\documentclass{article}

\usepackage[preprint]{neurips_2026}

\usepackage[utf8]{inputenc} 
\usepackage[T1]{fontenc}    
\usepackage{hyperref}       
\usepackage{url}            
\usepackage{booktabs}       
\usepackage{amsfonts}       
\usepackage{nicefrac}       
\usepackage{microtype}      
\usepackage{xcolor}         

\usepackage{xfrac}
\usepackage{amsmath}
\usepackage{bm}
\usepackage{amssymb}
\usepackage{accents}
\usepackage{graphicx}
\usepackage{mathrsfs}
\usepackage{float}

\newcommand{\vardbtilde}[1]{\tilde{\raisebox{0pt}[0.87\height]{$\tilde{#1}$}}}

\usepackage{amsthm}
\newtheoremstyle{neuripsresult}
  {3pt}   
  {3pt}   
  {}      
  {}      
  {\bfseries} 
  {.}     
  {.5em}  
  {\thmname{#1}\thmnumber{ #2}\thmnote{ \textbf{(#3)}}}

\theoremstyle{neuripsresult}
\newtheorem{assumption}{Assumption}[section]
\newtheorem{result}{Result}[section]

\title{Subcritical Signal Propagation at Initialization in Normalization-Free Transformers}

\author{%
  Sergey Alekseev \\
  Department of Physics and Astronomy\\
  Stony Brook University\\
  Stony Brook, NY 11794 \\
  \texttt{sergey.alekseev@stonybrook.edu}
}

\begin{document}

\maketitle

\begin{abstract}
We study signal propagation at initialization in transformers through the averaged partial Jacobian norm (APJN), a measure of gradient amplification across layers. We extend APJN analysis to transformers with bidirectional attention and permutation-symmetric input token configurations by deriving recurrence relations for activation statistics and APJNs across layers. 
Our theory predicts how attention modifies the asymptotic behavior of the APJN at large depth and matches APJNs measured in deep vision transformers. 
The criticality picture known from residual networks carries over to transformers: the pre-LayerNorm architecture exhibits power-law APJN growth, whereas transformers with LayerNorm replaced by elementwise $\tanh$-like nonlinearities have stretched-exponential APJN growth, indicating that the latter are subcritical. Applied to Dynamic Tanh (DyT) and Dynamic erf (Derf) transformers, the theory explains why these architectures can be more sensitive to initialization and optimization choices and require careful tuning for stable training.
\end{abstract}


\section{Introduction}

Trainability of deep neural networks is closely tied to signal propagation at initialization \citep{pmlr-v9-glorot10a, pmlr-v28-pascanu13, schoenholz2017deepinformationpropagation, pennington2017resurrectingsigmoiddeeplearning, noci2022signalpropagationtransformerstheoretical}. Signal propagation can be formalized through several related observables, including gradient norms \citep{pmlr-v28-pascanu13, xiong2020layernormalizationtransformerarchitecture, kedia2024transformersstableendtoendsignal}, the evolution of input correlations with depth \citep{schoenholz2017deepinformationpropagation, noci2022signalpropagationtransformerstheoretical}, and Jacobian-based quantities \citep{pmlr-v9-glorot10a, pennington2017resurrectingsigmoiddeeplearning, novak2018sensitivitygeneralizationneuralnetworks, doshi2023criticalinitializationwidedeep}. Across these views, stable training requires \textit{criticality}: a signal-propagation regime in which the signal neither vanishes nor explodes exponentially with depth. We study criticality through the averaged partial Jacobian norm (APJN), for which criticality means that the APJN remains finite or grows according to a power law with depth \citep{doshi2023criticalinitializationwidedeep}.

Architectural choices can strongly influence signal propagation and, in turn, training stability \citep{liu2023understandingdifficultytrainingtransformers,bachlechner2020rezeroneedfastconvergence,wang2022deepnetscalingtransformers1000,shleifer2021normformerimprovedtransformerpretraining}. Residual networks with pre-LayerNorm (pre-LN) achieve criticality for any weight initialization, so that gradients grow according to a power law from later layers to earlier ones, whereas other setups typically lead to exponential gradient growth or decay \citep{doshi2023criticalinitializationwidedeep}. In transformers, post-LayerNorm can lead to exponentially vanishing gradients \citep{kedia2024transformersstableendtoendsignal} and to larger gradient magnitudes in the last layers than pre-LN because of the difference in the final layer normalization \citep{xiong2020layernormalizationtransformerarchitecture}. As a result, pre-LN is generally more robust across optimization settings \citep{xiong2020layernormalizationtransformerarchitecture}.

Recent work replaces $\mathrm{LayerNorm}$ in transformers with elementwise $\tanh$-like nonlinearities, such as Dynamic Tanh (DyT) and Dynamic erf (Derf), motivated by the similarity of their input-output curves \citep{zhu2025transformersnormalization, chen2025strongernormalizationfreetransformers}. Following this literature, we refer to these architectures in the title as \textit{normalization-free} transformers. This raises a natural question: how does replacing LayerNorm with such nonlinearities affect signal propagation and training stability in deep transformers?

To address this question, we extend the APJN analysis of \citet{doshi2023criticalinitializationwidedeep} to transformers with bidirectional attention and permutation-symmetric input token configurations, following the framework of \citet{cowsik2024geometricdynamicssignalpropagation}. We show that replacing LayerNorm with elementwise $\tanh$-like nonlinearities leads to stretched-exponential APJN growth in deep layers, in contrast to the power-law growth of the pre-LN baseline. This regime was previously identified in residual networks and termed \textit{subcritical} \citep{yang2017meanfieldresidualnetworks, doshi2023criticalinitializationwidedeep}. Despite its simplifying assumptions, our theory captures both the scale and depth dependence of APJNs on real inputs in deep layers. In particular, it captures how attention modifies the scale parameter and critical exponent governing asymptotic APJN growth. Since the APJN provides a scalar measure of gradient amplification, this analysis also suggests a mechanism by which replacing LayerNorm with $\tanh$-like nonlinearities can make optimization more sensitive to architectural and initialization hyperparameters.

We further examine the optimization implications of this prediction through a
series of training-stability experiments on Vision Transformers
(ViTs) \citep{dosovitskiy2021imageworth16x16words}, comparing a pre-LN baseline with Derf variants initialized at different values of the scale parameter $\alpha$. In the tested ViT/CIFAR-100 settings, the
trainability of Derf models is more sensitive than that of pre-LN models to
hyperparameters such as model depth and weight initialization, especially at
larger $\alpha$. We also find that Derf models with smaller $\alpha$ train
stably over a wider range of learning rates, warmup durations, and network depths,
consistent with our theory that smaller $\alpha$ reduces gradient amplification.

This paper is organized as follows. Section~\ref{sec:background} introduces notation and relevant concepts, including the transformer architecture, the DyT/Derf operations, the APJN, and the assumptions underlying our signal propagation analysis. Section~\ref{sec:signal-prop} presents the signal propagation and APJN recurrences, analyzes their asymptotic behavior in deep networks, and compares the theoretical predictions with APJN values measured experimentally in a ViT model. Section~\ref{sec:stability-experiments} presents training stability experiments comparing Derf and pre-LN ViT models in the early stages of training.

\section{Background and Setup}\label{sec:background}

\subsection{Transformer}

Consider a transformer \citep{vaswani2017attention} with context size $n$ and $L$ layers, alternating between bidirectional self-attention and a position-wise MLP with ReLU nonlinearity. For notational simplicity, we restrict attention to the single-head case.

Let $\bm{h}_s^l \in \mathbb{R}^d$ denote the activation vector at position $s$ in layer $l$, with components $h_{s,i}^l$. Its evolution is given by
\begin{equation}
\bm{h}^{l+1}_s
=
\bm{h}^{l}_s
+
\begin{cases}
W_O^{l} W_V^{l} \displaystyle\sum_{t=1}^n A^{l}_{st}\,\tilde{\bm{h}}^{l}_{t},
& l \text{ even (attn)},\\[6pt]
W_2^{l}\,\mathrm{ReLU}\!\bigl(W_1^{l}\tilde{\bm{h}}^{l}_{s}\bigr),
& l \text{ odd (MLP)}.
\end{cases}
\label{eq:forward-pass}
\end{equation}
Here, $\tilde{\bm{h}}^{l}_{s}$ is the input to the residual branch after normalization or nonlinearity, defined below. The attention weights $A_{st}^l$ are defined in the standard way by softmax-normalized scaled dot products; see Eq.~(\ref{eq:attn-scores-mha}). We call each attention or MLP component a \textit{layer}, and each attention-MLP pair a \textit{block}. Thus $l=0,\ldots,L-1$ indexes layers, $b=0,\ldots,B-1$ indexes blocks, and $L=2B$.

All weights are initialized from zero-mean Gaussian distributions, with variances shared across transformer blocks. In the attention layers, $W_O$, $W_V \in \mathbb{R}^{d\times d}$ have component-wise variances $\sigma_O^2/d$ and $\sigma_V^2/d$, respectively. In the MLP layers, $W_1 \in \mathbb{R}^{4d\times d}$ and $W_2 \in \mathbb{R}^{d\times 4d}$ have component-wise variances $\sigma_1^2/d$ and $\sigma_2^2/(4d)$, respectively.


\subsection{LayerNorm and DyT/Derf}
The inputs to the residual branches in Eq.~(\ref{eq:forward-pass}), denoted by $\tilde{\bm h}_s^l$, are obtained from $\bm h_s^l$ either by applying LayerNorm \citep{ba2016layernormalization} or by an elementwise transformation:
\begin{equation}
\tilde{\bm h}_s^l =
\mathrm{LayerNorm}(\bm h_s^l),
\qquad \text{or} \qquad
\tilde{\bm h}_s^l = \bm\gamma \odot \phi_\alpha(\bm h_s^l) + \bm\beta,
\label{eq:norm-act-vector}
\end{equation}
where $\alpha \in \mathbb{R}$, $\bm{\beta}, \bm{\gamma} \in \mathbb{R}^d$ are learnable parameters, $\odot$ denotes elementwise multiplication, and the nonlinearity $\phi_\alpha$ is applied elementwise.

We focus on $\tanh$-like nonlinearities, namely antisymmetric, monotonically increasing functions that approach $1$ as their argument tends to $+\infty$. The main examples we consider are
\begin{equation}
\phi_\alpha(h)=\tanh(\alpha h)
\qquad \text{and} \qquad
\phi_\alpha(h)=\mathrm{erf}(\alpha h),
\end{equation}
corresponding to Dynamic Tanh (DyT) and Dynamic erf (Derf), respectively \citep{zhu2025transformersnormalization,chen2025strongernormalizationfreetransformers}.\footnote{The term ``dynamic'' refers to the fact that $\alpha$ is a learnable parameter; however, this does not play a role in the signal propagation analysis at initialization considered below.} When the dependence on $\alpha$ is not central, we suppress it and write simply $\phi$.

Since our theory concerns signal propagation at initialization, we set $\bm\gamma=\bm 1$ and $\bm\beta=\bm 0$, which is their standard initialization in both LayerNorm and DyT/Derf, and suppress them in the notation below. The initialization of $\alpha$ is one of the aspects studied below. For LayerNorm, we additionally assume that the feature-wise mean is close to zero, so that LayerNorm effectively reduces to RMSNorm \citep{zhang2019rootmeansquarelayer}.

\subsection{Signal Propagation at Initialization} \label{mft}

We make three simplifying assumptions to derive closed signal-propagation recurrences at initialization.

\begin{assumption}[Signal-propagation setting]
\label{ass:signal-prop}
\leavevmode
\begin{enumerate}
    \item \textit{Permutation-symmetric token configurations.} Following \citet{cowsik2024geometricdynamicssignalpropagation}, we restrict our analysis to permutation-symmetric input token configurations, in which all token vectors have the same squared norm and pairwise dot product. 
    This symmetry is preserved across layers in expectation, so the layerwise geometry of the activation vectors is fully characterized by two quantities: the normalized self-dot product
$q^l=\mathbb{E}_\theta\!\left[\bm{h}^{l}_s \cdot \bm{h}^{l}_s/d\right]$,
which is independent of $s$, and the normalized dot product between
activation vectors at different positions, $
p^l=\mathbb{E}_\theta\!\left[\bm{h}^{l}_s \cdot \bm{h}^{l}_t/d\right],\ s\neq t$,
which is independent of the choice of distinct positions $s,t$. Here,
$\mathbb{E}_\theta$ denotes expectation over random weight initializations.
    \item \textit{Gaussian mean-field approximation.} We adopt the large-width mean-field approximation
\citep{poole2016exponentialexpressivitydeepneural,
schoenholz2017deepinformationpropagation,
cowsik2024geometricdynamicssignalpropagation}, under which the components of the activation vectors at layer $l$ are modeled as jointly Gaussian. Specifically, for any feature index $i$ and any two positions $s \ne t$,
    \begin{equation}
    (h_{s,i}^l,\, h_{t,i}^l) \sim \mathcal{N}(0,\,\Sigma^l),
    \qquad
    \Sigma^l =
    \begin{pmatrix}
    q^l & p^l \\
    p^l & q^l
    \end{pmatrix},
    \label{eq:gaussian-mean-field}
    \end{equation}
    while components with different feature indices are uncorrelated.
    \item \textit{Uniform attention at initialization.} At initialization, we use the uniform-attention approximation $A^l_{st}\approx 1/n$, following \citet{noci2022signalpropagationtransformerstheoretical}. Under this approximation, attention acts as uniform mixing of the activation vectors.
\end{enumerate}
\end{assumption}

\subsection{Averaged Partial Jacobian Norm}\label{subsec:apjn}
The Jacobian of the mapping from layer $l$ to layer $l'$, with $l' > l$, is defined as
\begin{equation}
    J^{{l'},l}_{si, tj}=\frac{\partial h_{s,i}^{l'}}{\partial {h}^{l}_{t,j}}.
\end{equation}
It characterizes both forward perturbation propagation and backward gradient propagation between layers $l$ and $l'$ \citep{schoenholz2017deepinformationpropagation}. Its squared Frobenius norm, averaged over weight initializations, defines the \textit{averaged partial Jacobian norm (APJN)}, a scalar measure of gradient amplification or attenuation under backpropagation from layer $l'$ to layer $l$ \citep{doshi2023criticalinitializationwidedeep}:
\begin{equation}
\mathcal{J}^{{l'},l}
= \frac{1}{nd}\, \mathbb{E}_\theta\lVert J^{{l'},l}\rVert_F^2= \frac{1}{nd}\mathbb{E}_\theta \left[\sum_{st,ij}\left(\frac{\partial {h}_{s,i}^{l'}}{\partial {h}^{l}_{t,j}}\right)^2\right].
\label{eq:apjn-def}
\end{equation}

\textbf{Notation and terminology.} We refer to $\mathcal{J}^{L,l}$, viewed as a function of $l$, as the \textit{backward APJN}, and to $\mathcal{J}^{l,0}$ as the \textit{forward APJN}. At block level, we define
\begin{equation}
    Q^{b}=q^{2b},\quad P^b=p^{2b},\quad \mathscr{J}^{b',b}=\mathcal{J}^{2b',\,2b},
    \label{eq:block-level-notation}
\end{equation}
We use the same terminology for the block-level functions $\mathscr{J}^{B,b}$ and $\mathscr{J}^{b,0}$, referring to them as the \textit{backward APJN} and \textit{forward APJN}, respectively; the distinction from layer-level quantities will be clear from context.

\section{Theory of Signal Propagation}\label{sec:signal-prop}

\subsection{Covariance Recurrence Relations and APJN}

We now present covariance and APJN recurrences for transformers with either LayerNorm or an elementwise nonlinearity replacing LayerNorm. Under standard initialization, the same recurrences hold for multi-head attention; see Appendices~\ref{subsubsec:cov-prop-mha} and \ref{subsubsec:j_k_recurrence_attn}. We then compare the resulting APJN predictions with measurements. Derivations are provided in Appendices~\ref{sec:cov-prop-derivation-appendix} and \ref{sec:apjn-recurrence-derivation-appendix}.

\begin{result}[Covariance recurrence relations]
\label{res:covariance-recurrence}
Under Assumption~\ref{ass:signal-prop}, and in the large-context limit, Eq.~(\ref{eq:forward-pass}) gives the recurrence relation
\begin{equation}
q^{l+1}
=
q^{l}
+
\begin{cases}
\sigma_{OV}^2\tilde p^l,
& \text{if } l \text{ is even (attn)},\\[6pt]
\dfrac{1}{2}\sigma_{21}^2\tilde q^{l},
& \text{if } l \text{ is odd (MLP),}
\end{cases}
\label{eq:eq-q-recurrence}
\end{equation}
and similarly
\begin{equation}
p^{l+1}
= p^{l} +
\begin{cases}
\sigma_{OV}^2\tilde p^l,
& \text{if } l \text{ is even (attn)},\\[8pt]
\dfrac{1}{2}\sigma_{21}^2\tilde q^{l}\kappa\!\left(\dfrac{\tilde p^{l}}{\tilde q^{l}}\right),
& \text{if } l \text{ is odd (MLP).}
\end{cases}
\label{eq:eq-p-recurrence}
\end{equation}
Here $\sigma_{OV}=\sigma_O\sigma_V$ and $\sigma_{21}=\sigma_2\sigma_1$, while $\tilde q^l$ and $\tilde p^l$ denote the covariance components after propagation through either an elementwise nonlinearity or $\mathrm{LayerNorm}$. The function $\kappa$, which comes from covariance propagation through ReLU, is given in Eq.~(\ref{eq:kappa}) of Appendix~\ref{subsec:cov_prop_mlp_derivation}.
\end{result}

For an elementwise nonlinearity $\phi$, the propagated covariance components are defined by
\begin{equation}
\tilde \Sigma_{st}^l
=
\mathbb{E}_{(h_1,h_2)\sim \mathcal{N}(0,\,\Sigma^l)}
\bigl[\phi(h_s)\phi(h_t)\bigr],
\qquad s,t \in \{1,2\},
\label{eq:norm-cov-prop}
\end{equation}
where $\tilde\Sigma^l_{11}=\tilde\Sigma^l_{22}=\tilde q^l$, $\tilde\Sigma^l_{12}=\tilde\Sigma^l_{21}=\tilde p^l$, and $\Sigma^l_{11}=\Sigma^l_{22}=q^l$, $\Sigma^l_{12}=\Sigma^l_{21}=p^l$. In the LayerNorm case, $\tilde q^l=1$ and $\tilde p^l=p^l/q^l$, since LayerNorm rescales each activation vector to unit normalized squared norm while preserving the cosine similarity between activation vectors.

The recurrence relations in Eqs.~(\ref{eq:eq-q-recurrence}) and (\ref{eq:eq-p-recurrence}) are supplemented by the initial conditions $(q^0,p^0)$. Although permutation-symmetric input token configurations may admit negative values of $p^0$, we restrict the analysis to the simplified setting $p^0 \ge 0$. Related recurrence relations were derived, for example, by \citet{cowsik2024geometricdynamicssignalpropagation}; our derivation uses a simplified uniform-attention approximation and extends the analysis beyond LayerNorm.

\begin{result}[Simplified APJN recurrence]
\label{res:apjn-recurrence}
Under Assumption~\ref{ass:signal-prop} and in the large-context limit, after neglecting $\mathcal{O}(n^{-1})$ attention contributions and the cross-positional Jacobian correlation term defined in Appendix~\ref{sec:apjn-recurrence-derivation-appendix}, the forward APJN satisfies
\begin{equation}
    \mathcal{J}^{l+1,0}
    =
    \chi^l_{\mathcal J}\mathcal{J}^{l,0},
    \qquad l\ge 0,
\label{eq:apjn-recurrence}
\end{equation}
where
\begin{equation}
\chi^l_{\mathcal J}
=
\begin{cases}
1,
& \text{if } l \text{ is even (attn)},\\[6pt]
1+\dfrac{1}{2}\sigma_{21}^2\hat q^{\,l},
& \text{if } l \text{ is odd (MLP).}
\end{cases}
\label{eq:eq-chi-J}
\end{equation}
Here, $\hat q^l$ denotes the variance obtained by propagating $q^l$ through the Jacobian of the nonlinearity $\phi$ or of $\mathrm{LayerNorm}$.
\end{result}

For elementwise functions,
\begin{equation}
\hat q^l = \mathbb{E}_{h \sim \mathcal{N}(0,\,q^l)} \left[\phi'(h)^2\right],
\label{eq:q-hat}
\end{equation}
where the prime denotes differentiation. For $\mathrm{LayerNorm}$, $\hat q^l=1/q^l$, as follows from its Jacobian. For the specific choice $\phi_\alpha(h)=\mathrm{erf}(\alpha h)$, explicit formulas for $\tilde q^l$, $\tilde p^l$, and $\hat q^l$ are given in Appendix~\ref{sec:derf-explicit}.

The APJN recurrence relation in Eq.~(\ref{eq:apjn-recurrence}) is supplemented by the initial condition $\mathcal{J}^{0,0}=1$. The sequence $\mathcal{J}^{l,0}$ is obtained by first solving the covariance propagation equations, Eqs.~(\ref{eq:eq-q-recurrence}) and (\ref{eq:eq-p-recurrence}), and then evaluating Eq.~(\ref{eq:eq-chi-J}) using the resulting $q^l$. Under the approximation described above, the backward APJN is given by $\mathcal{J}^{L,l}=\mathcal{J}^{L,0}/\mathcal{J}^{l,0}$.

In this simplified recurrence, attention affects the APJN only indirectly through its effect on covariance propagation in Eqs.~(\ref{eq:eq-q-recurrence}) and (\ref{eq:eq-p-recurrence}). The recurrence is analogous to that derived in \citet{doshi2023criticalinitializationwidedeep} for feedforward and residual networks without attention in the large-width limit. In Appendix~\ref{sec:apjn-recurrence-derivation-appendix}, we present and derive the extended recurrence, which involves, in addition to the APJN, a cross-positional Jacobian correlation term that may become non-negligible for larger values of $\sigma_{OV}^2$.\footnote{In all figures showing or using theoretical APJNs, we use the extended theory for $(\sigma_{21},\,\sigma_{OV})=(0.6,\,1.2)$ without stating this explicitly each time, since it provides a better match to the empirical APJN measurements. Fig.~\ref{fig:k_j_ratio} in Appendix~\ref{sec-app:supplement-sig-prop} shows that, in this case, the contribution of the neglected terms is no longer non-negligible.}

\begin{figure}
  \centering
\includegraphics[width=14.0cm]{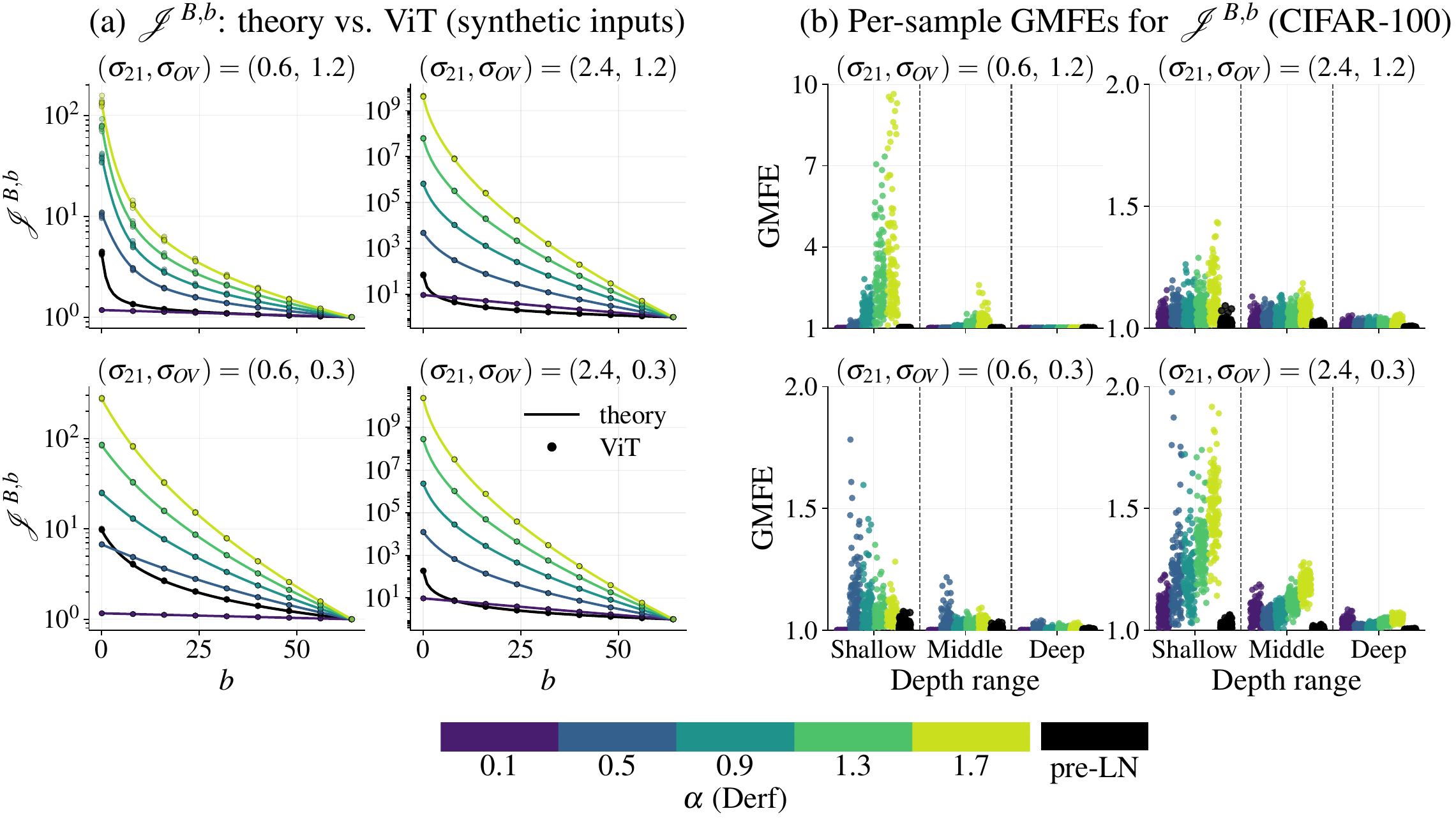}
\caption{
\textbf{(a)} Backward APJN $\mathscr{J}^{B,b}$ on synthetic permutation-symmetric inputs: ViT measurements versus theory for pre-LN and Derf variants. 
\textbf{(b)} Per-sample GMFE values, interpreted as typical multiplicative error factors, between theoretical and measured $\mathscr{J}^{B,b}$ values in a 128-block ViT on 120 CIFAR-100 inputs, evaluated in early, middle, and deep blocks.
}
  \label{fig:q_pq_apjn}
\end{figure}

\textbf{Comparison with experiments.} We compare the backward APJN predicted by the theoretical recurrence relations with those measured in ViT across several setups. Our main findings are summarized below; experimental details, including the ViT configuration, are provided in Appendix~\ref{sec-app:exp-details}, and additional figures and results are given in Appendix~\ref{sec-app:supplement-sig-prop}.

\textbf{(1) On synthetic permutation-symmetric input token configurations, the theory accurately predicts backward APJNs in both pre-LN and Derf ViTs.}
We compare the backward APJNs $\mathscr{J}^{B,b}$ measured in a ViT model on synthetic permutation-symmetric inputs with $(q^0,p^0)=(1.0,\,0.2)$ to the predictions of Eqs.~(\ref{eq:apjn-recurrence}) and (\ref{eq:eq-chi-J}). We consider both the pre-LN model and Derf variants with multiple values of $\alpha$. As shown in Fig.~\ref{fig:q_pq_apjn}(a), the agreement is strong, consistent with the fact that these inputs satisfy the simplifying assumptions of the theory. The corresponding values of $Q^b$, $P^b$, and $P^b/Q^b$ are shown in Fig.~\ref{fig:equiang_p_q_pq}.

\begin{figure}
  \centering
\includegraphics[width=14.0cm]{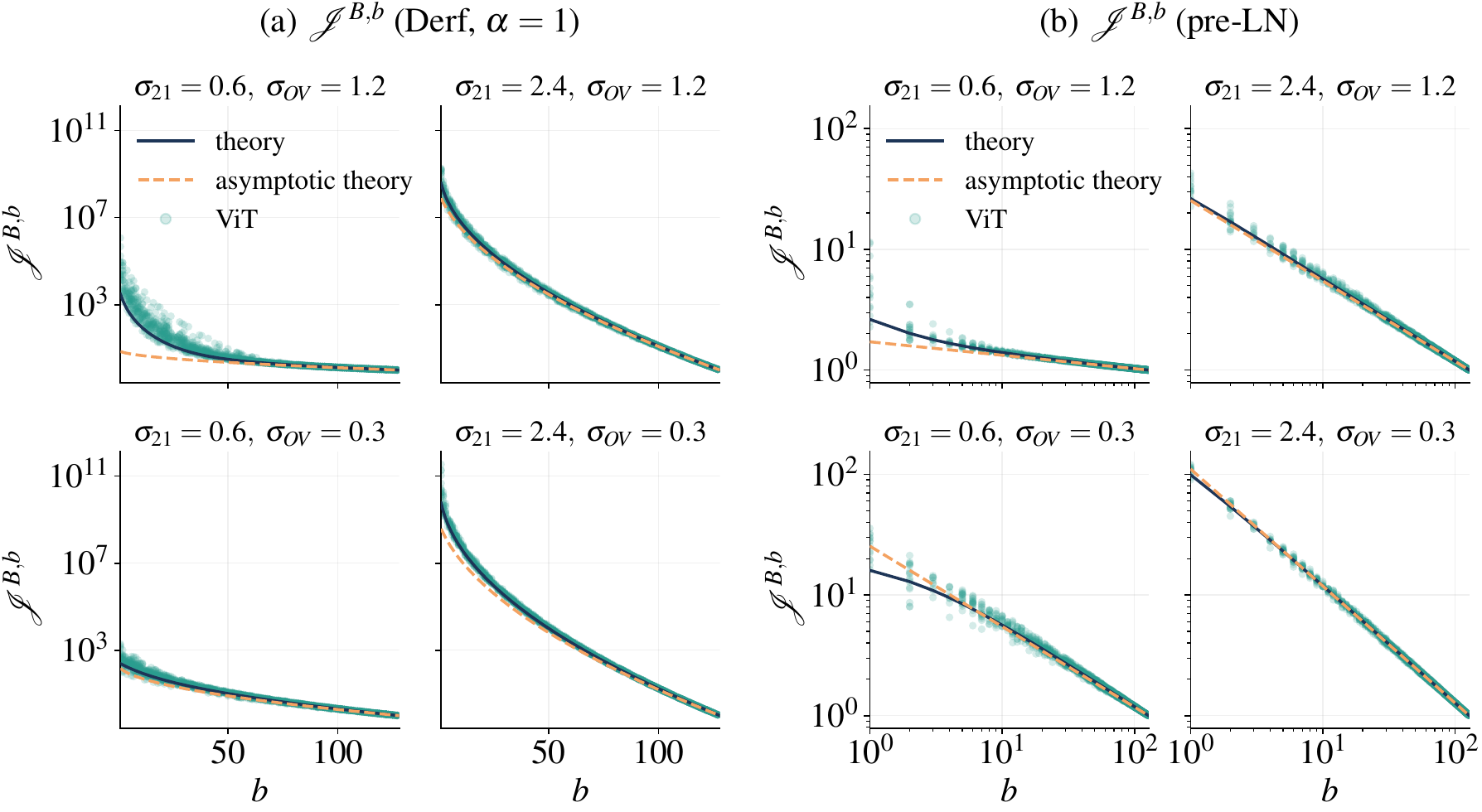}
\caption{
Backward APJNs for Derf and pre-LN at several values of $(\sigma_{21},\sigma_{OV})$: ViT vs.\ theory.
\textbf{(a)} Derf with $\alpha=1$. \textbf{(b)} pre-LN.
Teal circles show ViT measurements of $\mathscr{J}^{B,b}$ on 100 random CIFAR-100 samples. Black solid lines show full-theory predictions from Eqs.~(\ref{eq:eq-q-recurrence})-(\ref{eq:eq-chi-J}) with $(q^0,p^0)=(0.5,\,0.25)$, and orange dashed lines show the large-$b$ asymptotics: Eq.~(\ref{eq:derf-asymptotic-apjn}) in (a) and Eq.~(\ref{eq:pre-ln-asymptotic-apjn}) in (b).
}
  \label{fig:asymp}
\end{figure}

\textbf{(2) On CIFAR-100, the theoretical backward-APJN curves approximate measured APJNs in middle and deep layers within modest multiplicative error.} We compare the theory with backward APJNs measured on CIFAR-100 samples using a 128-block ViT, for both pre-LN and Derf variants across several values of $\alpha$. Since CIFAR-100 inputs do not satisfy the simplifying assumptions of the theory, we set the per-sample initial condition by taking $q^0$ to be the empirical average normalized self-dot product and $p^0$ to be the empirical average normalized cross-positional dot product at the input to the first transformer block. As shown in Fig.~\ref{fig:q_pq_apjn}(b), the resulting theoretical curves approximate the measured backward APJNs in middle and deep layers, with per-sample geometric mean fold errors (GMFEs) typically below 1.25 and often substantially lower. GMFE is interpreted as a typical multiplicative error factor and is defined in Appendix~\ref{sec-app:exp-details}.

The reasonable agreement between the theory and measured backward APJNs on CIFAR-100 inputs is consistent with the observation that, as depth increases, the relative fluctuations of the self- and cross-positional dot-product statistics across positions decrease, making the empirical activations closer to the permutation-symmetric setting assumed by the theory; see Figs.~\ref{fig:qp_within_hist_preln} and \ref{fig:qp_within_hist_derf}. The most challenging case for the theory is $(\sigma_{21}, \sigma_{OV})=(0.6,\,1.2)$, where larger attention weights may make the APJNs more sensitive to violations of the theoretical assumptions.
Representative comparisons and analogous forward-APJN results are shown in Figs.~\ref{fig:apjn-fits} and \ref{fig:forward-ajns-synth-gmfes}.

Fig.~\ref{fig:asymp} further illustrates the depth dependence and sample-to-sample variability of backward APJNs on 100 CIFAR-100 training samples, comparing ViT measurements with theory for Derf with $\alpha=1$ and for pre-LN. The theoretical curves are computed using the fixed initial condition $(q^0,p^0)=(0.5,\,0.25)$, chosen to be close to the average of the empirical per-sample $(q^0,p^0)$ values. Agreement improves in deeper blocks, where $\mathscr{J}^{B,b}$ has smaller sample-to-sample variance; see Figs.~\ref{fig:qp_uniformize_preln} and \ref{fig:qp_uniformize_derf} Fig.~\ref{fig:two_alpha_theory_asympt} shows analogous plots for Derf models with different values of $\alpha$.

\subsection{$\mathrm{LayerNorm}$ vs. $\tanh$-Like Nonlinearity: Asymptotic Behavior}
\label{subsec:apjn-asymptotics}

We now summarize large-depth APJN behavior for LayerNorm transformers and for transformers where LayerNorm is replaced by a $\tanh$-like nonlinearity. Since attention and MLP layers affect covariance dynamics and APJNs differently, we work at the block level, using $Q^b$, $P^b$, $\mathscr{J}^{b,0}$, and $\mathscr{J}^{B,b}$ from Eq.~(\ref{eq:block-level-notation}).

In both architectures, the activation variance $Q^b$ grows linearly with depth. The difference in APJN behavior comes from the dependence of $\hat q$ on $q$: for LayerNorm, $\hat q=1/q$, leading to power-law APJN growth, whereas for $\tanh$-like nonlinearities, $\hat q\sim 1/\sqrt{q}$ at large depth, leading to stretched-exponential APJN growth. We refer the reader to Appendix~\ref{sec:large-l-derivation-appendix} for details.

\begin{result}[Asymptotic growth of $Q^b$]
\label{res:asymptotic-covariance-growth}
Using the covariance recurrence in Result~\ref{res:covariance-recurrence}, assume that $\tilde q^l$ and $\tilde p^l$ approach limiting values $\tilde q_\star$ and $\tilde p_\star$ as $l\to\infty$. Then, for both LayerNorm and $\tanh$-like nonlinearities, at large block index $b$,
\begin{equation}
    Q^{b} \sim b\left(\frac{1}{2}\sigma_{21}^2\tilde q_\star + \sigma_{OV}^2\tilde p_\star\right).
\label{eq:q-asymptotic}
\end{equation}
In both cases, $\tilde q_\star=1$. The value of $\tilde p_\star$ is determined by the asymptotic cosine similarity between activation vectors at different positions, denoted by $c_\star$, via $\tilde p_\star=\tilde p(c_\star)$, where $\tilde p(c)=\frac{2}{\pi}\arcsin c$ for $\tanh$-like nonlinearities and $\tilde p(c)=c$ for LayerNorm. The value $c_\star$ satisfies
\begin{equation}
    c_\star = \frac{
    \frac{1}{2}\sigma_{21}^2\kappa\left(\tilde p(c_\star)\right)+
    \sigma_{OV}^2\tilde p(c_\star)}{
    \frac{1}{2}\sigma_{21}^2+
    \sigma_{OV}^2\tilde p(c_\star)}.
\label{eq:c-star-fixed-point}
\end{equation}
\end{result}

The linear growth in Eq.~(\ref{eq:q-asymptotic}) follows directly from the covariance recurrence in Eq.~(\ref{eq:eq-q-recurrence}): once $\tilde q^l$ and $\tilde p^l$ approach limiting values, each transformer block adds an asymptotically constant contribution to $Q^b$. In both architectures considered here, $\tilde q_\star=1$; this is immediate for LayerNorm and follows from saturation of $\phi$ at large pre-activations for $\tanh$-like nonlinearities.

For LayerNorm, the unique solution of Eq.~(\ref{eq:c-star-fixed-point}) is $c_\star=1$, so $\tilde p_\star=1$. For $\tanh$-like nonlinearities with $\sigma_{21}^2\ne0$, Eq.~(\ref{eq:c-star-fixed-point}) has the solution $c_\star=1$ and another solution with $c_\star<1$; the former is unstable, so the system generically converges to the stable solution with $c_\star<1$. The derivation of Eq.~(\ref{eq:c-star-fixed-point}) and the stability analysis are given in Appendix~\ref{subsubsec:fixed-point-derivation}.

\begin{figure}
  \centering
\includegraphics[width=12.0cm]{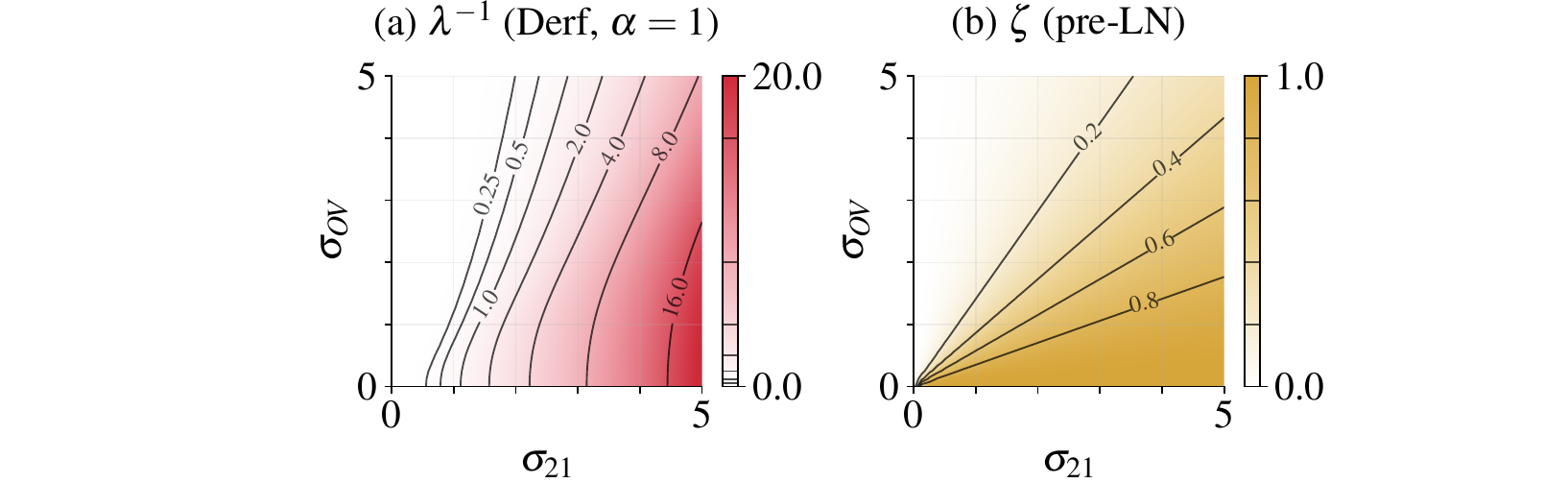}
  \caption{
Theoretical values of the scale parameter $\lambda^{-1}$ and the critical exponent $\zeta$ that determine the asymptotic behavior of the forward and backward APJNs.
\textbf{(a)} The scale parameter $\lambda^{-1}$ as a function of $\sigma_{21}$ and $\sigma_{OV}$ for Derf with $\alpha=1$. \textbf{(b)} The critical exponent $\zeta$ as a function of $\sigma_{21}$ and $\sigma_{OV}$ for pre-LN.
}
  \label{fig:asymp_lambda_zeta}
\end{figure}

\begin{result}[Pre-LN: power-law APJN growth]
\label{res:pre-ln-apjn-asymptotics}
Combining the asymptotic growth of $Q^b$ in Result~\ref{res:asymptotic-covariance-growth} with the simplified APJN recurrence of Result~\ref{res:apjn-recurrence}, the pre-LN APJNs satisfy, at large $b$,
\begin{equation}
 \mathscr{J}^{b,0}\sim b^{\zeta},
 \qquad
 \mathscr{J}^{B,b}\sim \left(\frac{B}{b}\right)^\zeta,
 \label{eq:pre-ln-asymptotic-apjn}
\end{equation}
where
\begin{equation}
\zeta = \frac{\frac{1}{2}\sigma_{21}^2}{\frac{1}{2}\sigma_{21}^2+\sigma_{OV}^2}.
\label{eq:zeta}
\end{equation}
\end{result}

Thus, pre-LN lies in the \emph{critical} regime of \citet{doshi2023criticalinitializationwidedeep}, characterized by power-law growth of the APJN with critical exponent $\zeta$. Specifically, the forward APJN increases with the block index, whereas the backward APJN increases toward earlier blocks. The derivation is provided in Appendix~\ref{sec:large-l-derivation-appendix}.

\begin{result}[$\tanh$-like nonlinearities: stretched-exponential APJN growth]
\label{res:tanh-like-apjn-asymptotics}
For $\tanh$-like nonlinearities $\phi_\alpha$, the APJNs satisfy, at large $b$,
\begin{equation}
\mathscr{J}^{b, 0} \sim e^{\sqrt{b/\lambda}},\qquad
\mathscr{J}^{B, b} \sim e^{\left(\sqrt{B}-\sqrt{b}\right)/\sqrt{\lambda}},
\label{eq:derf-asymptotic-apjn}
\end{equation}
with scale parameter
\begin{equation}
    \lambda^{-1}
    =
    \frac{C_\alpha^2\sigma_{21}^4}{
    \frac{1}{2}\sigma_{21}^2+\sigma_{OV}^2\tilde p_\star},
    \qquad
    C_\alpha=\frac{1}{\sqrt{2\pi}}\int_\mathbb{R}\phi'_\alpha(h)^2\,dh.
\label{eq:lambda}
\end{equation}
\end{result}

Following \citet{doshi2023criticalinitializationwidedeep}, the network lies in the \emph{subcritical} regime at initialization. In this regime, the forward APJN grows asymptotically faster than any power law, and therefore faster than in pre-LN. Likewise, for sufficiently large network depth, the backward APJN increases more rapidly than in pre-LN from the final block toward earlier blocks. Moreover, unlike the bounded pre-LN exponent $\zeta$ in Eq.~(\ref{eq:zeta}), the scale parameter $\lambda^{-1}$ is unbounded, so the stretched-exponential APJN growth can be made arbitrarily steep by varying the weight initialization. The derivation, including the refined asymptotic expression with the power-law prefactor in Eq.~(\ref{eq:derf-asymp-back-apjn-prefactor}), is provided in Appendix~\ref{subsubsec-app:tanh-asymp-apjn}.

\textbf{Role of $\alpha$ in DyT/Derf.} Decreasing $\alpha$ in the nonlinearity $\phi_\alpha$ reduces the effective residual-branch contribution and therefore slows the growth of both $Q^b$ and the APJN. This has two consequences. First, the asymptotic expression in Eq.~(\ref{eq:derf-asymptotic-apjn}) becomes valid only at greater depths for smaller $\alpha$. Second, at fixed depth, sufficiently small $\alpha$ can make Derf APJNs comparable to pre-LN, as in Fig.~\ref{fig:q_pq_apjn}(a), but matching pre-LN at increasing depth requires progressively smaller $\alpha$.


\textbf{Comparison with experiment.}
Fig.~\ref{fig:asymp} compares backward APJN values measured in ViT for pre-LN and Derf with the corresponding asymptotic predictions: Eq.~(\ref{eq:pre-ln-asymptotic-apjn}) for pre-LN, and for Derf the refined asymptotic expression in Eq.~(\ref{eq:derf-asymp-back-apjn-prefactor}), including the power-law prefactor, derived in Appendix~\ref{subsubsec-app:tanh-asymp-apjn}. In both cases, the asymptotic expressions become increasingly accurate as the depth increases. Fig.~\ref{fig:asymp_lambda_zeta} shows the theoretical values of the scale parameter $\lambda^{-1}$ and the critical exponent $\zeta$ as functions of $\sigma_{21}$ and $\sigma_{OV}$. Consistent with Eqs.~(\ref{eq:pre-ln-asymptotic-apjn}) and (\ref{eq:derf-asymptotic-apjn}), these quantities determine the slope of the asymptotic APJN curves in the two regimes.

\begin{figure}
  \centering
\includegraphics[width=14.0cm]{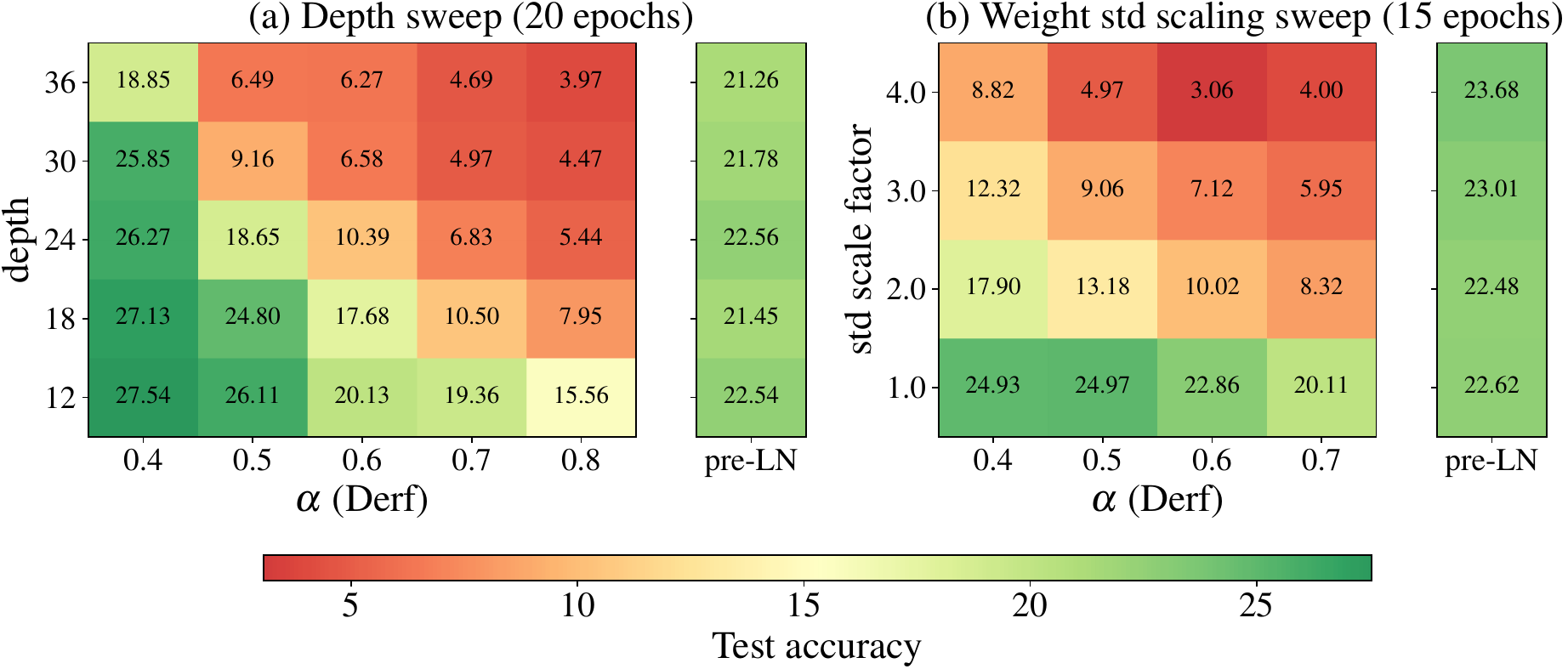}
  \caption{
Training stability comparison between Derf and pre-LN ViTs. \textbf{(a)} Test accuracy on CIFAR-100 after $20$ epochs of training versus model depth.
\textbf{(b)} Test accuracy on CIFAR-100 after $15$ epochs of training versus the MLP and attention weight standard deviation scale (depth $B=12$). 
}
  \label{fig:depth_std_tiles}
\end{figure}

\section{Training Stability Experiments}\label{sec:stability-experiments}

The theory above suggests several empirical predictions. Since the backward APJN provides a scalar measure of gradient amplification during backpropagation, Derf is expected to be less robust to hyperparameter choices than pre-LN. Specifically, increasing model depth or the weight-initialization scale is expected to impair trainability more strongly in Derf than in pre-LN, whereas decreasing $\alpha$ is expected to improve trainability in Derf by reducing gradient amplification, both across transformer blocks and at the final Derf layer; the latter effect is discussed in Appendix~\ref{sec-app:final-norm-role}.

We test these predictions in ViT models trained on CIFAR-100, comparing Derf and pre-LN during early training. Although fully training a ViT model on CIFAR-100 may require hundreds of epochs, we typically truncate training after 10--20 epochs, which is sufficient to identify configurations with pathologically slow convergence. Model and training details are provided in Appendix~\ref{sec-app:exp-details}. Supplementary experiments and example training/test curves are provided in Appendix~\ref{sec-app:supplement-stability-exps}.

The main results are as follows.

\textbf{(1) Derf trainability is more sensitive to model depth than pre-LN trainability.}
Fig.~\ref{fig:depth_std_tiles}(a) shows that Derf models train more slowly as depth increases and eventually fail to converge beyond a critical depth, with smaller initialization values of $\alpha$ pushing this instability to greater depth.
We choose the range of $\alpha$ values so that this instability boundary is visible. Although we expect the destabilization to occur even for smaller values of $\alpha$, it becomes less relevant at depths typically used in practice.
By comparison, the pre-LN baseline, with the same weight initialization, maintains an approximately constant convergence rate across the full tested depth range without hyperparameter tuning.

\textbf{(2) Derf trainability is more sensitive to weight initialization than pre-LN trainability.}
We vary the initialization scale of the MLP and attention weights by a common factor up to $4.0$, excluding query and key weights to avoid entropy-collapse effects \citep{zhai2023stabilizingtransformertrainingpreventing,giorlandino2026failuremodesdeeptransformers}, while keeping the model depth fixed at $B=12$.
Fig.~\ref{fig:depth_std_tiles}(b) shows that Derf variants with larger $\alpha$ tend to become unstable at smaller weight initialization scales, whereas the pre-LN baseline remains stable over the entire range considered.

\begin{figure}
  \centering
\includegraphics[width=14.0cm]{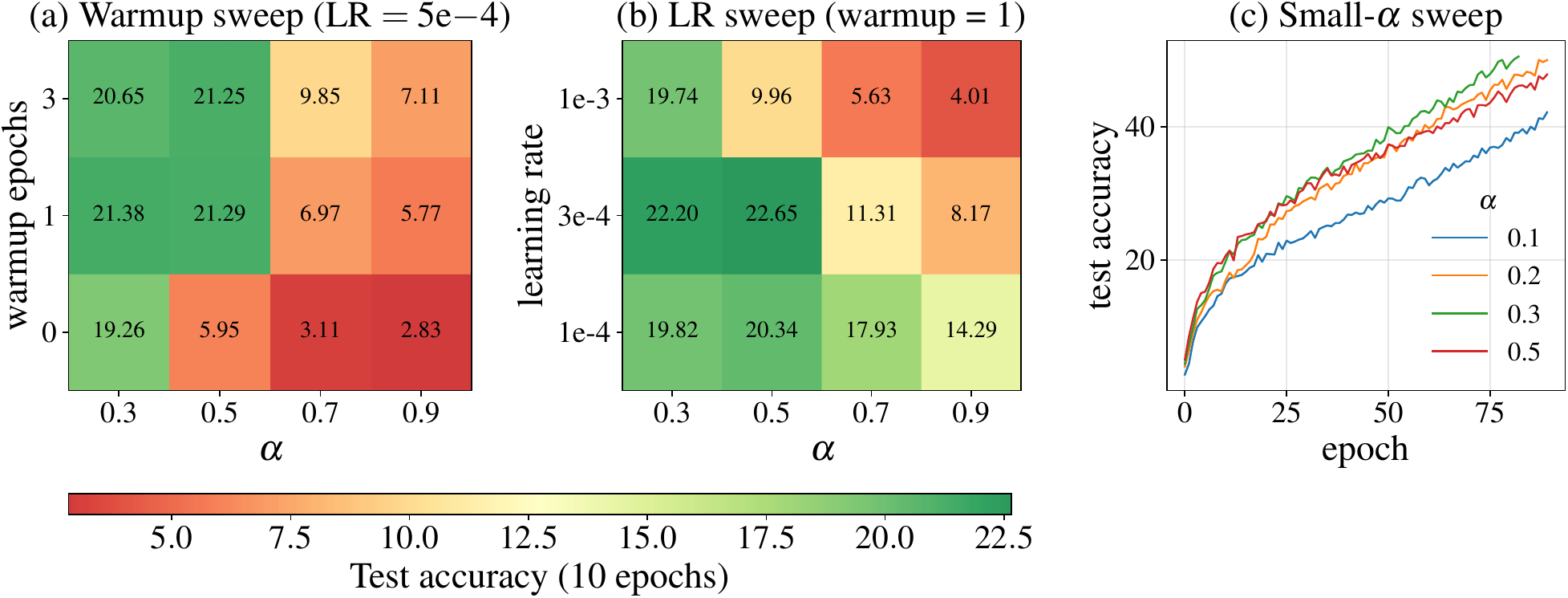}
\caption{
Effects of warmup, learning rate, and small initial $\alpha$ in Derf ViTs on CIFAR-100.
\textbf{(a)} Test accuracy after $10$ epochs versus number of warmup epochs.
\textbf{(b)} Test accuracy after $10$ epochs versus learning rate.
\textbf{(c)} Test accuracy over $90$ epochs for Derf models initialized with small $\alpha$.
}

  \label{fig:warmup_lr_tiles}
\end{figure}

\textbf{(3) Derf models with larger $\alpha$ benefit from a smaller learning rate or a longer warmup.}
At fixed depth $B=12$, we separately sweep the number of warmup epochs and the maximum learning rate. Figs.~\ref{fig:warmup_lr_tiles}(a) and (b) show that both changes improve the trainability of Derf models at larger $\alpha$, with reducing the learning rate having a particularly strong effect.

\textbf{(4) Choosing an overly small initial value of $\alpha$ in Derf models can slow convergence.}
Improving stability by decreasing $\alpha$ comes with a trade-off. Fig.~\ref{fig:warmup_lr_tiles}(c) shows that choosing $\alpha$ to be too small, such as $\alpha=0.1$, can substantially slow convergence. Fig.~\ref{fig:appendix-small-alpha} in Appendix~\ref{sec-app:supplement-stability-exps} shows that this effect is robust across different settings.

Overall, these results agree qualitatively with the predicted role of gradient amplification, which we measure directly in Appendix~\ref{subsec-app:grad_amplification}. Although $\alpha$ is trainable in Derf models, its initialization remains important; freezing $\alpha$ can further worsen convergence, as shown in Fig.~\ref{fig:frozen-alpha-warmup-lr-tiles}.

\section{Conclusion}

We extended the APJN framework to transformers with uniform bidirectional attention and permutation-symmetric input token configurations. Despite these simplifying assumptions, the theory reasonably approximates APJNs measured in deeper ViT layers on real inputs. We derived APJN asymptotics for both LayerNorm transformers and transformers in which LayerNorm is replaced by $\tanh$-like nonlinearities, showing that the latter exhibit stretched-exponential APJN growth, as in residual networks without attention.

The subcritical behavior of DyT/Derf, together with the overall gradient rescaling induced by the final normalization layer, can lead to larger gradients and thus requires more careful tuning of initialization parameters, such as the weight standard deviation and the parameter $\alpha$. Across several experiments comparing the training stability of Derf models initialized with different values of $\alpha$ with that of the pre-LN baseline, we observed the trends predicted by the theory. 

\textbf{Limitations.}
Our experiments use ViT models on CIFAR-100, limiting conclusions about other architectures, modalities, or tasks. Although truncated runs reveal early instability boundaries, they do not assess long-term training dynamics or final performance.
Theoretically, extending the framework to more complex token geometries remains important.
Finally, since the theory describes signal propagation at initialization, it explains trends in trainability and gradient amplification, but not full training dynamics or final performance.


{\small
\bibliographystyle{plainnat} 
\bibliography{references}
}


\appendix
\section*{Appendix}

The appendix is organized as follows. Sec.~\ref{sec-app:exp-details} provides experimental details for the results presented in the main text. Sec.~\ref{sec:cov-prop-derivation-appendix} contains derivations of the covariance propagation equations. Sec.~\ref{sec:apjn-recurrence-derivation-appendix} derives the APJN recurrence relations, and Sec.~\ref{sec:large-l-derivation-appendix} presents derivations of the asymptotic APJN behavior. 
Sec.~\ref{sec:derf-explicit} provides explicit expressions for the covariance components propagated through the $\mathrm{erf}$ nonlinearity that appear in the recurrence relations in the main text. Sec.~\ref{sec-app:final-norm-role} discusses the role of the final normalization layer following the transformer blocks.
Sec.~\ref{sec-app:supplement-sig-prop} provides supplementary material on signal propagation experiments. Finally, Sec.~\ref{sec-app:supplement-stability-exps} presents supplementary training stability experiments.

\section{Experimental Details}
\label{sec-app:exp-details}

\textbf{Resources.} All experiments were conducted in Google Colab. For the APJN computation experiments, we used a single T4 or L4 GPU, taking approximately 48 hours in total. 
The most computationally intensive experiments were the training-stability runs, for which we used a single A100 GPU to train models with up to 24 transformer blocks and a single RTX PRO 6000 GPU (G4 VM) to train the deeper models. Training a ViT model with 12 blocks for 30 epochs takes around 1 hour on an A100 GPU. The experiments included in the paper took approximately 200 GPU-hours in total.

\textbf{ViT model.} As the base model, we use $\texttt{vit\_base\_patch16\_224}$ from $\texttt{timm}$. By default, this model has 12 transformer blocks, hidden dimension $d = 768$, GELU activations in its MLP layers, context size $n = 196$ on CIFAR-100, and weights initialized with standard deviation $0.02$. LayerNorm is applied to the inputs of the residual branches and to the final output. In the Derf models, we replace all of these LayerNorm layers, including the final one, with the Derf nonlinearity parameterized by $\alpha$, following \citet{zhu2025transformersnormalization, chen2025strongernormalizationfreetransformers}.

For the experiments in Section~\ref{sec:signal-prop}, we replace the GELU activations in the MLP layers with ReLU. To match the weight initialization of the ViT model in our theory, we set $\sigma_O=\sigma_V=\sigma_Q=\sigma_K=\sigma_1=0.02\times\sqrt{d}$ and $\sigma_2=0.02\times\sqrt{4d}$ in the theoretical model. Consequently, $\sigma_{OV}\approx 0.31$ and $\sigma_{21}\approx 0.61$.

In experiments where we vary $\sigma_{OV}$, we scale $\sigma_O$ and $\sigma_V$ by the same factor while keeping $\sigma_Q$ and $\sigma_K$ fixed. Analogously, when varying $\sigma_{21}$, we scale $\sigma_1$ and $\sigma_2$ by the same factor.

\textbf{Training stability experiments.} Unless stated otherwise, we train the models using AdamW ($\beta_1 = 0.9$, $\beta_2 = 0.999$) with weight decay $0.05$. We use a linear warmup schedule, after which the learning rate is held constant. We do not use stochastic depth \citep{huang2016deepnetworksstochasticdepth}, in order to properly capture depth-scaling effects.

The training-stability sweeps are single-run experiments for each configuration.
We therefore interpret these sweeps as qualitative evidence for large instability
trends and approximate instability boundaries, rather than as statistically
precise estimates of final accuracy or small performance differences.

The peak learning rate, number of warmup epochs, and batch size used in each training stability experiment in the main text are listed below:
\begin{itemize}
    \item Fig.~\ref{fig:depth_std_tiles}(a): $\mathrm{LR}=3\times 10^{-4},\ \mathrm{warmup}=3,\ \mathrm{batch}=128$;
    \item Fig.~\ref{fig:depth_std_tiles}(b): $\mathrm{LR}=3\times 10^{-4},\ \mathrm{warmup}=3,\ \mathrm{batch}=256$;
    \item Fig.~\ref{fig:warmup_lr_tiles}(a): $\mathrm{LR}=5\times 10^{-4},\ \mathrm{batch}=256$;
    \item Fig.~\ref{fig:warmup_lr_tiles}(b): $\mathrm{warmup}=1,\ \mathrm{batch}=256$;
    \item Fig.~\ref{fig:warmup_lr_tiles}(c): $\mathrm{LR}=3\times 10^{-4},\ \mathrm{warmup}=10,\ \mathrm{batch}=256$.
\end{itemize}

Images from CIFAR-100 were resized to $224 \times 224$. During training, we used the default augmentations from the \texttt{timm} repository: color jitter with magnitude 0.4, RandAugment (\texttt{rand-m9-mstd0.5-inc1}), bicubic interpolation, random erasing with probability 0.25 in pixel mode and count 1, and mixup/cutmix with mixup 0.8, cutmix 1.0, application probability 1.0, switch probability 0.5, and batch mode. We trained using \texttt{SoftTargetCrossEntropy}, with label smoothing set to 0.1.

\textbf{Fig.~\ref{fig:q_pq_apjn}(a). Backward APJN: ViT (synthetic permutation-symmetric inputs) vs. theory.} The APJN values are computed using 8 generated permutation-symmetric inputs with $(q^0, p^0) = (1.0,\ 0.2)$. The components of each input token are drawn from a Gaussian distribution.
For each input, we compute the APJNs across the full set of models: pre-LN and Derf variants with several values of $\alpha$. Note that the variance across inputs is visible only for $(\sigma_{21}, \sigma_{OV}) = (0.6, 1.2)$. The Jacobian Frobenius norms are estimated using Hutchinson's method \citep{hoffman2019robustlearningjacobianregularization, doshi2023criticalinitializationwidedeep} with 10 random draws. These Frobenius norms are then averaged over 5 random weight initializations to estimate the APJNs. 

\textbf{Fig.~\ref{fig:q_pq_apjn}(b). Backward APJN: ViT with CIFAR-100 inputs vs. theory.}
For a randomly selected CIFAR-100 training example, we measure APJNs at every fourth transformer block of a 128-block ViT model, excluding the endpoints. The geometric mean fold error (GMFE), used to compare the measured APJN values with theory, is defined as
\begin{equation}
    \mathrm{GMFE}(q^0,p^0)
    = \exp\left(\frac{1}{|S|}
    \sum_{b\in S}
    \left|\log
    \frac{
    \mathscr{J}_{\mathrm{theory}}^{B,b}(q^0,p^0)
    }{
    \mathscr{J}_{\mathrm{ViT}}^{B,b}
    }
    \right|
    \right),
\end{equation}
where $S$ is a set of blocks in the first third of the network (early), the second third (middle), or the last third (deep). To compute the theoretical values, we estimate $q^0$ and $p^0$ for the given input as
\begin{equation}
    q^0 = \frac{1}{n} \sum_{s=1}^n \frac{\bm{h}^0_s\cdot \bm{h}^0_s}{d}, \quad
    p^0 = \frac{1}{n(n-1)} \sum_{s\ne t} \frac{\bm{h}^0_s\cdot \bm{h}^0_t}{d}.
\end{equation}
We repeat this measurement and comparison for multiple CIFAR-100 samples and multiple values of $(\sigma_{21}, \sigma_{OV})$, as shown in Fig.~\ref{fig:q_pq_apjn}(b). The APJNs are obtained by averaging Jacobian Frobenius norms over 8 independent model-weight initializations. Each Jacobian Frobenius norm is estimated using Hutchinson's method with 10 random matrix draws.

The input statistics used to evaluate the initial condition $(q^0,p^0)$ are extracted immediately before the first transformer block, after the CIFAR-100 image has passed through the initial patch embedding and positional embedding.

\textbf{Fig.~\ref{fig:asymp}. APJNs: ViT (CIFAR-100 inputs) vs. theory vs. asymptotic theory.} For each CIFAR-100 input, we randomly choose 16 blocks at which the APJNs are evaluated. The APJNs are obtained by averaging Jacobian Frobenius norms over 4 model weight initializations. Each Jacobian Frobenius norm is estimated using Hutchinson's method with 10 random matrix draws.

\section{Covariance Propagation}
\label{sec:cov-prop-derivation-appendix}

This section derives Result~\ref{res:covariance-recurrence}.

One of the main ingredients in the signal propagation calculation is covariance propagation through elementwise nonlinearities followed by linear layers. We illustrate this for a transformation of the form $W\phi(\bm{h})$, where the entries of $W$ are i.i.d. $\mathcal{N}(0,\sigma_W^2/d)$. Consider two activation vectors at different positions, $\bm{h}_1$ and $\bm{h}_2$, with component-wise covariance matrix $\Sigma$, such that $\Sigma_{11}=\Sigma_{22}=q$ and $\Sigma_{12}=\Sigma_{21}=p$. Equivalently, for any component index $i$, $(h_{1,i}, h_{2,i}) \sim \mathcal{N}(0, \Sigma)$, and different components are uncorrelated. The covariance after applying the transformation is then $\sigma_W^2 \Sigma^\phi$, where
\begin{equation}
\Sigma^\phi_{ab}
=
\mathbb{E}_{(h_1,h_2)\sim \mathcal{N}(0,\,\Sigma)}
\bigl[\phi(h_a)\,\phi(h_b)\bigr],
\qquad a,b \in \{1,2\}.
\end{equation}
In this expression, each $h_a$ is a scalar random variable; the component index $i$ is suppressed.

In what follows, we derive the recurrence relations given in Eqs.~(\ref{eq:eq-q-recurrence}) and (\ref{eq:eq-p-recurrence}).

\subsection{Covariance Propagation Through MLP}
\label{subsec:cov_prop_mlp_derivation}
Given the MLP-layer activation vector update in Eq. (\ref{eq:forward-pass}),
\begin{equation}
\bm{h}^{l+1}_a
=
\bm{h}^{l}_a
+
W_2^{l}\,\mathrm{ReLU}\!\bigl(W_1^{l}\tilde{\bm{h}}^{l}_{a}\bigr),
\label{eq:mlp}
\end{equation}
we split the computation into three steps.

\textbf{1. Covariance propagation through} $W_1\phi(\cdot)$ \textbf{or} $W_1\mathrm{LayerNorm}(\cdot)$\textbf{.} Here $W_1\in \mathbb{R}^{d_{\text{ff}}\times d}$ has i.i.d. entries $\mathcal{N}(0,\, \sigma_1^2/d)$. Assume the input activations have covariance matrix $\Sigma_{ab}$, $a,b\in\{1,2\}$, with
$\Sigma_{11}=\Sigma_{22}=q$ and $\Sigma_{12}=\Sigma_{21}=p$. The propagated covariance is
\begin{equation}
\Sigma_{ab}'=\frac{1}{d_{\text{ff}}}\mathbb{E}_\theta \left[(W_1\tilde{\bm{h}}_a)\cdot (W_1\tilde{\bm{h}}_b)\right] =\frac{\sigma_1^2}{d}\mathbb{E}_\theta\left[ \tilde{\bm{h}}_a\cdot \tilde{\bm{h}}_b\right].
\end{equation}
For $\mathrm{LayerNorm}$, evaluating the dot product gives $q' = \sigma_1^2$ and $p'=\sigma_1^2\,p/q$. For elementwise functions, we obtain:
\begin{equation}
    \Sigma_{ab}' = \sigma_1^2\,\mathbb{E}_{(h_1,h_2)\sim \mathcal{N}(0, \Sigma)}\left[\phi(h_a)\phi(h_b)\right].
\end{equation}

\textbf{2. Covariance propagation through $W_2\mathrm{ReLU}(\cdot)$.} Here $W_2\in \mathbb{R}^{d\times d_{\text{ff}}}$ has i.i.d. entries $\mathcal{N}(0, \sigma_2^2/d_{\text{ff}})$. Similarly, given an input covariance matrix $\Sigma_{ab}$, the propagated covariance is
\begin{equation}
    {\Sigma}_{ab}' = \sigma_2^2\,\mathbb{E}_{(h_1,h_2)\sim \mathcal{N}(0,\Sigma)}\left[\mathrm{ReLU}(h_a)\mathrm{ReLU}(h_b)\right].
\end{equation}
These Gaussian integrals can be evaluated explicitly, giving:
\begin{equation}
    q'=\frac{1}{2}\sigma_2^2 q, \quad p' = \frac{1}{2}\sigma_2^2\kappa\left(\frac{p}{q}\right)q, 
\end{equation}
where 
\begin{equation}
    \kappa(\rho)=\frac{1}{\pi}\left(\sqrt{1-\rho^2}+\rho(\pi-\arccos\rho)\right).
\label{eq:kappa}
\end{equation}

\textbf{3. Residual addition.} The residual addition adds the covariance contributions from the residual stream and the residual branch, because they are independent.

Combining the results above gives the MLP contributions in Eqs. (\ref{eq:eq-q-recurrence}) and (\ref{eq:eq-p-recurrence}).

\subsection{Covariance Propagation Through Attention}
Given the attention-layer activation update in Eq. (\ref{eq:forward-pass}) 
\begin{equation}
\bm{h}^{l+1}_a
=
\bm{h}^{l}_a
+
W_O^{l} W_V^{l} \displaystyle\sum_{b=1}^n A^{l}_{ab}\,\tilde{\bm{h}}^{l}_{b}
\label{eq:attn}
\end{equation}
we split the computation into two steps.

\textbf{1. Covariance propagation through attention.} Define $\bm{h}'_a=W_O W_V\sum_{b=1}^n A_{ab}\,\tilde{\bm{h}}_{b}$. Assume the input activations have covariance matrix $\Sigma_{ab}$, $a,b=1,\ldots,n$, with $\Sigma_{aa}=q$ and $\Sigma_{ab}=p$ for $a\ne b$. In the uniform-attention approximation \citep{noci2022signalpropagationtransformerstheoretical}, the propagated covariance is
\begin{equation}
    \Sigma_{ab}'=\frac{1}{d} \mathbb{E}_\theta\left[\bm{h}'_a\cdot\bm{h}'_b\right]=\sigma_{OV}^2\sum_{t,s=1}^n\frac{1}{n^2}\mathbb{E}_\theta\left[\frac{\tilde{\bm{h}}_t\cdot\tilde{\bm{h}}_s}{d}\right].
    \label{eq:cov-prop-attn}
\end{equation}
Let $\tilde \Sigma_{ts}=\mathbb{E}_\theta\left[\tilde{\bm{h}}_t\cdot\tilde{\bm{h}}_s/d\right]$, with $t,s=1,\ldots,n$, denote the covariance after propagation through $\phi$, whose entries satisfy $\tilde \Sigma_{tt}=\tilde q$ and $\tilde \Sigma_{ts}=\tilde p$ for $t\ne s$. As in the previous section, it is given by 
\begin{equation}
    \tilde \Sigma_{ts} = \mathbb{E}_{\bm{h}\sim \mathcal{N}(0, \Sigma)}\left[\phi(h_t)\phi(h_s)\right].
\end{equation} 
Therefore,
\begin{equation}
\Sigma_{ab}'= \frac{\sigma_{OV}^2}{n^2}\sum_{t',s'=1}^n \tilde \Sigma_{t's'}=\sigma_{OV}^2\left(\frac{1}{n}\tilde q+\frac{n-1}{n}\tilde p\right). 
\end{equation}
Thus, for large context size $n$, one obtains approximately:
\begin{equation}
    q' \approx \sigma_{OV}^2\tilde p,\quad p'\approx \sigma_{OV}^2\tilde p.
\end{equation}

\textbf{2. Residual addition.} The residual addition adds the covariance contributions from the residual stream and the residual branch, because they are independent:
\begin{equation}
q^{l+1}\approx q^l + \sigma_{OV}^2\tilde{p},\quad p^{l+1}\approx p^l+\sigma_{OV}^2\tilde{p}.
\label{eq:attn-cov-upd}
\end{equation}

\subsection{Covariance Propagation Through Multi-Head Attention}
\label{subsubsec:cov-prop-mha}
In the case of multiple attention heads, Eq. (\ref{eq:attn-cov-upd}) remains unchanged under the standard weight initialization scheme. Define $\bm{h}'_a=\sum_{h=1}^HW_O^h W_V^h\sum_{b=1}^n A_{ab}^h\,\tilde{\bm{h}}_{b}$, where $H$ denotes the total number of attention heads, and the attention scores now depend on the head index $h$ due to separate query and key embeddings:
\begin{equation}
A_{ab}^h
=
\frac{
\exp\!\left((W_Q^h\tilde{\bm{h}}_{a})\cdot(W_K^{h}\tilde{\bm{h}}_{b})/\sqrt{d_{\text{h}}}\right)
}{
\sum_{c=1}^n
\exp\!\left((W_Q^h\tilde{\bm{h}}_{a})\cdot(W_K^h\tilde{\bm{h}}_{c})/\sqrt{d_\text{h}}\right)
}.
\label{eq:attn-scores-mha}
\end{equation}
Here $d_\text{h}=d/H$, and $W_Q^h\in\mathbb{R}^{d_{\text{h}}\times d}$, $W_K^h\in\mathbb{R}^{d_{\text{h}}\times d}$, $W_V^h\in\mathbb{R}^{d_{\text{h}}\times d}$, and $W_O^h\in\mathbb{R}^{d\times d_{\text{h}}}$ have i.i.d. Gaussian entries with zero mean and variances $\sigma_Q^2/d$, $\sigma_K^2/d$, $\sigma_V^2/d$, and $\sigma_O^2/d$, respectively. Note that, in standard Transformer implementations, the variance of $W_O^h\in\mathbb{R}^{d\times d_{\text{h}}}$ scales as $1/d$, which is crucial for this derivation. Let us denote $\vardbtilde{\bm{h}}^h_a=\sum_{b=1}^n A_{ab}^h\,\tilde{\bm{h}}_{b}$. Then, in the uniform-attention approximation, the propagated covariance is
\begin{equation}
\begin{split}
&\Sigma_{ab}'=\frac{1}{d} \mathbb{E}_\theta\left[\bm{h}'_a\cdot\bm{h}'_b\right]=
\sum_{h=1}^H \frac{\sigma_O^2}{Hd_{\text{h}}}\mathbb{E}_\theta\left[ \left(W_V\vardbtilde{\bm{h}}^h_a\right)\cdot
\left(W_V\vardbtilde{\bm{h}}^h_b\right)\right]= 
\frac{1}{H} \sum_{h=1}^H \frac{\sigma_{OV}^2}{d}\mathbb{E}_\theta 
\left[
\vardbtilde{\bm{h}}^h_a\cdot
\vardbtilde{\bm{h}}^h_b
\right]= \\
& = \frac{\sigma_{OV}^2}{H}\sum_{h=1}^H\sum_{t,s=1}^n \frac{1}{n^2}\mathbb{E}_\theta\left[
\frac{\tilde{\bm{h}}_t\cdot \tilde{\bm{h}}_s}{d}\right]= 
\sigma_{OV}^2\sum_{t,s=1}^n \frac{1}{n^2} \mathbb{E}_\theta\left[ 
\frac{\tilde{\bm{h}}_t\cdot \tilde{\bm{h}}_s}{d}\right].
\end{split}
\end{equation}
In the chain of equations above, we performed the weight averages sequentially to keep track of the factors of $d$, $d_\text{h}$, and $H$: first over $W_O$, and then over $W_V$. Thus, we obtain an equation analogous to Eq. (\ref{eq:cov-prop-attn}). This proves that the covariance update for multi-head attention is the same as for single-head attention in Eq. (\ref{eq:attn-cov-upd}), in the uniform-attention approximation.

\section{APJN Recurrence Relations}
This section states and derives the full APJN recurrence relations, including the direct attention contribution and the cross-positional Jacobian-correlation terms; see Results~\ref{res:full-forward-apjn-recurrence} and~\ref{res:full-backward-apjn-recurrence}. The simplified main-text recurrence in Result~\ref{res:apjn-recurrence} follows by neglecting the direct attention contribution and the cross-positional Jacobian-correlation terms.

\label{sec:apjn-recurrence-derivation-appendix}
\subsection{Statement of the APJN Recurrences}
\subsubsection{Forward APJN Recurrence Relations}

\begin{result}[Full forward APJN and cross-positional correlation recurrences]
\label{res:full-forward-apjn-recurrence}
Under Assumption~\ref{ass:signal-prop}, the forward APJN satisfies
\begin{equation}
\mathcal{J}^{l+1,0}
=
\begin{cases}
\left(1+\dfrac{\sigma_{OV}^2}{n}\hat{q}^l\right)\mathcal{J}^{l,0}
+\sigma_{OV}^2\hat{p}^l\mathcal{K}^{l,0},
& \text{if } l \text{ is even (attn)},\\[8pt]
\left(1+\dfrac{1}{2}\sigma_{21}^2\hat q^{\,l}\right)\mathcal{J}^{l,0},
& \text{if } l \text{ is odd (MLP).}
\end{cases}
\label{eq:j-forward-recurrence}
\end{equation}
Here, $\mathcal{K}^{l,0}$ denotes the forward cross-positional Jacobian correlation,
\begin{equation}
\mathcal{K}^{l,0}
=
\frac{1}{n(n-1)d}\,
\mathbb{E}_\theta\bigg[
\sum_{\substack{a\ne f,\ b, \\\mu\nu}}
\frac{\partial h^l_{a\mu}}{\partial h^{0}_{b\nu}}
\frac{\partial h^l_{f\mu}}{\partial h^{0}_{b\nu}}
\bigg].
\label{eq:k-forward-def}
\end{equation}
This quantity satisfies
\begin{equation}
\mathcal{K}^{l+1,0}
=
\begin{cases}
\left(1+\sigma_{OV}^2\hat{p}^l\right)\mathcal{K}^{l,0}
+\dfrac{\sigma_{OV}^2}{n}\hat{q}^l\mathcal{J}^{l,0},
& \text{if } l \text{ is even (attn)},\\[8pt]
\left(1+\sigma_{21}^2\hat{\kappa}^l\hat p^{l}\right)\mathcal{K}^{l,0},
& \text{if } l \text{ is odd (MLP).}
\end{cases}
\label{eq:k-forward-recurrence}
\end{equation}
In Eqs.~(\ref{eq:j-forward-recurrence}) and (\ref{eq:k-forward-recurrence}), $\hat q^l$ and $\hat p^l$ denote the covariance components obtained by propagating the covariance through either the derivative of the nonlinearity $\phi$ or the derivative of LayerNorm. For an elementwise nonlinearity $\phi$,
\begin{equation}
\hat \Sigma_{st}^l
=
\mathbb{E}_{(h_1,h_2)\sim \mathcal{N}(0,\Sigma^l)}
\bigl[\phi'(h_s)\phi'(h_t)\bigr],
\qquad s,t \in \{1,2\},
\label{eq:norm-derivative-cov-prop}
\end{equation}
where $\hat\Sigma^l_{11}=\hat\Sigma^l_{22}=\hat q^l$, $\hat\Sigma^l_{12}=\hat\Sigma^l_{21}=\hat p^l$, and $\Sigma^l_{11}=\Sigma^l_{22}=q^l$, $\Sigma^l_{12}=\Sigma^l_{21}=p^l$. In the LayerNorm case, $\hat q^l=\hat p^l=1/q^l$. The quantity $\hat \kappa^l$ arises from covariance propagation through the derivative of $\mathrm{ReLU}$ and is given by
\begin{equation}
    \hat \kappa^l
    =
    \frac{1}{4}
    +
    \frac{1}{2\pi}
    \arcsin\left(\frac{\tilde p^l}{\tilde q^l}\right).
\label{eq:hat-kappa}
\end{equation}
The recurrence relations in Eqs.~(\ref{eq:j-forward-recurrence}) and (\ref{eq:k-forward-recurrence}) are supplemented with the initial conditions
\begin{equation}
\mathcal{J}^{0,0}=1,
\qquad
\mathcal{K}^{0,0}=0.
\end{equation}
\end{result}

Because $\mathcal{K}^{0,0}=0$, the cross-positional Jacobian correlation can be neglected when $\sigma_{OV}^2/n$ is sufficiently small, producing the simplified APJN recurrence given in Eq.~(\ref{eq:eq-chi-J}) in the main text.

\subsubsection{Backward APJN Recurrence Relations}

\begin{result}[Full backward APJN and cross-positional correlation recurrences]
\label{res:full-backward-apjn-recurrence}
Under Assumption~\ref{ass:signal-prop}, the backward APJN satisfies
\begin{equation}
\mathcal{J}^{L,l}
=
\begin{cases}
\left(1+\dfrac{\sigma_{OV}^2}{n}\hat{q}^l\right)\mathcal{J}^{L,l+1}
+\sigma_{OV}^2\hat{q}^l\mathcal{K}^{L,l+1},
& \text{if } l \text{ is even (attn)},\\[8pt]
\left(1+\dfrac{1}{2}\sigma_{21}^2\hat q^{\,l}\right)\mathcal{J}^{L,l+1},
& \text{if } l \text{ is odd (MLP).}
\end{cases}
\label{eq:j-backward-recurrence}
\end{equation}
Here, $\mathcal{K}^{L,l}$ denotes the backward cross-positional Jacobian correlation,
\begin{equation}
\mathcal{K}^{L,l}
=
\frac{1}{n(n-1)d}\,
\mathbb{E}_\theta\bigg[
\sum_{\substack{b\ne f,\ a, \\\mu\nu}}
\frac{\partial h^L_{a\mu}}{\partial h^{l}_{b\nu}}
\frac{\partial h^L_{a\mu}}{\partial h^{l}_{f\nu}}
\bigg].
\label{eq:k-backward-def}
\end{equation}
This quantity satisfies
\begin{equation}
\mathcal{K}^{L,l}
=
\begin{cases}
\left(1+\sigma_{OV}^2\hat{p}^l\right)\mathcal{K}^{L,l+1}
+\dfrac{\sigma_{OV}^2}{n}\hat{p}^l\mathcal{J}^{L,l+1},
& \text{if } l \text{ is even (attn)},\\[8pt]
\left(1+\sigma_{21}^2\hat{\kappa}^l\hat p^{l}\right)\mathcal{K}^{L,l+1},
& \text{if } l \text{ is odd (MLP).}
\end{cases}
\label{eq:k-backward-recurrence}
\end{equation}
The quantities $\hat q^l$, $\hat p^l$, and $\hat\kappa^l$ are defined as in Result~\ref{res:full-forward-apjn-recurrence}. The recurrence relations in Eqs.~(\ref{eq:j-backward-recurrence}) and (\ref{eq:k-backward-recurrence}) are supplemented by the initial conditions
\begin{equation}
\mathcal{J}^{L,L}=1,
\qquad
\mathcal{K}^{L,L}=0.
\end{equation}
\end{result}

Analogously to the forward APJN case, when $\sigma_{OV}^2/n$ is sufficiently small, the cross-positional Jacobian correlation can be neglected, giving the simplified backward APJN recurrence:
\begin{equation}
\mathcal{J}^{L,l}
\approx
\begin{cases}
\mathcal{J}^{L,l+1},
& \text{if } l \text{ is even (attn)},\\[8pt]
\left(1+\dfrac{1}{2}\sigma_{21}^2\hat q^{\,l}\right)\mathcal{J}^{L,l+1},
& \text{if } l \text{ is odd (MLP).}
\end{cases}
\label{eq:j-backward-recurrence-simple}
\end{equation}
In this case, the backward APJN is approximately the inverse of the forward APJN, up to the overall factor $\mathcal{J}^{L,0}$:
\begin{equation}
\mathcal{J}^{L,l}\approx\mathcal{J}^{L,0}/\mathcal{J}^{l,0}.
\end{equation}

\subsection{Derivation of APJN Recurrence Relations}
\label{subsec:apjn-factors-derivation}
In this section, we derive the APJN recurrence relations in Eqs.~(\ref{eq:j-forward-recurrence}) and (\ref{eq:k-forward-recurrence}). To clearly distinguish positional and feature indices, we use Latin letters for the former and Greek letters for the latter. The $\mu$th component of the activation vector at position $a$ in layer $l$ is denoted by $h_{a\mu}^l$. We begin by separating the contribution of the last layer from the rest in the APJN:
\begin{equation}
\begin{split}
    &\mathcal{J}^{l,0}=\frac{1}{nd}\mathbb{E}_\theta \lVert J^{l,0}\rVert^2_F=\frac{1}{nd}\mathbb{E}_\theta \sum_{ab,\mu\nu}\left(\frac{\partial h_{a\mu}^l}{\partial h^0_{b\nu}}\right)^2=\frac{1}{nd}
    \mathbb{E}_\theta \sum_{\substack{abcd, \\ \mu\nu\xi\rho}}
    \frac{\partial h_{a\mu}^l}{\partial h^{l-1}_{c\xi}} \frac{\partial h_{a\mu}^l}{\partial h^{l-1}_{d\rho}}
    \frac{\partial h_{c\xi}^{l-1}}{\partial h^{0}_{b\nu}} \frac{\partial h_{d\rho}^{l-1}}{\partial h^{0}_{b\nu}} = \\
    & =\frac{1}{nd}\sum_{\substack{abcd, \\ \mu\nu\xi\rho}}\mathbb{E}_\theta
    \left[\frac{\partial h_{a\mu}^l}{\partial h^{l-1}_{c\xi}} \frac{\partial h_{a\mu}^l}{\partial h^{l-1}_{d\rho}}\right]
    \mathbb{E}_\theta
    \left[\frac{\partial h_{c\xi}^{l-1}}{\partial h^{0}_{b\nu}} \frac{\partial h_{d\rho}^{l-1}}{\partial h^{0}_{b\nu}}\right].
\end{split}
\label{eq:apjn-starting-point}
\end{equation}
In the last equality, we used the infinite-width factorization property: although the local Jacobian 
$\partial h^l/\partial h^{l-1}$ depends on $h^{l-1}$, the conditional second moment of its entries self-averages to a deterministic function of the layer-$(l-1)$ activation order parameters, such as the self-overlap $q^{l-1}$ and the cross-positional overlap $p^{l-1}$. Analogously, one can separate the contribution of layer $l$ from the rest in the cross-positional Jacobian correlation $\mathcal{K}^{l,0}$:
\begin{equation}
\begin{split}
    &\mathcal{K}^{l,0}=\frac{1}{n^2d}\mathbb{E}_\theta \sum_{\substack{a\ne f,\ b, \\\mu\nu}}
    \frac{\partial h_{a\mu}^l}{\partial h^0_{b\nu}}\frac{\partial h_{f\mu}^l}{\partial h^0_{b\nu}}
    =\frac{1}{n^2d}\sum_{\substack{a\ne f,\ bcd, \\ \mu\nu\xi\rho}}\mathbb{E}_\theta
    \left[\frac{\partial h_{a\mu}^l}{\partial h^{l-1}_{c\xi}} \frac{\partial h_{f\mu}^l}{\partial h^{l-1}_{d\rho}}\right]
    \mathbb{E}_\theta
    \left[\frac{\partial h_{c\xi}^{l-1}}{\partial h^{0}_{b\nu}} \frac{\partial h_{d\rho}^{l-1}}{\partial h^{0}_{b\nu}}\right].
\end{split}
\label{eq:k-starting-point}
\end{equation}
Here and in what follows, we approximate $n(n-1)\approx n^2$, assuming that the context size $n$ is sufficiently large. Our goal is now to compute the first expectation in Eqs.~(\ref{eq:apjn-starting-point}) and (\ref{eq:k-starting-point}), which will allow us to derive the recurrence relations in Eqs.~(\ref{eq:j-forward-recurrence}) and (\ref{eq:k-forward-recurrence}).

\subsubsection{$\mathcal{J}^{l,0}$ Recurrence for MLP}
Consider the activation vector update in Eq. (\ref{eq:mlp}) due to the MLP. Let $\psi$ denote a vector-valued function that is either an elementwise nonlinearity or $\mathrm{LayerNorm}$, i.e., $\psi \in \{\phi,\, \mathrm{LayerNorm}\}$. Since MLP layers do not mix activation vectors across positions, the APJN is a sum of identical contributions from each position. It therefore suffices to consider a single positional index:
\begin{equation}
    \frac{\partial h_{\mu}^{l+1}}{\partial h^{l}_{\xi}}=\delta_{\mu\xi}+\sum_{\alpha\beta} (W_2)^l_{\mu\alpha}\mathrm{ReLU}'\left(\big(W_1\tilde{\bm{h}}^l\big)_\alpha\right)\left(W_1\right)^l_{\alpha\beta} 
    \frac{\partial \psi(\bm{h}^l)_\beta}{\partial h^l_\xi},
    \label{eq:mlp-jac-term}
\end{equation}
where, for an elementwise nonlinearity $\phi$,
\begin{equation}
    \frac{\partial \psi(\bm{h}^l)_\beta}{\partial h^l_\xi}=\phi'(h_\beta^l)\delta_{\beta\xi},
\end{equation}
whereas, for $\mathrm{LayerNorm}$, 
\begin{equation}
    \frac{\partial \psi(\bm{h}^l)_\beta}{\partial h^l_\xi}=\frac{\sqrt{d}}{\lVert\bm{h}^l\rVert}\left(\delta_{\beta\xi}-\frac{h^l_\beta h^l_\xi}{\lVert\bm{h}^l\rVert^2}\right).
    \label{eq:ln-jac}
\end{equation}
The second term in this expression is negligible in the large-$d$ limit. Taking the expectation of a product of terms (\ref{eq:mlp-jac-term}), as required for Eq. (\ref{eq:apjn-starting-point}), 
\begin{equation}
    \mathbb{E}_\theta
    \left[\frac{\partial h_{\mu}^{l+1}}{\partial h^{l}_{\xi}} \frac{\partial h_{\mu}^{l+1}}{\partial h^{l}_{\rho}}\right] = \delta_{\mu\xi}\delta_{\mu\rho}+\delta_{\xi\rho} \frac{\sigma_{21}^2}{d}\left(\frac{1}{d_{\text{ff}}}\mathrm{ReLU}'\!\left(W_1^l\tilde{\bm{h}}^l\right)\cdot \mathrm{ReLU}'\!\left(W_1^l\tilde{\bm{h}}^l\right)\right)\mathbb{E}_\theta\!\left[\phi'\left(h^l_\rho\right)^2\right],
\label{eq:mlp-expectation}
\end{equation}
where the prime denotes the derivative. The dot product in parentheses is the activation variance propagated through $\mathrm{ReLU}'$ (i.e., the Heaviside step function), which does not depend on the input and equals $1/2$. Moreover, the expectation $\mathbb{E}_\theta\!\left[\phi'\left(h^l_\rho\right)^2\right]$ does not depend on the index $\rho$. For $\mathrm{LayerNorm}$, this expectation should effectively be replaced by $d/\lVert\bm{h}^l\rVert^2=1/q^l$, as follows from Eq. (\ref{eq:ln-jac}) in the large-$d$ limit. Substituting the above expression into Eq. (\ref{eq:apjn-starting-point}) and restoring the positional indices, we obtain
\begin{equation}
\mathcal{J}^{l+1,0}=\frac{1}{nd}
\sum_{ab,\rho\nu}
\left(1+\frac{1}{2}\sigma_{21}^2\mathbb{E}_\theta\left[\phi'\!\left(h^{l}\right)^2\right]\right)
\mathbb{E}_\theta
\left[
\left(
\frac{\partial h^l_{a\rho}}{\partial h^0_{b\nu}}
\right)^2
\right]
\end{equation}
Thus, for the MLP, we have shown that $\mathcal{J}^{l+1,0} = \chi^l_\mathcal{J} \mathcal{J}^{l,0}$, where 
\begin{equation}
\chi^l_\mathcal{J}=1+\frac{\sigma_{21}^2}{2}\hat q^l, 
\label{eq:chi-J-appendix}
\end{equation}
and, for elementwise nonlinearities, we define 
\begin{equation}
\hat q^l = \mathbb{E}_{h \sim \mathcal{N}(0,\,q^l)} \left[\phi'(h)^2\right].
\label{eq:hat-q-def}
\end{equation}
For $\mathrm{LayerNorm}$, we set $\hat q^l = 1/q^l$, following the discussion above.

\subsubsection{$\mathcal{K}^{l,0}$ Recurrence for MLP}
Restoring the positional indices in Eq.~(\ref{eq:mlp-expectation}), we obtain
\begin{equation}
\begin{split}
    &\mathbb{E}_\theta
    \left[\frac{\partial h_{a\mu}^{l+1}}{\partial h^{l}_{c\xi}} \frac{\partial h_{f\mu}^{l+1}}{\partial h^{l}_{d\rho}}\right] = \\
&=\delta_{ac}\delta_{fd}\left[\delta_{\mu\xi}\delta_{\mu\rho}+
    \delta_{\xi\rho} \frac{\sigma_{21}^2}{d}\left(\frac{1}{d_{\text{ff}}}\mathrm{ReLU}'\!\left(W_1^l\tilde{\bm{h}}^l_c\right)\cdot\mathrm{ReLU}'\!\left(W_1^l\tilde{\bm{h}}^l_d\right)\right)\mathbb{E}_\theta\!\left[\phi'\left(h^l_{c\rho}\right)\phi'\left(h^l_{d\rho}\right)\right]\right].
\end{split}
\label{eq:mlp-expectation-for-k}
\end{equation}

Substituting this expression into Eq.~(\ref{eq:k-starting-point}), we find that only the terms with $c\ne d$ contribute to the sum, which yields the recurrence relation
\begin{equation}
\mathcal{K}^{l+1,0}=\left(1+\sigma_{21}^2 \hat{\kappa}^l \hat p^l\right)\mathcal{K}^{l,0},
\end{equation}
where $\hat p^l$ is the off-diagonal counterpart of $\hat q^l$ in Eq.~(\ref{eq:hat-q-def}):
\begin{equation}
    \hat{p}^l = \mathbb{E}_{(h_1,\,h_2)\sim \mathcal{N}(0,\, \Sigma^l)}\left[\phi'(h_1)\phi'(h_2)\right],
\end{equation}
and $\hat \kappa^l$ arises from covariance propagation through the derivative of $\mathrm{ReLU}$:
\begin{equation}
    \hat \kappa^l = \mathbb{E}_{(h_1,\,h_2)\sim \mathcal{N}(0,\, \tilde\Sigma^l)}\left[\mathrm{ReLU}'(h_1)\mathrm{ReLU}'(h_2)\right]=\frac{1}{4}+\frac{1}{2\pi}\arcsin\left(\frac{\tilde p^l}{\tilde q^l}\right).
\end{equation}
Here, $\Sigma^l$ is the covariance matrix of the activations $\bm{h}^l$, whereas $\tilde\Sigma^l$ is the covariance matrix of the normalized activations $\tilde{\bm{h}}^l$. Note that for LayerNorm, the expectation $\mathbb{E}_\theta\!\big[\phi'\left(h^l_{c\rho}\right)\phi'\left(h^l_{d\rho}\right)\big]$ is replaced by $d/\lVert\bm{h}^l\rVert^2 = 1/q^l$, analogously to the derivation in the previous subsection. Therefore, for LayerNorm, we set $\hat p^l = 1/q^l$.

\subsubsection{$\mathcal{J}^{l,0}$ and $\mathcal{K}^{l,0}$ Recurrence for Attention}
\label{subsubsec:j_k_recurrence_attn}
Consider the activation vector update in Eq.~(\ref{eq:attn}) due to attention. As argued in \citet{cowsik2024geometricdynamicssignalpropagation}, the terms arising from differentiating the attention scores are subleading in $1/d$. Therefore, the computation is similar to the MLP case: we differentiate the nonlinearity and additionally approximate the product of attention scores as $1/n^2$. This gives
\begin{equation}
\mathbb{E}_\theta
    \left[\frac{\partial h_{a\mu}^{l+1}}{\partial h^{l}_{c\xi}} \frac{\partial h_{f\nu}^{l+1}}{\partial h^{l}_{d\rho}}\right]=\delta_{ac}\delta_{fd}\delta_{\mu\xi}\delta_{\nu\rho}+\frac{\sigma_{OV}^2}{d}\delta_{\mu\nu}\delta_{\xi\rho} A_{ac}A_{fd}\mathbb{E}_\theta\left[\phi'(h^\rho_c)\phi'(h^\rho_d)\right].
\label{eq:attn-derivatives}
\end{equation}
Under the uniform-attention approximation, the second term does not depend on the indices $a$ and $f$ and therefore contributes equally to the recurrences for $\mathcal{J}^{l,0}$ and $\mathcal{K}^{l,0}$, as can be seen from Eqs.~(\ref{eq:apjn-starting-point}) and (\ref{eq:k-starting-point}). Substituting this expression into Eq.~(\ref{eq:apjn-starting-point}) and splitting the sum into the contributions with $c=d$ and $c\ne d$, we obtain
\begin{equation}
    \mathcal{J}^{l+1, 0} = \mathcal{J}^{l, 0} + \frac{\sigma_{OV}^2}{n}\hat{q}^l\frac{1}{nd}\sum_{c\xi,\,b\nu}\left(\frac{\partial h^l_{c\xi}}{\partial h^0_{b\nu}}\right)^2+\sigma_{OV}^2\hat p^l \frac{1}{n^2d}\sum_{\substack{c\ne d,\, b, \\ \rho\nu}} 
    \frac{\partial h_{c\rho}^l}{\partial h^0_{b\nu}}\frac{\partial h_{d\rho}^l}{\partial h^0_{b\nu}}.
\end{equation}
Here, we recover the APJN recurrence in Eq.~(\ref{eq:j-forward-recurrence}), as claimed. Analogously, substituting Eq.~(\ref{eq:attn-derivatives}) into Eq.~(\ref{eq:k-starting-point}) and again splitting the sum into the contributions with $c=d$ and $c\ne d$, we obtain
\begin{equation}
    \mathcal{K}^{l+1,0} = \mathcal{K}^{l,0}+\frac{\sigma_{OV}^2}{n} \hat{q}^l \frac{1}{nd} \sum_{c\xi,\,b\nu} \left(\frac{\partial h^l_{c\xi}}{\partial h^0_{b\nu}}\right)^2 + \sigma_{OV}^2 \hat p^l \frac{1}{n^2d}\sum_{\substack{c\ne d,\, b, \\ \rho\nu}} 
    \frac{\partial h_{c\rho}^l}{\partial h^0_{b\nu}}\frac{\partial h_{d\rho}^l}{\partial h^0_{b\nu}}.
\end{equation}
This proves the recurrence for the cross-positional Jacobian correlation in Eq.~(\ref{eq:k-forward-recurrence}).

\textbf{Multi-head attention.} In the case of multi-head attention, under the same assumptions as in Sec.~\ref{subsubsec:cov-prop-mha}, Eq.~(\ref{eq:attn-derivatives}) is modified as follows:
\begin{equation}
\mathbb{E}_\theta
    \left[\frac{\partial h_{a\mu}^{l+1}}{\partial h^{l}_{c\xi}} \frac{\partial h_{f\nu}^{l+1}}{\partial h^{l}_{d\rho}}\right]=\delta_{ac}\delta_{fd}\delta_{\mu\xi}\delta_{\nu\rho}+\frac{1}{H}\sum_{h=1}^H\frac{\sigma_{OV}^2}{d}\delta_{\mu\nu}\delta_{\xi\rho}A^h_{ac}A^h_{fd}\mathbb{E}_\theta\left[\phi'(h^\rho_c)\phi'(h^\rho_d)\right].
\label{eq:attn-derivatives-mha}
\end{equation}
Approximating the product of the attention scores by $1/n^2$, this equation reduces to Eq.~(\ref{eq:attn-derivatives}), leading to the same APJN recurrence relations. 

\subsubsection{Derivation of Backward APJN Recurrence Relations}
The derivation of the recurrence relations for the backward quantities, $\mathcal{J}^{L,l}$ and $\mathcal{K}^{L,l}$, in Eqs.~(\ref{eq:j-backward-recurrence}) and (\ref{eq:k-backward-recurrence}) closely parallels the derivation given above for the forward quantities. We begin by separating the contribution of layer $l+1$ from the rest in the backward APJN:
\begin{equation}
\begin{split}
    &\mathcal{J}^{L,l}=\frac{1}{nd}\mathbb{E}_\theta \sum_{\substack{ab, \\\mu\nu}}\left(
    \frac{\partial h_{a\mu}^L}{\partial h^l_{b\nu}}\right)^2
    =\frac{1}{nd}\sum_{\substack{abcd, \\ \mu\nu\xi\rho}}\mathbb{E}_\theta
    \left[\frac{\partial h_{a\mu}^L}{\partial h^{l+1}_{c\xi}} \frac{\partial h_{a\mu}^L}{\partial h^{l+1}_{d\rho}}\right]
    \mathbb{E}_\theta
    \left[\frac{\partial h_{c\xi}^{l+1}}{\partial h^{l}_{b\nu}} \frac{\partial h_{d\rho}^{l+1}}{\partial h^{l}_{b\nu}}\right].
\end{split}
\label{eq:j-backward-starting-point}
\end{equation}
Analogously, we separate the contribution of layer $l+1$ from the rest in the cross-positional Jacobian correlation $\mathcal{K}^{L,l}$:
\begin{equation}
\begin{split}
    &\mathcal{K}^{L,l}=\frac{1}{n^2d}\mathbb{E}_\theta \sum_{\substack{b\ne f,\ a, \\\mu\nu}}
    \frac{\partial h_{a\mu}^L}{\partial h^l_{b\nu}}\frac{\partial h_{a\mu}^L}{\partial h^l_{f\nu}}
    =\frac{1}{n^2d}\sum_{\substack{b\ne f,\ acd, \\ \mu\nu\xi\rho}}\mathbb{E}_\theta
    \left[\frac{\partial h_{a\mu}^L}{\partial h^{l+1}_{c\xi}} \frac{\partial h_{a\mu}^L}{\partial h^{l+1}_{d\rho}}\right]
    \mathbb{E}_\theta
    \left[\frac{\partial h_{c\xi}^{l+1}}{\partial h^{l}_{b\nu}} \frac{\partial h_{d\rho}^{l+1}}{\partial h^{l}_{f\nu}}\right].
\end{split}
\label{eq:k-backward-starting-point}
\end{equation}
Since the rest of the derivation is identical to that given above, we omit it here.

\section{Asymptotic Signal Propagation}
\label{sec:large-l-derivation-appendix}

From the covariance propagation recurrence relations, Eqs.~(\ref{eq:eq-q-recurrence}) and (\ref{eq:eq-p-recurrence}), it follows that $q^l$ and $p^l$ increase with $l$. As a result, for $\mathrm{LayerNorm}$ or $\tanh$-like nonlinearities, $\tilde q^l$ and $\tilde p^l$ saturate, which in turn leads to asymptotically linear growth of both $q^l$ and $p^l$ at large $l$. Therefore, the output of the $b$th transformer block has the following covariance components:
\begin{equation}
\begin{split}
    & q^{2b} \sim b\left(\frac{1}{2}\sigma_{21}^2\tilde q_\star + \sigma_{OV}^2\tilde p_\star\right), \\
    &p^{2b} \sim b\left(\frac{1}{2}\sigma_{21}^2\kappa\left(\frac{\tilde p_\star}{\tilde q_\star}\right)\tilde q_\star + \sigma_{OV}^2\tilde p_\star\right),
\end{split}
\label{eq:q-p-linear-growth-appendix}
\end{equation}
where $\tilde q_\star$ and $\tilde p_\star$ are the limiting values of $\tilde q^{2b}$ and $\tilde p^{2b}$ as $b\to\infty$, respectively. Assuming that $\sigma_{OV}^2/n$ is sufficiently small, so that the cross-positional Jacobian correlation terms $\mathcal{K}^{l,0}$ and $\mathcal{K}^{L,l}$ are negligible, only the odd (MLP) layers contribute to the APJN:
\begin{equation}
\mathcal{J}^{2b,0}=\prod_{k=0}^{b-1}\chi^{2k+1}_\mathcal{J}=\prod_{k=0}^{b-1} \left(1+\frac{\sigma_{21}^2}{2}\hat q^{2k+1}\right).
\label{eq:apjn-prod-simplified}
\end{equation}

\subsection{Asymptotic Behavior of APJN}

This section derives Results~\ref{res:pre-ln-apjn-asymptotics} and~\ref{res:tanh-like-apjn-asymptotics}.

\subsubsection{LayerNorm}
From Eq. (\ref{eq:apjn-prod-simplified}) and using the asymptotic expression for $q^{2k}$ in Eq. (\ref{eq:q-p-linear-growth-appendix}), we obtain:
\begin{equation}
\chi^{2k+1}_\mathcal{J}=1+\frac{\sigma_{21}^2}{2}\frac{1}{q^{2k+1}}\sim1+\frac{\frac{1}{2}\sigma_{21}^2}{k\left(\frac{1}{2}\sigma_{21}^2+\sigma_{OV}^2\tilde p_\star\right)}=1+\frac{\zeta}{k},
\end{equation}
where $\zeta$ is defined in Eq. (\ref{eq:zeta}) in the main text. Thus, for large transformer block index $b$,
\begin{equation}
    \log \mathcal{J}^{2b, 0}\sim \sum_{k=K}^b \log\left(1+\frac{\zeta}{k}\right)=\sum_{k=K}^b\left[ \frac{\zeta}{k}+O(k^{-2})\right]\sim \zeta \log b,
\end{equation}
where $K$ is a sufficiently large integer whose choice does not affect the asymptotics. Therefore, the APJN exhibits power-law growth:
\begin{equation}
    \mathcal{J}^{2b,0}\sim b^{\zeta}.
\end{equation}

\subsubsection{$\tanh$-like nonlinearities}
\label{subsubsec-app:tanh-asymp-apjn}
First, for $\tanh$-like nonlinearities, Eq.~(\ref{eq:q-hat}) implies the following large-$q^l$ asymptotic form for $\hat q^l$:
\begin{equation}
    \hat q^l\approx \frac{C}{\sqrt{q^l}},
    \qquad
    C=\frac{1}{\sqrt{2\pi}}\int_\mathbb{R}\phi'(h)^2\,dh.
\end{equation}
Next, proceeding as in the previous subsection, we obtain the following asymptotic expression for $\chi^{2k+1}_\mathcal{J}$:
\begin{equation}
\chi^{2k+1}_\mathcal{J}=1+\frac{\sigma_{21}^2}{2}\frac{C}{\sqrt{q^{2k+1}}}\sim1+\frac{\frac{1}{2}\sigma_{21}^2C}{\sqrt{k\left(\frac{1}{2}\sigma_{21}^2+\sigma_{OV}^2\tilde p_\star\right)}}=1+\frac{1}{2\sqrt{k\lambda}},
\end{equation}
where $\lambda$ is defined in Eq. (\ref{eq:lambda}) in the main text. Thus, for large transformer block index $b$, 
\begin{equation}
    \log \mathcal{J}^{2b, 0}\sim \sum_{k=K}^b \log\left(1+\frac{1}{2\sqrt{k\lambda}}\right)=\sum_{k=K}^b\left[ \frac{1}{2\sqrt{k\lambda}}-\frac{1}{8k\lambda}+O(k^{-3/2})\right]\sim \sqrt{\frac{b}{\lambda}}-\frac{1}{8\lambda}\log b,
\end{equation}
where $K$ is a sufficiently large integer. Therefore, the APJN grows stretched-exponentially with $b$, with a power-law prefactor:
\begin{equation}
    \mathcal{J}^{2b,0}\sim b^{-\frac{1}{8\lambda}}e^{\sqrt{b/\lambda}}.
\end{equation}
This power-law prefactor modifies the asymptotic behavior of the backward APJN in Eq.~(\ref{eq:derf-asymptotic-apjn}) to
\begin{equation}
    \mathscr{J}^{B,\,b}\sim \left(\frac{B}{b}\right)^{-\frac{1}{8\lambda}}e^{\left(\sqrt{B}-\sqrt{b}\right)/\sqrt{\lambda}}.
\label{eq:derf-asymp-back-apjn-prefactor}
\end{equation}

\subsection{Limiting Value of $c=p/q$}

This section derives the equation for the limiting value of the inter-token cosine similarity $c=p/q$ used in Result~\ref{res:asymptotic-covariance-growth}, analyzes the rate at which $c$ approaches this limiting value, and shows that $c_\star=1$ is unstable for $\tanh$-like nonlinearities.

\subsubsection{Equation for the Limiting Value of $c=p/q$}
\label{subsubsec:fixed-point-derivation}

Let us denote by $c^b=p^{2b}/q^{2b}$ the ratio of the covariance components, which is equal to the cosine similarity between activation vectors at different positions in the output of transformer block $b$. For large $b$, the increments of $q^{2b}$ and $p^{2b}$ per transformer block are approximately
\begin{equation}
\begin{split}
    &\Delta q(c) = \frac{1}{2}\sigma_{21}^2+\sigma_{OV}^2 \tilde p(c), \\
    &\Delta p(c) = \frac{1}{2}\sigma_{21}^2\kappa(\tilde p(c))+\sigma_{OV}^2 \tilde p (c),
\end{split}
\label{eq:p-q-increments}
\end{equation}
evaluated at $c=c^b$. Here, $\tilde p(c)$ denotes the asymptotic form of $\tilde p$ for large $q$ and $p$, as follows from Eq. (\ref{eq:norm-cov-prop}):
\begin{equation}
\tilde p(c)=\frac{2}{\pi}\arcsin c
\label{eq:tilde-p}
\end{equation}
for an elementwise $\tanh$-like nonlinearity, and 
\begin{equation}
\tilde p(c)=c
\end{equation}
for $\mathrm{LayerNorm}$. Therefore, the asymptotic dynamics of $c^b$, induced by those of $p^{2b}$ and $q^{2b}$, are given by 
\begin{equation}
    c^{b+1}= c^b + \frac{\Delta p(c^b)}{q^{2b}}-c^b\frac{\Delta q(c^b)}{q^{2b}}+o\left(\frac{1}{q^{2b}}\right).
    \label{eq:cos-eq}
\end{equation}
For convenience, let us define
\begin{equation}
g(c)=\Delta p(c)-c\Delta q(c).
\label{eq:gc}
\end{equation}
The fixed points of the asymptotic dynamics are given by the zeros of $g(c)$. Thus, any fixed point $c_\star$ satisfies $g(c_\star)=0$, or equivalently, $c_\star=\Delta p(c_\star)/\Delta q(c_\star)$, i.e.,
\begin{equation}
    c_\star = \frac{\frac{1}{2}\sigma_{21}^2\kappa(\tilde p(c_\star))+\sigma_{OV}^2 \tilde p (c_\star)}{\frac{1}{2}\sigma_{21}^2+\sigma_{OV}^2 \tilde p(c_\star)}.
\end{equation}

\subsubsection{Rate of Convergence of $c$ to the Limiting Value}
At a fixed point, $g(c_\star)=0$, so we expand $g(c)$ in Eq. (\ref{eq:cos-eq}) to first order in $c-c_\star$. This yields
\begin{equation}
\Delta c \sim \frac{g'(c_\star)(c-c_\star)}{b\Delta q(c_\star)}.    
\end{equation}
From this equation, we obtain power-law convergence to the fixed point,
\begin{equation}
    |c-c_\star| = b^{-\mu},
\end{equation}
with exponent
\begin{equation}
    \mu = - \frac{g'(c_\star)}{\Delta q(c_\star)}.
\end{equation}
The derivative $g'(c)$ follows from the definition of $g(c)$: 
\begin{equation}
g'(c)= \tilde p'(c)\left[\frac{1}{2}\sigma_{21}^2\kappa'(\tilde p(c))+(1-c)\sigma_{OV}^2\right]-\frac{1}{2}\sigma_{21}^2-\sigma_{OV}^2\tilde p(c).
\end{equation}
For $\mathrm{LayerNorm}$, the exponent simplifies to $\mu = \sigma_{OV}^2/\left(\sigma_{OV}^2+\frac{1}{2}\sigma_{21}^2\right)$.

\subsubsection{
Instability of the Asymptotic Fixed Point $c_\star = 1$ for $\tanh$-Like Nonlinearities}

From Eq. (\ref{eq:cos-eq}), a transformer block increases the alignment between activation vectors, as measured by $c$, if $g(c)>0$, and decreases it if $g(c)<0$. Let us show that for $c=1-\varepsilon$ with sufficiently small $\varepsilon$, $g(c)$ is negative for $\tanh$-like nonlinearities. For $\tanh$-like nonlinearities, from Eq. (\ref{eq:tilde-p}), $\tilde p(1-\varepsilon)=1-\frac{2\sqrt{2}}{\pi}\sqrt{\varepsilon}+O(\varepsilon^{3/2})$. Additionally, from Eq. (\ref{eq:kappa}), for small $x$, $\kappa(1-x)=1-x+o(x)$. Thus, from Eqs. (\ref{eq:gc}) and (\ref{eq:p-q-increments}), a straightforward computation yields:
\begin{equation}
g(1-\varepsilon)= - \frac{\sqrt{2}}{\pi}\sigma_{21}^2\sqrt{\varepsilon}+o(\sqrt{\varepsilon}),
\end{equation}
which is negative for sufficiently small $\varepsilon$. This implies that $c$ decreases according to Eq. (\ref{eq:cos-eq}).

\section{Explicit Expressions for $\mathrm{erf}$-Nonlinearity}
\label{sec:derf-explicit}
For $\phi(h)=\mathrm{erf}(\alpha h)$, the covariance components propagated through $\phi$, denoted by $\tilde q^l$ and $\tilde p^l$ and defined in Eq.~(\ref{eq:norm-cov-prop}), can be computed analytically as
\begin{equation}
    \tilde q^l = \frac{2}{\pi}\arcsin\left(\frac{2\alpha^2 q^l}{1+2\alpha^2 q^l}\right),\qquad
    \tilde p^l = \frac{2}{\pi}\arcsin\left(\frac{2\alpha^2 p^l}{1+2\alpha^2 q^l}\right).
\end{equation}
Similarly, the covariance components propagated through $\phi'$, denoted by $\hat q^l$ and $\hat p^l$ and defined in Eq.~(\ref{eq:norm-derivative-cov-prop}), are given by
\begin{equation}
    \hat q^l = \frac{4\alpha^2}{\pi\sqrt{1+4\alpha^2 q^l}},\qquad
    \hat p^l = \frac{4\alpha^2}{\pi\sqrt{(1+2\alpha^2 q^l)^2 - 4\alpha^4 (p^l)^2}}.
\end{equation}
In particular, these expressions explicitly show why smaller values of $\alpha$ in Derf lead to slower growth of both $q^l$, according to Eqs.~(\ref{eq:eq-q-recurrence}) and (\ref{eq:eq-p-recurrence}), and the APJN, according to Eqs.~(\ref{eq:apjn-recurrence}) and (\ref{eq:eq-chi-J}).

\section{Role of the Final Normalization Layer}
\label{sec-app:final-norm-role}

Pre-LN transformers typically use a final LayerNorm, whereas DyT/Derf models use the corresponding $\tanh$-like nonlinearity at the output \citep{zhu2025transformersnormalization, chen2025strongernormalizationfreetransformers}. This final normalization layer contributes an additional factor $\hat Q^B$ to the backward APJN: specifically, the APJN from block $b$ to the output of the final normalization layer satisfies \begin{equation}
\mathscr{J}^{\text{out},b}=\hat{Q}^B\mathscr{J}^{B,b}.
\end{equation}
For LayerNorm,
\begin{equation}
    \hat{Q}^B=1/Q^B,
\end{equation}
whereas for a final $\tanh$-like nonlinearity parameterized by $\alpha$, $\hat{Q}^B$ has the same form as in Eq.~(\ref{eq:hat-q-def}):
\begin{equation}
\hat{Q}^B=\mathbb{E}_{h\sim\mathcal{N}(0,\,Q^B)} \left[\phi'(h)^2\right],
\end{equation}
In particular, for a final $\mathrm{erf}$ nonlinearity parameterized by $\alpha$,
\begin{equation}
\hat{Q}^B=\frac{4\alpha^2}{\pi\sqrt{1+4\alpha^2Q^B}}.
\label{eq:final-derf-role}
\end{equation}
This factor is governed by the same mechanism as the layer-to-layer APJN growth discussed in the main text. At fixed $\alpha$ in DyT/Derf, increasing model depth or weight scale increases the last-block variance $Q^B$; once $Q^B$ is sufficiently large, $\hat Q^B$ is larger for DyT/Derf than for LayerNorm. At fixed model depth, however, decreasing $\alpha$ suppresses $\hat Q^B$ and can make it smaller than in pre-LN. Thus, replacing the final LayerNorm with DyT/Derf can further increase backward gradient amplification in deep networks, though sufficiently small $\alpha$ can weaken or reverse this effect.

\section{Supplementary Results on Signal Propagation}
\label{sec-app:supplement-sig-prop}

\begin{figure}[tbp]
  \centering
\includegraphics[width=14.0cm]{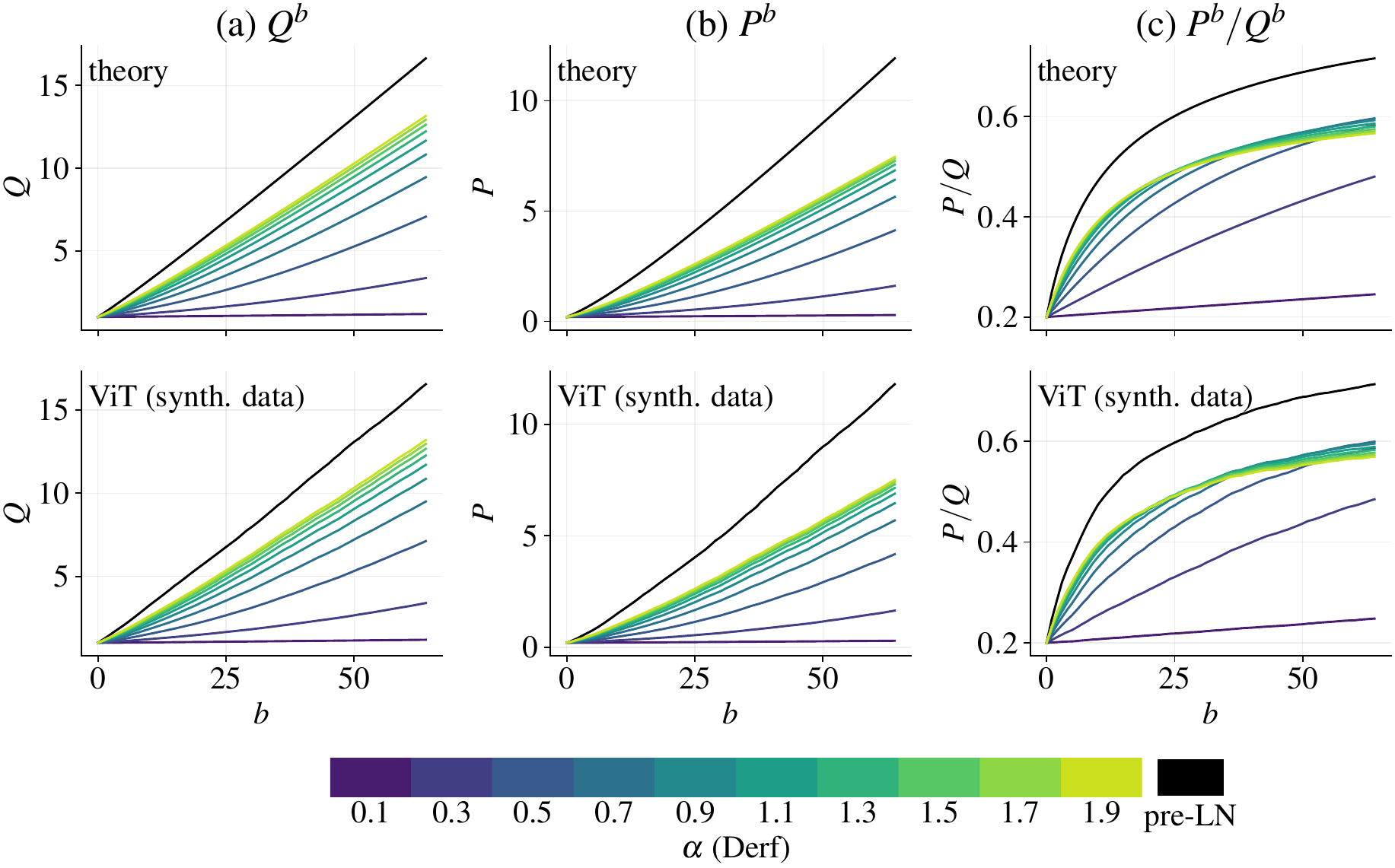}
\caption{
The activation covariance components \textbf{(a)} $Q^b$ and \textbf{(b)} $P^b$, as well as their ratio \textbf{(c)} $P^b/Q^b$. \textbf{Upper row}: theoretical values predicted by the recurrence relations (\ref{eq:eq-q-recurrence}) and (\ref{eq:eq-p-recurrence}); \textbf{lower row}: values measured in a ViT model for a permutation-symmetric input token configuration with $(q^0,p^0)=(1.0,\ 0.2)$.
}
\label{fig:equiang_p_q_pq}
\end{figure}

\textbf{Activation covariance components $Q^b$ and $P^b$, and their ratio $P^b/Q^b$.} Fig.~\ref{fig:equiang_p_q_pq} shows these quantities as computed by the theory and measured in a ViT model on a permutation-symmetric input token configuration with $(q^0,\, p^0)=(1.0,\, 0.2)$, averaged over 16 weight initializations.

\begin{figure}[tbp]
  \centering
\includegraphics[width=14.0cm]{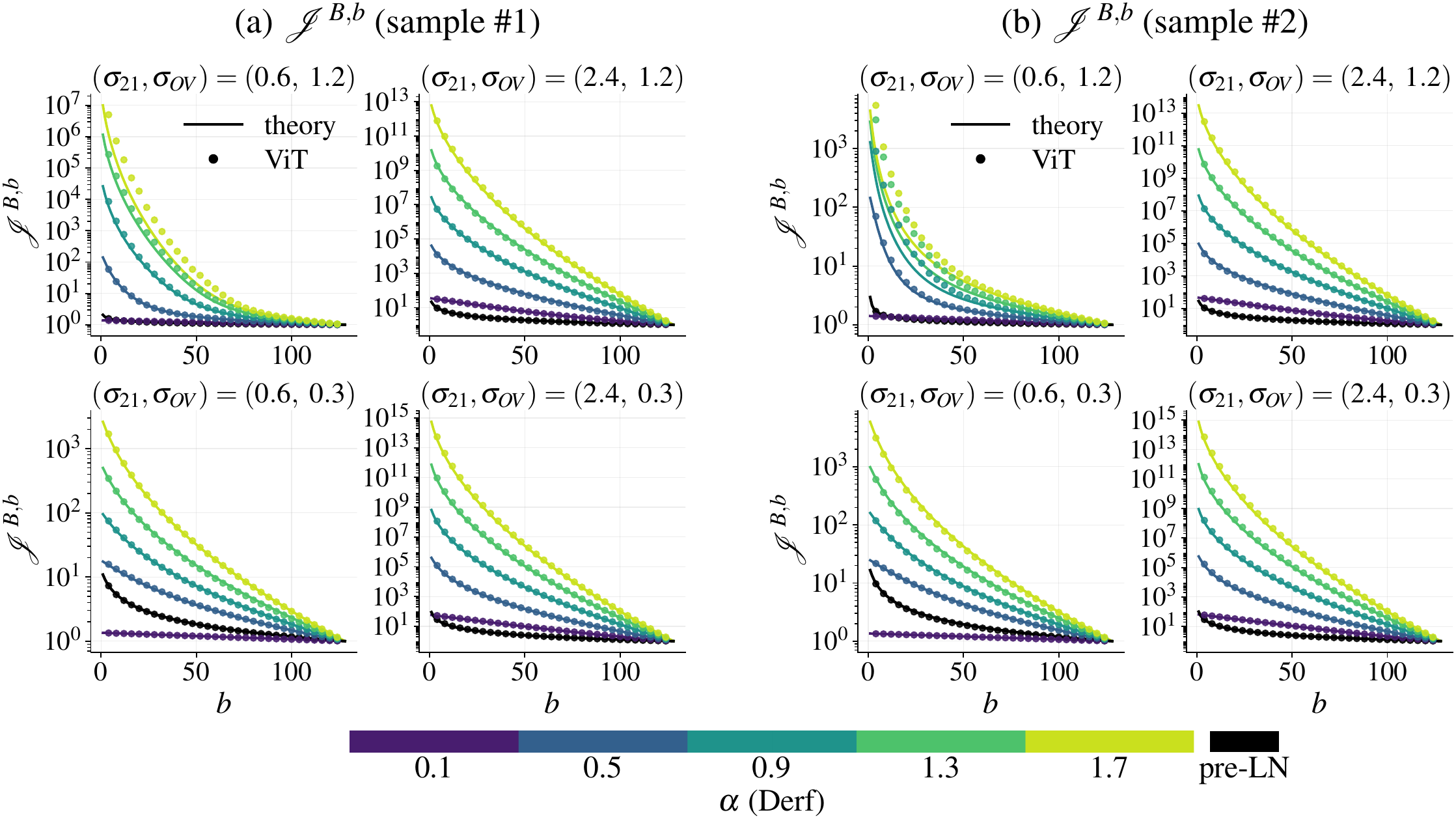}
\caption{
Backward APJN $\mathscr{J}^{B,b}$: ViT on CIFAR-100 inputs versus theory comparison. \textbf{(a)} and \textbf{(b)} show two different CIFAR-100 input samples. Circles denote APJN values measured in ViT models (the pre-LN variant and Derf variants with multiple values of $\alpha$). Solid lines denote the backward APJN values predicted by the theoretical model in Eqs.~(\ref{eq:eq-q-recurrence})--(\ref{eq:eq-chi-J}).
}
\label{fig:apjn-fits}
\end{figure}

\textbf{Backward-APJN comparison examples.} Fig.~\ref{fig:apjn-fits} shows example comparisons between measured and theoretical backward-APJN curves for two randomly selected samples from the experiment shown in Fig.~\ref{fig:q_pq_apjn}(b). As discussed in the main text, the case $(\sigma_{21}, \sigma_{OV})=(0.6,\,1.2)$ shows worse agreement, likely because larger attention weights make the APJNs more sensitive to deviations from the permutation-symmetry assumption of the theory.

\begin{figure}[tbp]
  \centering
\includegraphics[width=14.0cm]{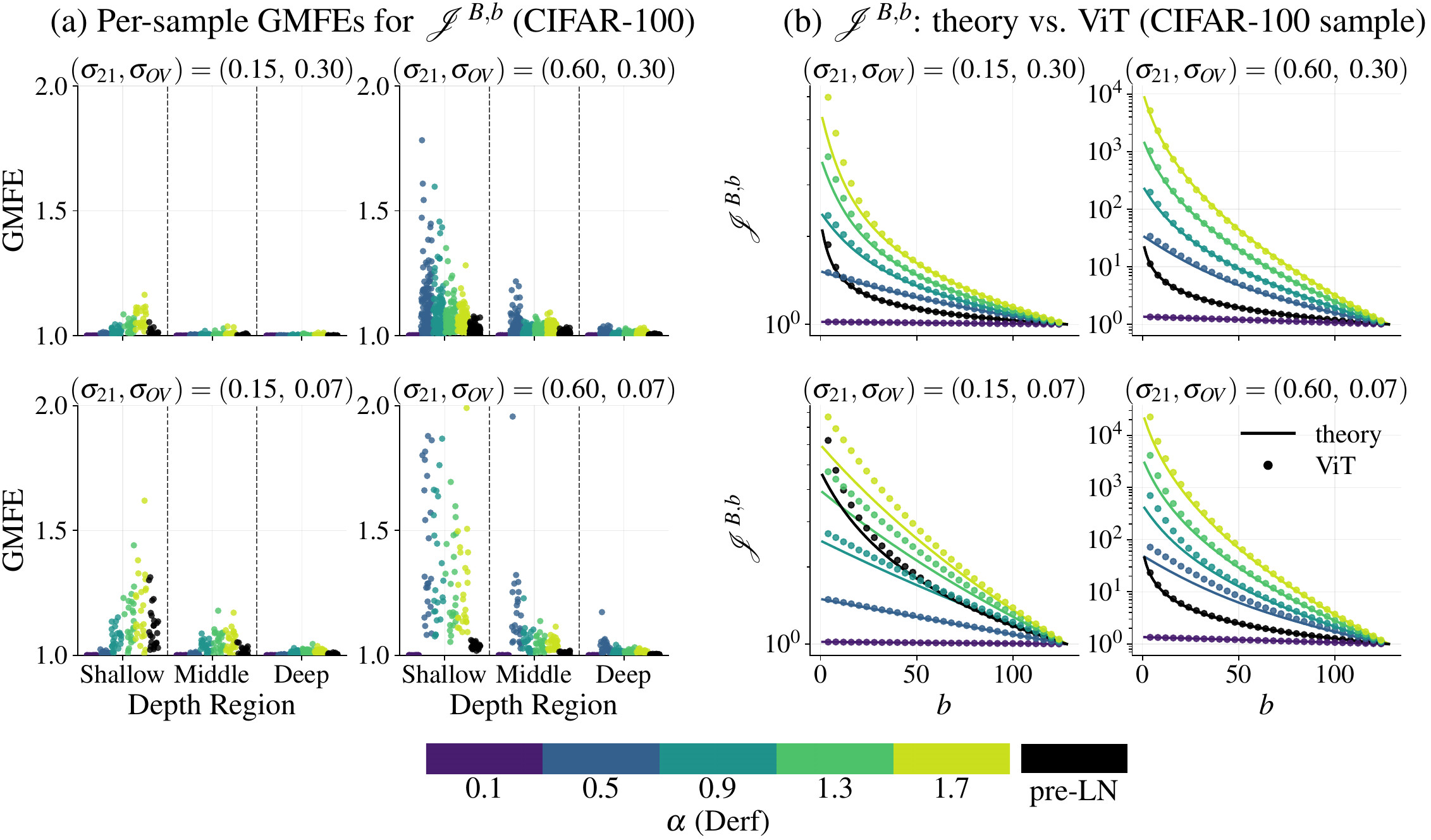}
\caption{Backward APJN: ViT with CIFAR-100 inputs vs. theory at additional values of $(\sigma_{21}, \sigma_{OV})$.
\textbf{(a)} Per-sample GMFE values between the theoretical values of $\mathscr{J}^{B,b}$ and those measured in a ViT with $B=128$ blocks on CIFAR-100 inputs, evaluated in early, middle, and deep layers.
\textbf{(b)} Backward APJN $\mathscr{J}^{B,b}$ for a ViT on a random CIFAR-100 input, compared with theory. Circles denote APJN values measured in ViT models, and solid lines denote the backward APJN values predicted by the theoretical model in Eqs.~(\ref{eq:eq-q-recurrence})--(\ref{eq:eq-chi-J}).
}
\label{fig:apjn-back-more}
\end{figure}

\textbf{Backward APJN: ViT with CIFAR-100 inputs vs. theory at additional values of $(\sigma_{21}, \sigma_{OV})$.}
Fig.~\ref{fig:apjn-back-more} repeats the experiments in Fig.~\ref{fig:q_pq_apjn} from the main text for backward APJNs at additional, smaller values of $(\sigma_{21}, \sigma_{OV})$. Note that smaller weight variances cause the activations to preserve a complex distribution of dot products up to deeper transformer blocks. As a result, the theoretical assumptions become applicable only in deeper layers, and one should not expect the theory to accurately capture the APJN curve shape in this regime, as illustrated, for example, by the case $(\sigma_{21}, \sigma_{OV})=(0.15,\,0.07)$ in Fig.~\ref{fig:apjn-back-more}(b).

\begin{figure}[tbp]
  \centering
\includegraphics[width=14.0cm]{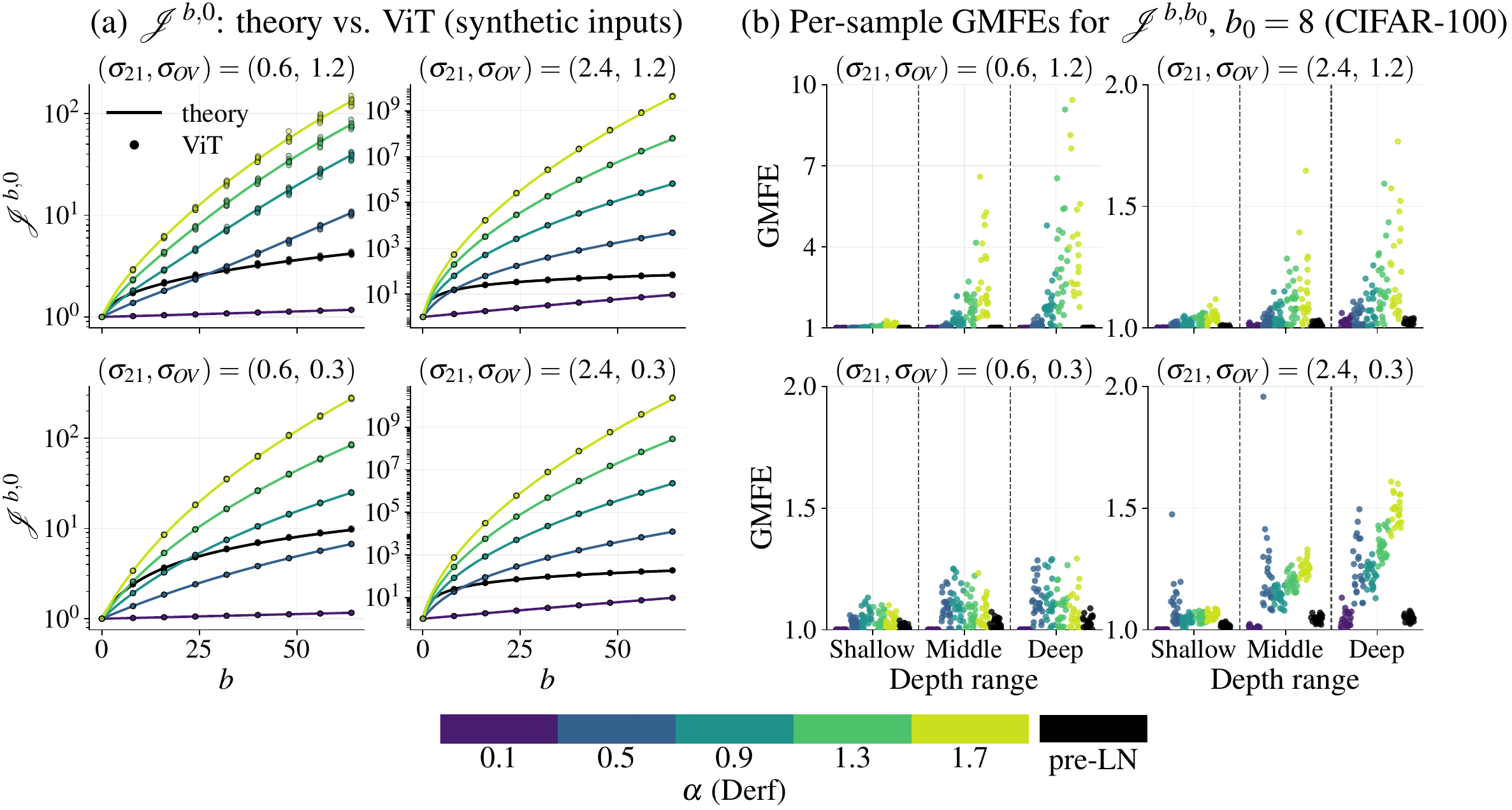}
\caption{
\textbf{(a)} Forward APJN $\mathscr{J}^{b,0}$: theory vs.\ ViT on synthetic permutation-symmetric inputs. Circles denote forward APJN values measured in the ViT model on several permutation-symmetric inputs, for both the pre-LN variant and Derf variants with several values of $\alpha$. Solid lines denote the forward APJN values predicted by the theoretical model in Eqs.~(\ref{eq:eq-q-recurrence})--(\ref{eq:eq-chi-J}).
\textbf{(b)} Per-sample GMFE values between the theoretical values of $\mathscr{J}^{b,b_0},\ b_0=8$ and those measured in a ViT with $B=128$ blocks on 24 CIFAR-100 inputs, in early, middle, and deep layers.
}
\label{fig:forward-ajns-synth-gmfes}
\end{figure}

\textbf{Forward APJNs.} Fig.~\ref{fig:forward-ajns-synth-gmfes} repeats the experiments in Fig.~\ref{fig:q_pq_apjn} from the main text for forward APJNs.

Fig.~\ref{fig:forward-ajns-synth-gmfes}(a) compares the forward APJNs $\mathscr{J}^{b,0}$ measured in ViT model variants on several synthetic permutation-symmetric inputs with $(q^0, p^0) = (1.0,\ 0.2)$ against the APJNs predicted by the theoretical recurrence relations in Eqs.~(\ref{eq:eq-q-recurrence}), (\ref{eq:eq-p-recurrence}), and (\ref{eq:eq-chi-J}).

\begin{figure}[tbp]
  \centering
\includegraphics[width=13.0cm]{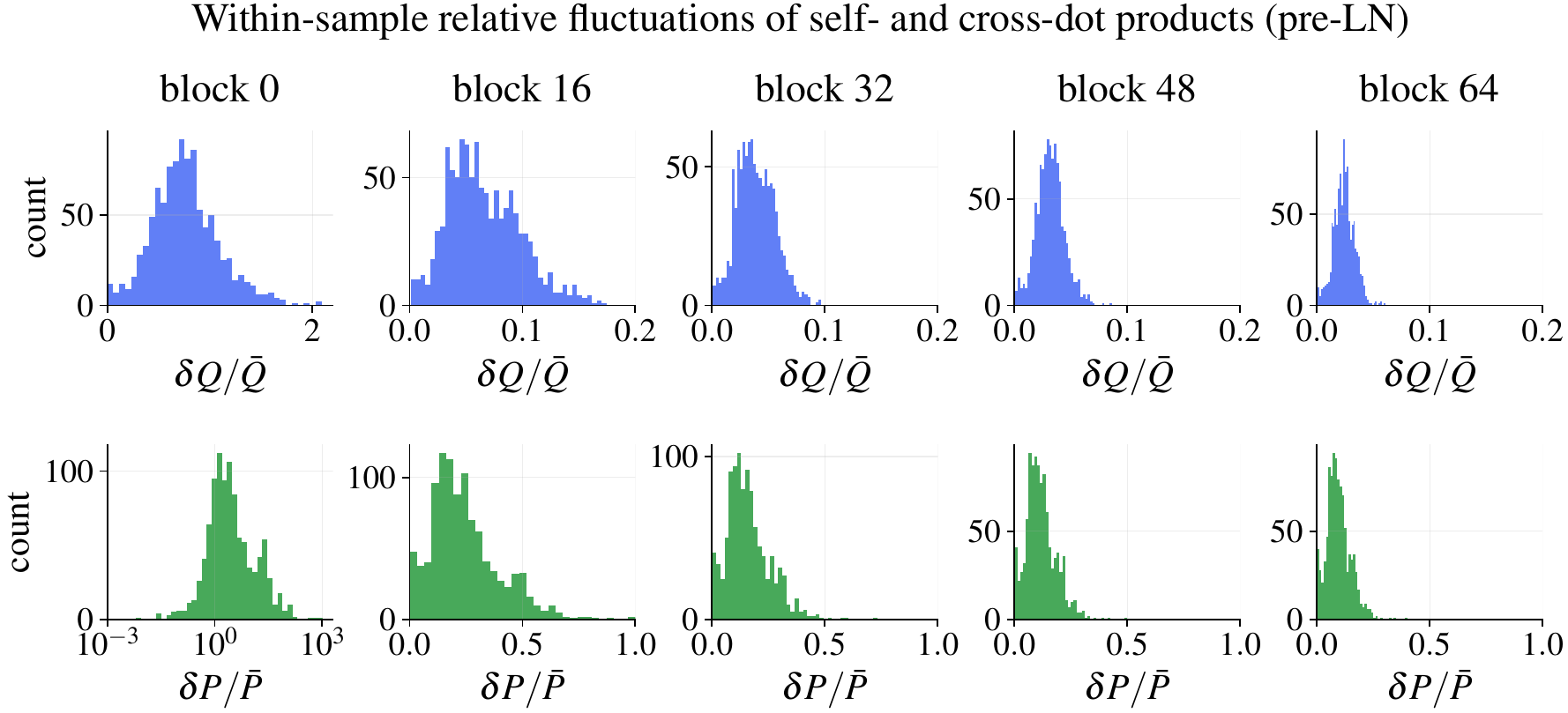}
  \caption{Pre-LN: Histograms of within-sample relative fluctuations of self-dot products (\textbf{upper row}) and cross-positional dot products between activation vectors (\textbf{lower row}) at multiple transformer blocks.
  }
  \label{fig:qp_within_hist_preln}
\end{figure}

\begin{figure}[tbp]
  \centering
\includegraphics[width=13.0cm]{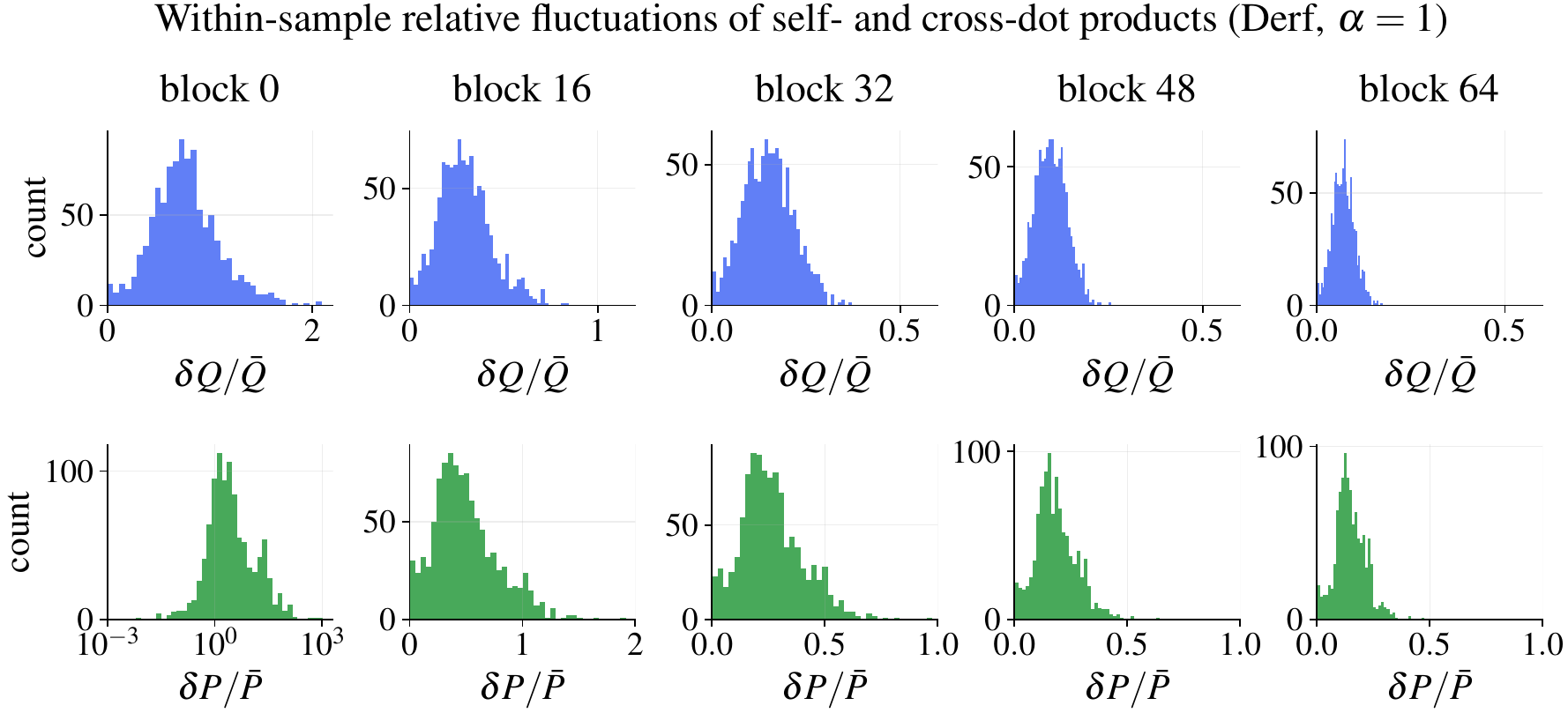}
  \caption{Derf, $\alpha=1$: Histograms of within-sample relative fluctuations of self-dot products (\textbf{upper row}) and cross-positional dot products between activation vectors (\textbf{lower row}) at multiple transformer blocks.
  }
  \label{fig:qp_within_hist_derf}
\end{figure}

Fig.~\ref{fig:forward-ajns-synth-gmfes}(b) shows per-sample GMFE scores for forward APJN values, comparing theoretical predictions with measurements on randomly selected CIFAR-100 samples. For this comparison, we shift the reference block to $b_0=8$ and compare the measured $\mathscr{J}^{b,b_0}$ with theory. We do so because the theory does not accurately capture the overall scale of the measured pre-LN forward APJN values when the reference is taken at the input; this scale is determined by the first few transformer blocks. This discrepancy may arise because forward APJNs involve differentiation with respect to the inputs, whose statistics differ substantially from those assumed in the theory. By contrast, activations in later blocks are expected to more closely satisfy the theoretical assumptions: after a few transformer blocks, the variances of the self-dot products and cross-dot products among activation vectors become smaller relative to their mean values, as illustrated in Figs.~\ref{fig:qp_within_hist_preln} and \ref{fig:qp_within_hist_derf}.

\begin{figure}[tbp]
  \centering
\includegraphics[width=14.0cm]{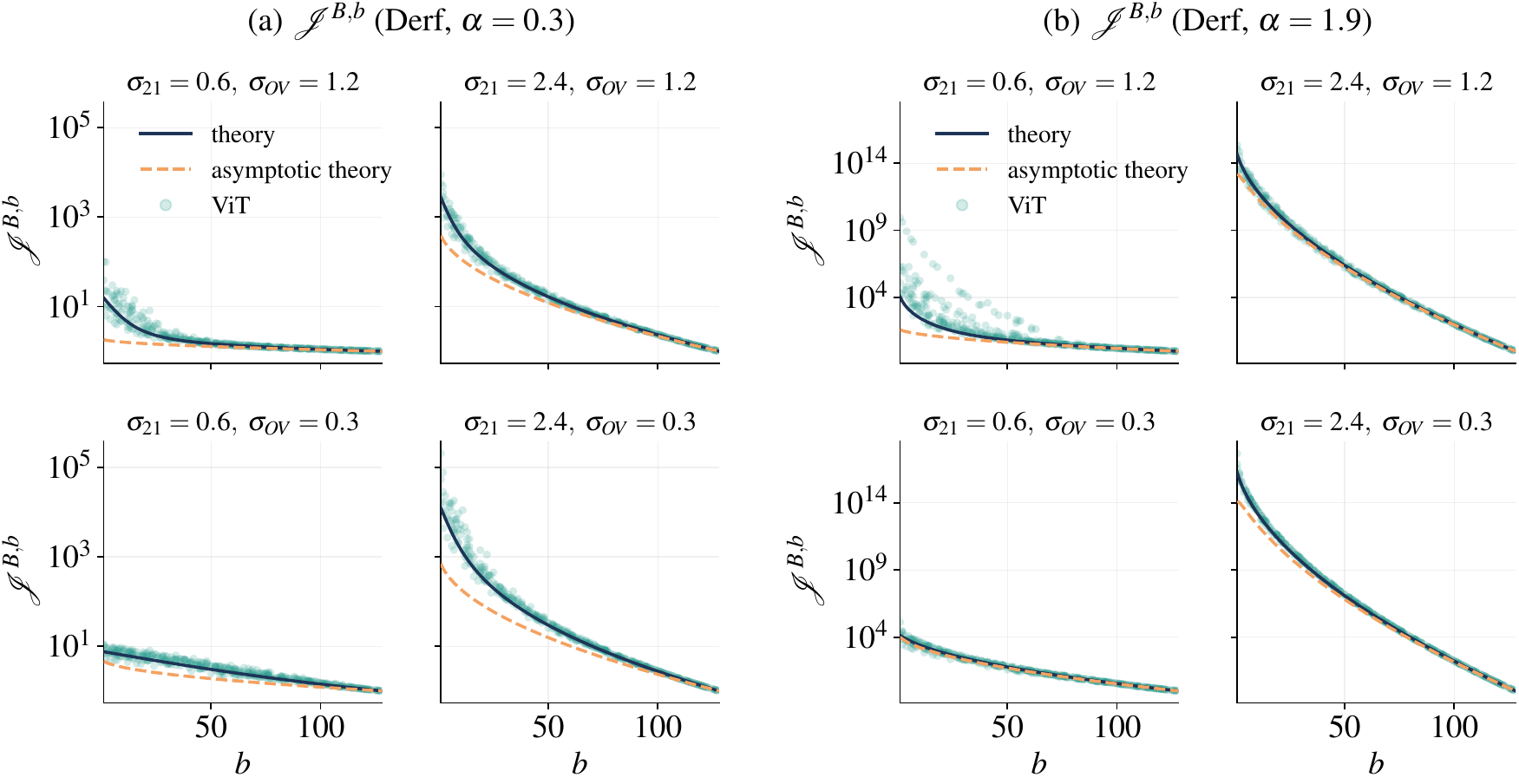}
  \caption{
Backward APJN $\mathscr{J}^{B,b}$ for ViT with Derf, together with the full-theory prediction and the asymptotic prediction from Eq.~(\ref{eq:derf-asymptotic-apjn}).
\textbf{(a)} $\alpha=0.3$; \textbf{(b)} $\alpha=1.9$.
Teal circles indicate backward APJN values measured in ViT on 100 random training samples from CIFAR-100. Black solid lines show the full-theory predictions obtained from the recurrence relations (\ref{eq:eq-q-recurrence}), (\ref{eq:eq-p-recurrence}), and (\ref{eq:eq-chi-J}) with $(q^0,\,p^0)=(0.5,\,0.25)$, while orange dashed lines show the large-$b$ asymptotic predictions.
}
  \label{fig:two_alpha_theory_asympt}
\end{figure}

\textbf{Comparison of backward APJNs on CIFAR-100 inputs for Derf with $\alpha = 0.3$ and $\alpha = 1.9$ against the theory.} Fig.~\ref{fig:two_alpha_theory_asympt} is analogous to Fig.~\ref{fig:asymp} in the main text for different values of $\alpha$. 
The asymptotic expression for the APJN is less accurate for models with smaller $\alpha$, even in networks with more than one hundred transformer blocks. This is because the derivation of the asymptotic APJN assumes that $l$ is large enough for $\tilde q^l$ and $\tilde p^l$ in Eqs.~(\ref{eq:eq-q-recurrence}) and (\ref{eq:eq-p-recurrence}) to approach their asymptotic values. However, the layer at which this occurs depends on the nonlinearity parameter $\alpha$ and may lie beyond the network depth. Indeed, as discussed in the main text, smaller $\alpha$ leads to slower growth of $q^l$ with $l$. Consequently, the asymptotic APJN expressions in Eqs.~(\ref{eq:pre-ln-asymptotic-apjn}) and (\ref{eq:derf-asymptotic-apjn}) become valid only in deeper layers for smaller $\alpha$ and may therefore not apply in finite-depth networks. Nevertheless, the full APJN theory remains accurate at small $\alpha$, as illustrated in Fig.~\ref{fig:two_alpha_theory_asympt}(a) for $\alpha = 0.3$ in Appendix~\ref{sec-app:supplement-sig-prop}.

\begin{figure}[tbp]
  \centering
\includegraphics[width=14.0cm]{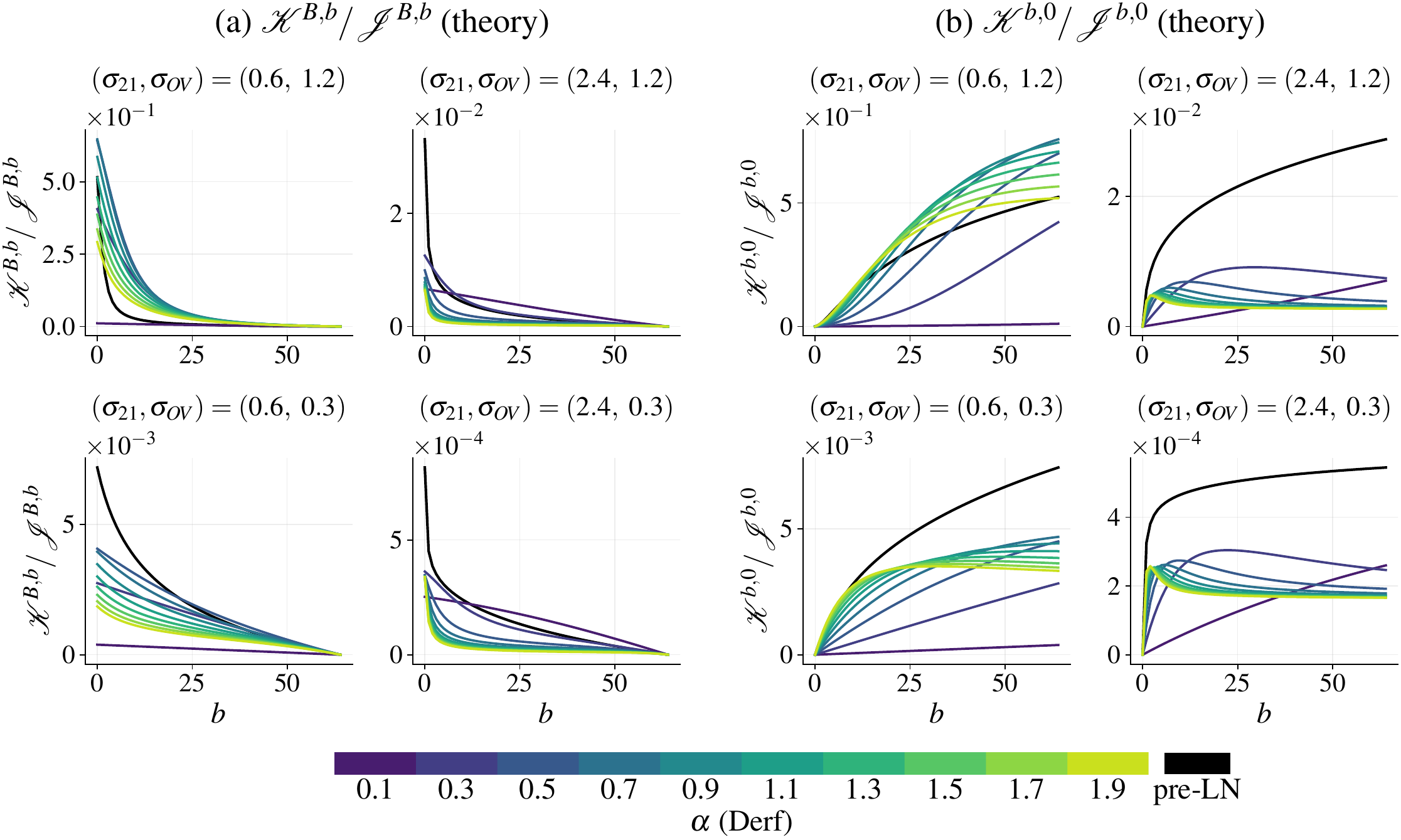}
  \caption{
Ratio of the cross-positional Jacobian correlation to the APJN, computed from theory for multiple values of $(\sigma_{21}, \sigma_{OV})$.
\textbf{(a)} Ratio $\mathscr{K}^{B,b}/\mathscr{J}^{B,b}$.
\textbf{(b)} Ratio $\mathscr{K}^{b,0}/\mathscr{J}^{b,0}$.
The cross-positional Jacobian correlation becomes non-negligible for larger values of $\sigma_{OV}$.
}
\label{fig:k_j_ratio}
\end{figure}

\textbf{Ratio of the cross-positional Jacobian-correlation term to the APJN in the full recurrence relations of Eqs.~(\ref{eq:j-forward-recurrence}) and (\ref{eq:j-backward-recurrence});} shown in Fig.~\ref{fig:k_j_ratio} for several values of $(\sigma_{21}, \sigma_{OV})$. The cross-positional Jacobian-correlation terms, $\mathscr{K}^{B,b}$ and $\mathscr{K}^{b,0}$, can become non-negligible for larger values of $\sigma_{OV}$, as illustrated by the case $(\sigma_{21}, \sigma_{OV}) = (0.6, 1.2)$. The quantity $\mathscr{K}^{b', b}$ is defined analogously to the other block-level quantities in the text: $\mathscr{K}^{b', b} = \mathcal{K}^{2b', 2b}$.

\begin{figure}[tbp]
  \centering
\includegraphics[width=13.0cm]{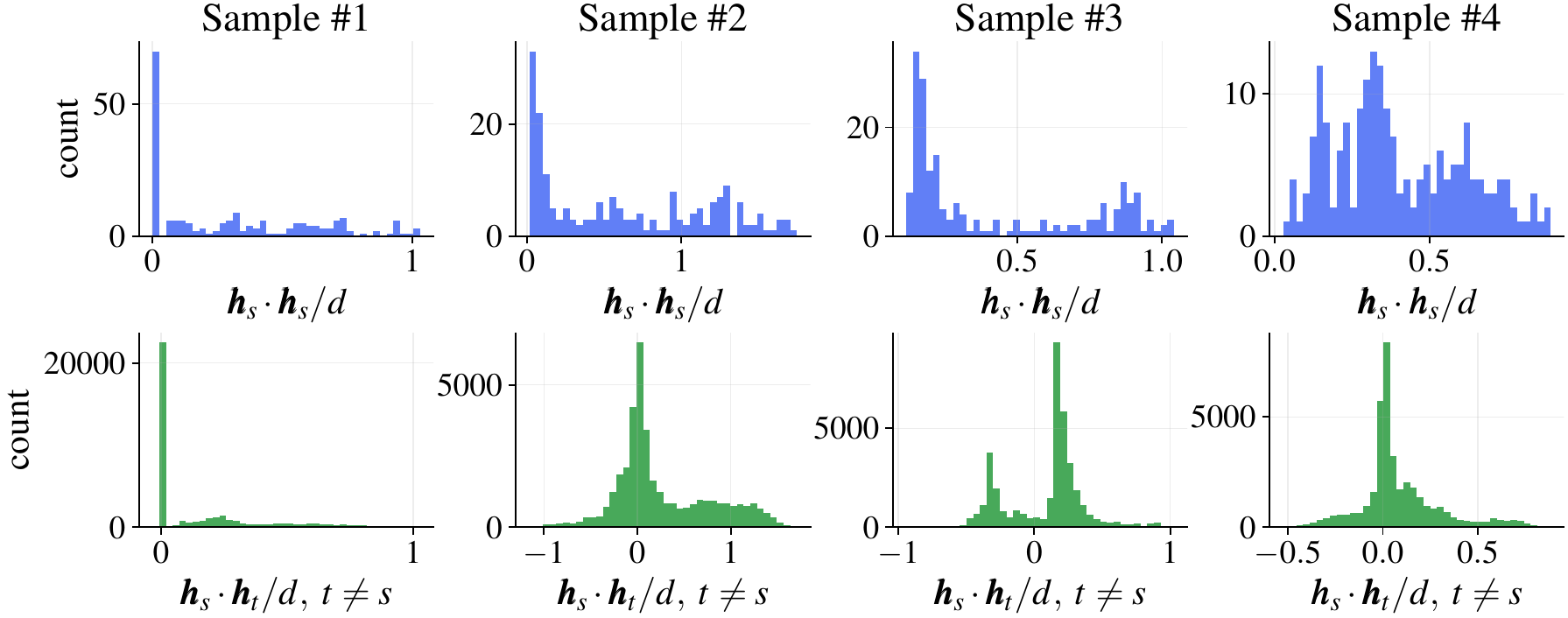}
  \caption{Histograms of normalized activation vector norms and normalized dot products between activation vectors at different positions for four CIFAR-100 training samples, measured after random augmentation, feature extraction, and positional encoding, immediately before the first transformer block.
  }
  \label{fig:q_p_histograms}
\end{figure}

\textbf{Input statistics and the evolution of activation statistics across transformer blocks.} Fig.~\ref{fig:q_p_histograms} shows the input statistics for four random CIFAR-100 samples, namely the normalized activation-vector norms and cross-positional dot products. These statistics clearly violate the simplifying assumption of permutation invariance among activation vectors used in the theoretical derivation in the main text.

\begin{figure}[tbp]
  \centering
\includegraphics[width=13.0cm]{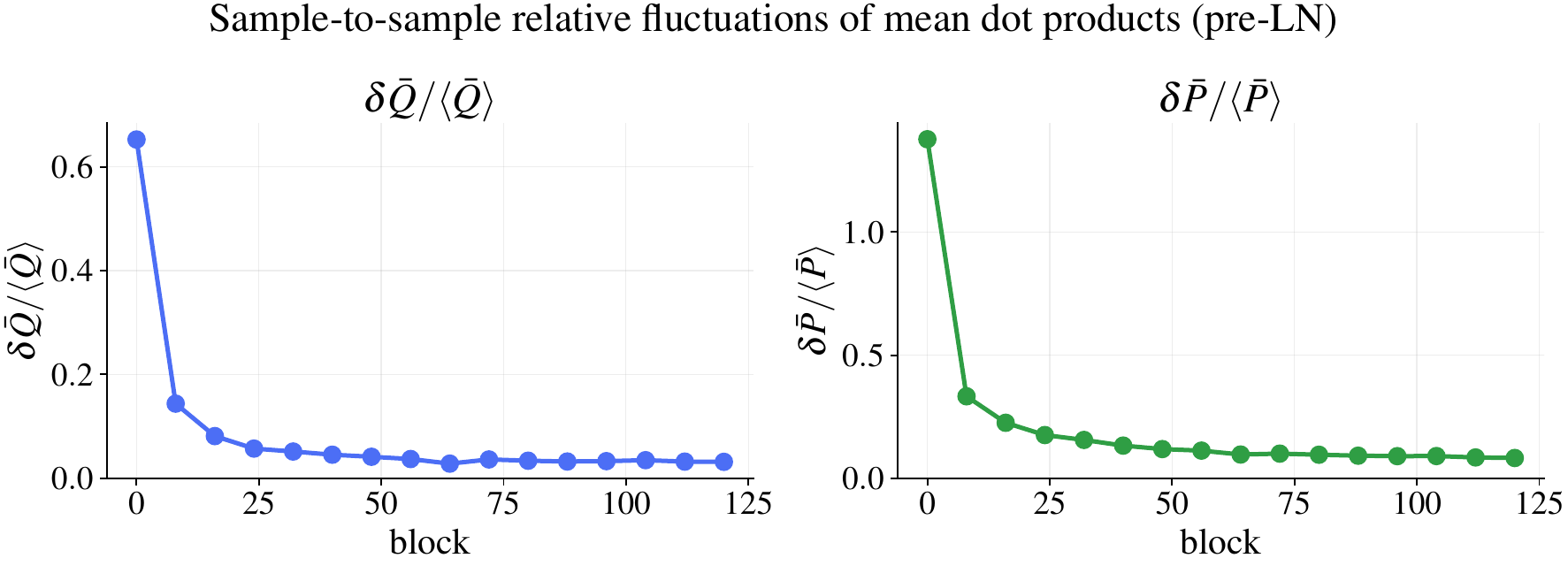}
  \caption{Pre-LN: Sample-to-sample relative fluctuations of mean self-dot products (\textbf{left}) and mean cross-positional dot products between activation vectors (\textbf{right}) across transformer blocks.
  }
  \label{fig:qp_uniformize_preln}
\end{figure}

\begin{figure}[tbp]
  \centering
\includegraphics[width=13.0cm]{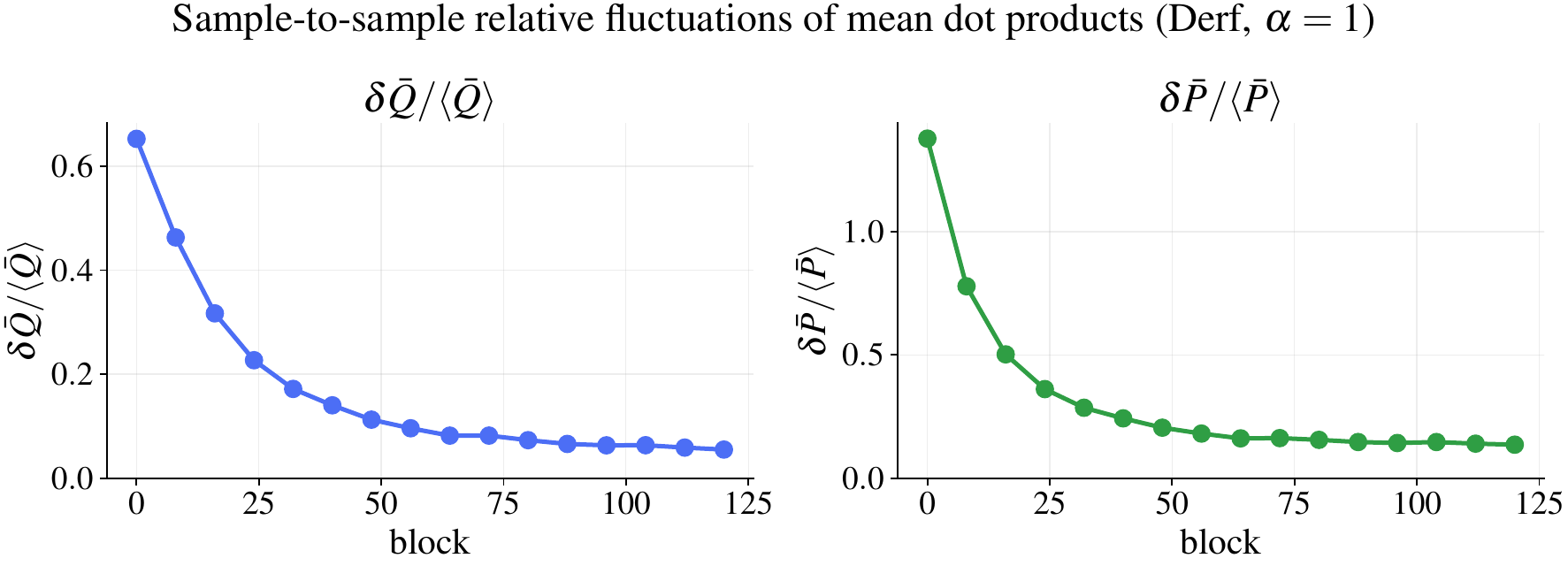}
  \caption{
  Derf, $\alpha=1$: Sample-to-sample relative fluctuations of mean self-dot products (\textbf{left}) and mean cross-positional dot products between activation vectors (\textbf{right}) across transformer blocks.
  }
  \label{fig:qp_uniformize_derf}
\end{figure}

Figs.~\ref{fig:qp_within_hist_preln}--\ref{fig:qp_uniformize_derf} show how the activation statistics evolve across transformer blocks in the pre-LN ViT model and the Derf ViT model with $\alpha = 1$, based on 1000 CIFAR-100 input samples. With a slight abuse of notation, $\bar{Q}$ and $\bar{P}$ denote the averages over positions of the normalized self-dot products and cross-positional dot products of activation vectors, respectively, at a given transformer block for a fixed input sample. $\delta Q$ and $\delta P$ denote the corresponding standard deviations; these are therefore within-sample quantities. In turn, $\langle \bar{Q} \rangle$ and $\langle \bar{P} \rangle$ denote averages across input samples, with corresponding standard deviations $\delta \bar{Q}$ and $\delta \bar{P}$.

Specifically, Figs.~\ref{fig:qp_within_hist_preln} and \ref{fig:qp_within_hist_derf} show how the distribution of the within-sample relative dot-product fluctuations, $\delta Q/\bar{Q}$ and $\delta P/\bar{P}$, evolves across transformer blocks. For both self-dot products and cross-positional dot products, these fluctuations decrease with depth. This suggests that the theoretical assumptions become more accurate for activations in deeper layers. In particular, the theory provides a better description of the empirically measured APJN $\mathscr{J}^{b',b}$ at larger values of $b$ and $b'$.

Additionally, Figs.~\ref{fig:qp_uniformize_preln} and \ref{fig:qp_uniformize_derf} show how the sample-to-sample relative fluctuations of these mean quantities, $\delta \bar{Q}/\langle\bar{Q}\rangle$ and $\delta \bar{P}/\langle\bar{P}\rangle$, evolve across transformer blocks. For both self-dot products and cross-positional dot products, these relative fluctuations decrease with depth. This suggests that the dependence on the specific input statistics gradually weakens.

\clearpage

\section{Supplementary Results on Training Stability Experiments}
\label{sec-app:supplement-stability-exps}
\subsection{Gradient Amplification in ViT: Measurements and Analysis}
\label{subsec-app:grad_amplification}
\begin{figure}[tbp]
  \centering
\includegraphics[width=14.0cm]{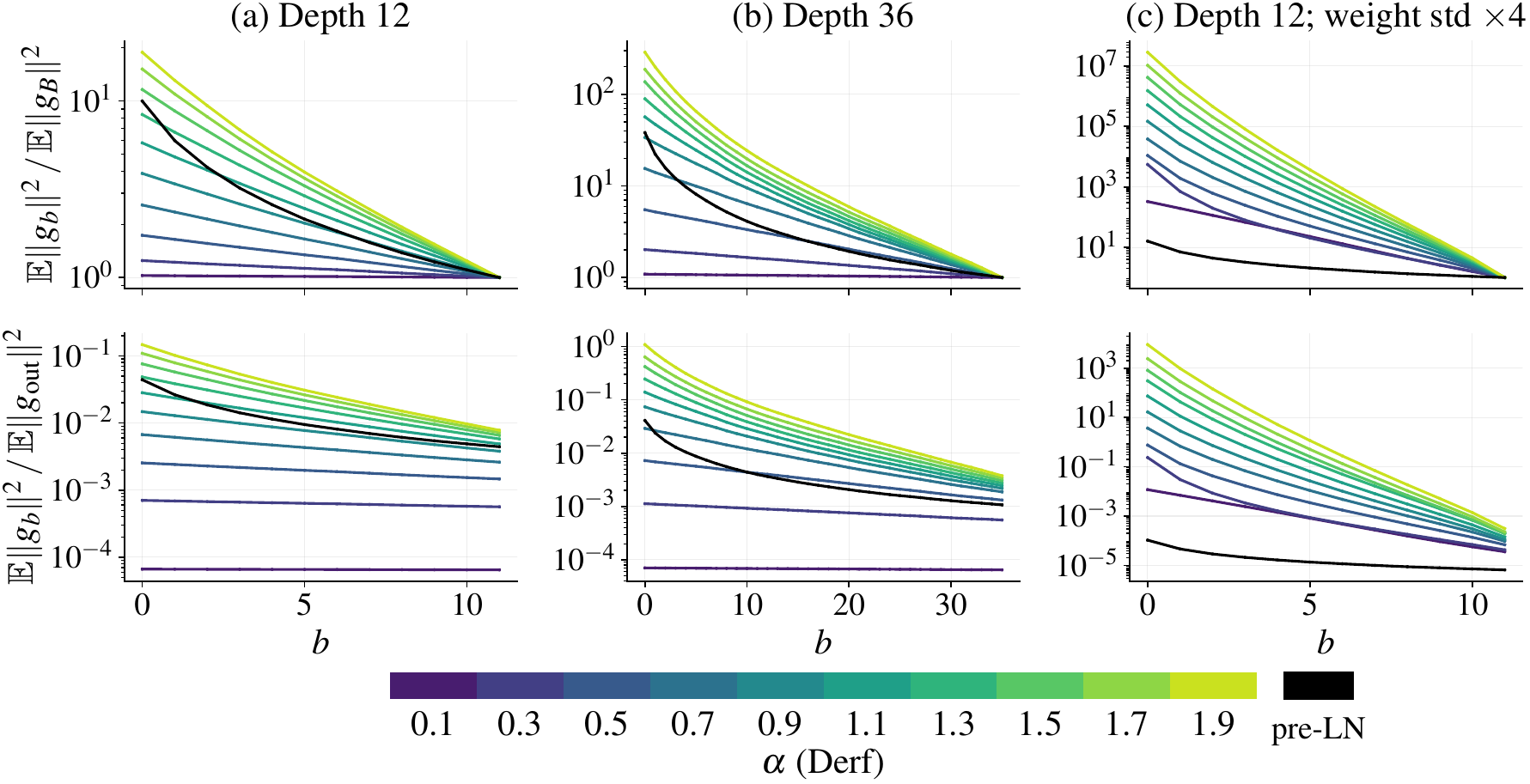}
  \caption{
  Gradient amplification in the ViT model from \textbf{(upper row)} the last transformer block to earlier blocks, and from \textbf{(lower row)} the final normalization layer to earlier blocks: Derf vs.\ pre-LN. \textbf{(a)} Depth $B=12$. \textbf{(b)} Depth $B=36$. As the network depth increases, the gradient amplification in pre-LN matches that of Derf models with smaller $\alpha$. In addition, smaller $\alpha$ results in smaller overall gradient magnitudes because of gradient propagation through the final normalization layer. \textbf{(c)} Depth $B=12$, with the MLP and attention weight standard deviations increased by a factor of $4$, leading to substantially stronger gradient amplification in Derf than in pre-LN.
  }
  \label{fig:grad_amplification}
\end{figure}

Fig.~\ref{fig:grad_amplification} shows the gradient amplification in the ViT model from the last transformer block to block $b$, $\mathbb{E}\lVert g_b\rVert^2 / \mathbb{E}\lVert g_B\rVert^2$ (upper row), and from the final normalization layer to block $b$, $\mathbb{E}\lVert g_b\rVert^2 / \mathbb{E}\lVert g_{\text{out}}\rVert^2$ (lower row), where $g_b$ denotes the gradient with respect to the output of the $b$th transformer block, and $g_{\text{out}}$ denotes the gradient with respect to the output of the final normalization layer. 
The expectations are estimated using a batch of 32 CIFAR-100 training examples. 

Fig.~\ref{fig:grad_amplification}(a) shows that, for $B=12$, the gradient amplification of pre-LN approximately matches that of Derf with $\alpha \approx 0.9$ in the deeper layers, while its overall gradient amplification is closer to that of Derf with $\alpha \approx 1.3$. Fig.~\ref{fig:grad_amplification}(b) shows that, at greater depth ($B=36$), the pre-LN baseline matches the gradient amplification of Derf variants with smaller values of $\alpha$, namely $\alpha \approx 0.5$--$0.7$, whereas Derf variants with larger values of $\alpha$ exhibit stronger gradient amplification.
Both statements hold for gradient amplification computed either from the last transformer block or from the output of the final normalization layer.
This behavior qualitatively agrees with our theory and may explain why Derf variants with larger values of $\alpha$ become unstable at smaller depth as model depth increases. 

However, we also observe that equal gradient amplification does not imply equal convergence rate. Even when Derf is matched to pre-LN in terms of gradient amplification at a given depth, it still tends to converge more slowly. For example, at depth $12$, Derf with $\alpha=0.8$ converges more slowly than pre-LN in Fig.~\ref{fig:depth_std_tiles}(a), whereas $\alpha=0.6$ yields a convergence rate closer to that of pre-LN, and $\alpha=0.4$ converges faster. 
In Fig.~\ref{fig:gamma-sweep}, we show that tuning the initialization parameter $\gamma$ in pre-LN can partially close this gap.

Fig.~\ref{fig:grad_amplification}(c) shows that increasing the weight scale leads to significantly stronger gradient amplification in the Derf variants than in pre-LN.

\subsection{Supplementary Experiments}

\begin{figure}[tbp]
  \centering
\includegraphics[width=14.0cm]{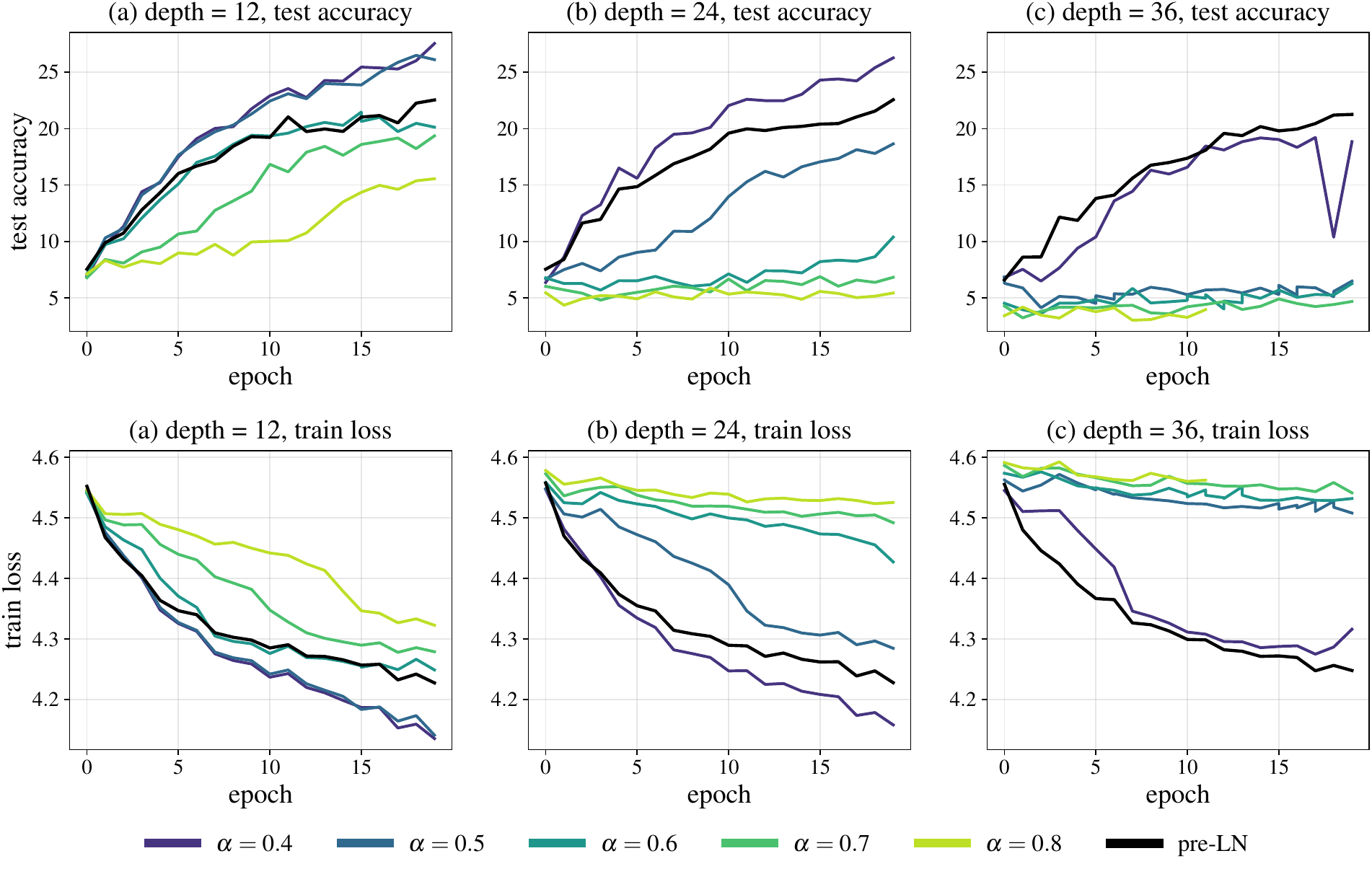}
  \caption{
  \textbf{Upper:} test performance curves; \textbf{lower:} training curves for the training-stability comparison between Derf and pre-LN when varying depth, shown in Fig.~\ref{fig:depth_std_tiles}(a).
  }
  \label{fig:train-test-curves-depth}
\end{figure}

\begin{figure}[tbp]
  \centering
\includegraphics[width=14.0cm]{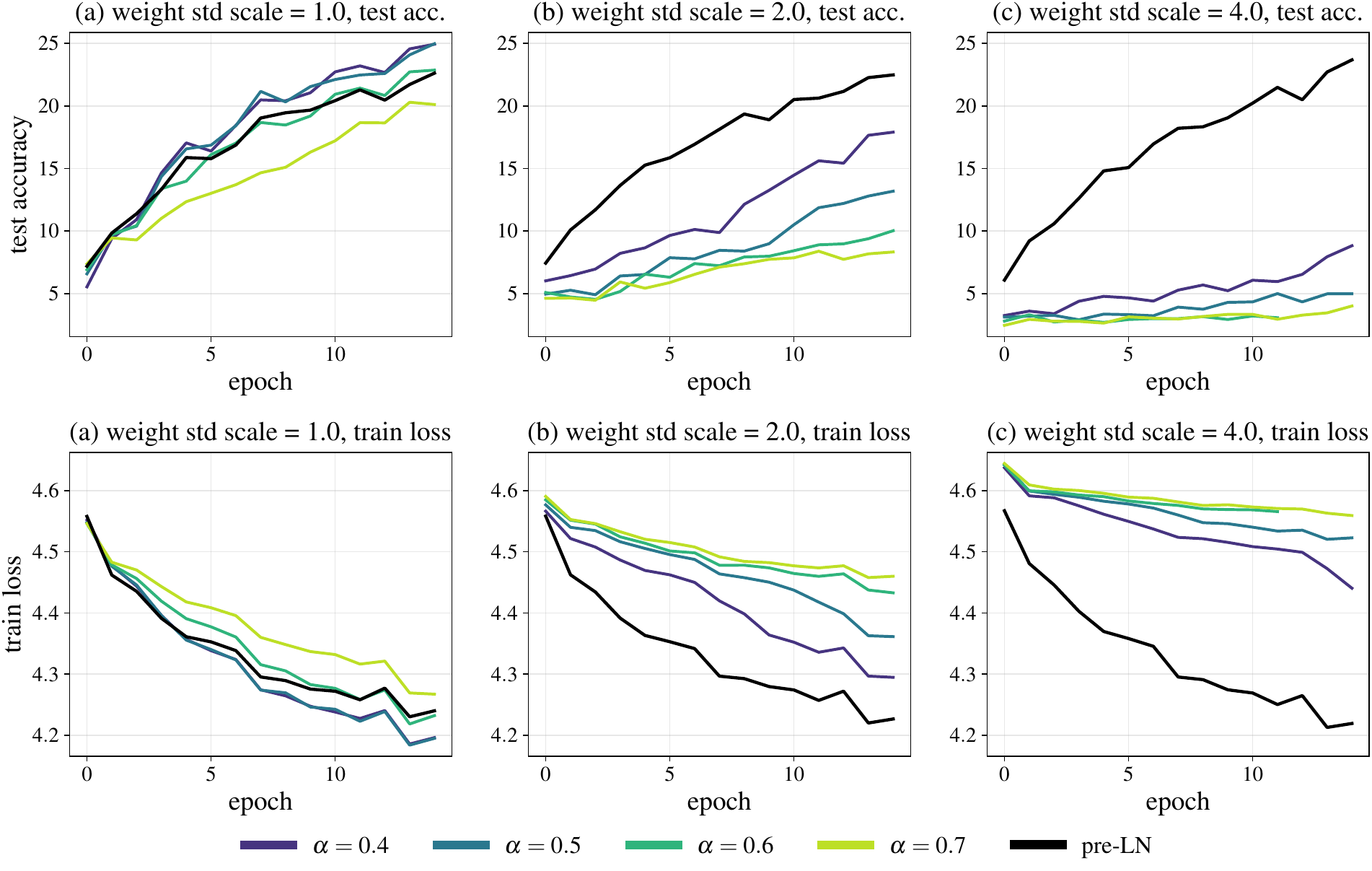}
  \caption{\textbf{Upper:} test performance curves; \textbf{lower:} training curves for the training-stability comparison between Derf and pre-LN when varying the weight standard deviation, shown in Fig.~\ref{fig:depth_std_tiles}(b).}
  \label{fig:train-test-curves-sigma}
\end{figure}

\textbf{Training loss and test performance curves.} Figs.~\ref{fig:train-test-curves-depth} and \ref{fig:train-test-curves-sigma} show the training loss and test performance curves for the experiments presented in Figs.~\ref{fig:depth_std_tiles}(a) and \ref{fig:depth_std_tiles}(b), respectively.

\begin{figure}[tbp]
  \centering
\includegraphics[width=14.0cm]{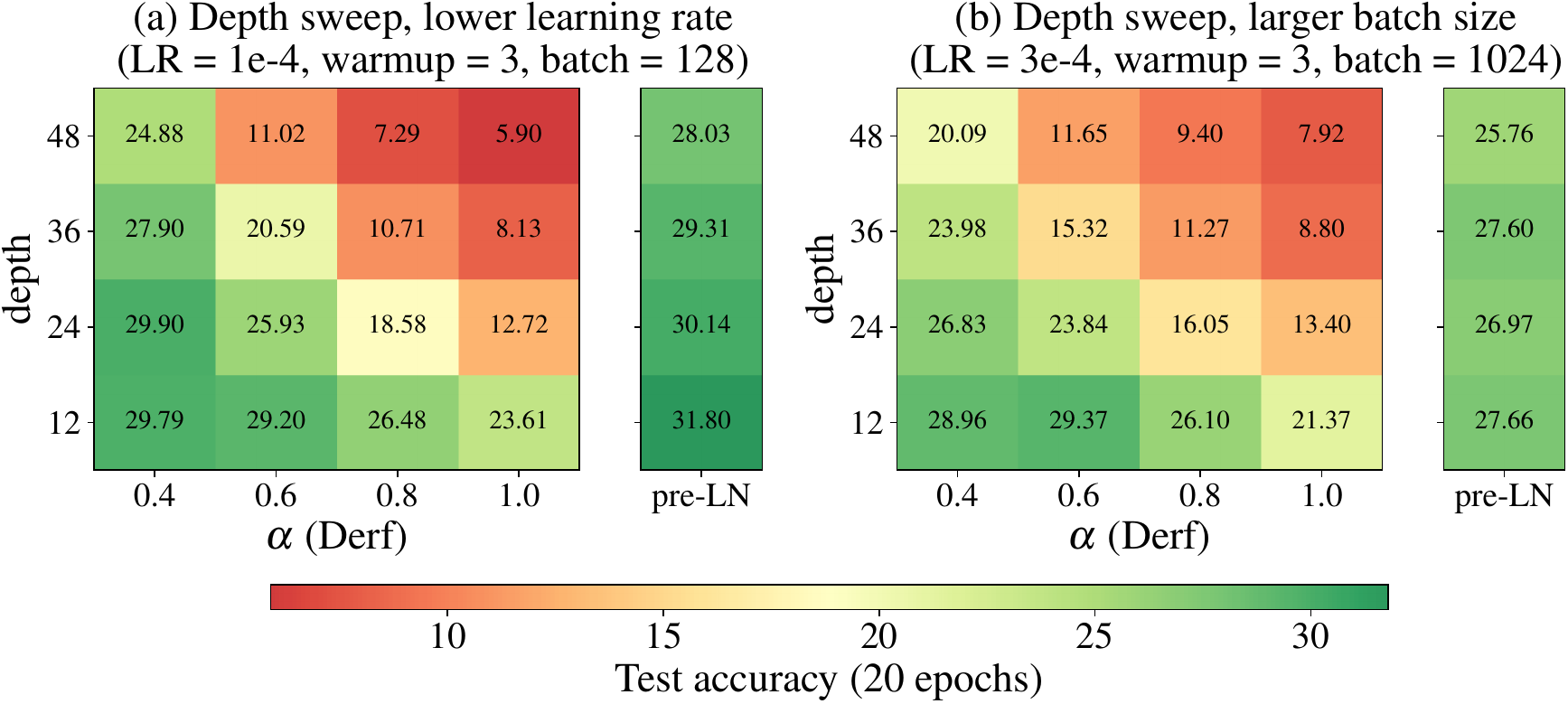}
  \caption{
  Training stability comparison between Derf and pre-LN as depth varies, using \textbf{(a)} a lower learning rate and \textbf{(b)} a larger batch size than in the main experiment in Fig.~\ref{fig:depth_std_tiles}(a). All other parameters are kept identical. Both panels show test accuracy on CIFAR-100 after $20$ training epochs versus depth. Compared to Fig.~\ref{fig:depth_std_tiles}(a), both decreasing the learning rate and increasing the batch size shift the instability boundary toward larger $\alpha$ and larger depth.
  }
  \label{fig:depth-sweep-lr-and-batch}
\end{figure}

\begin{figure}[tbp]
  \centering
\includegraphics[width=14.0cm]{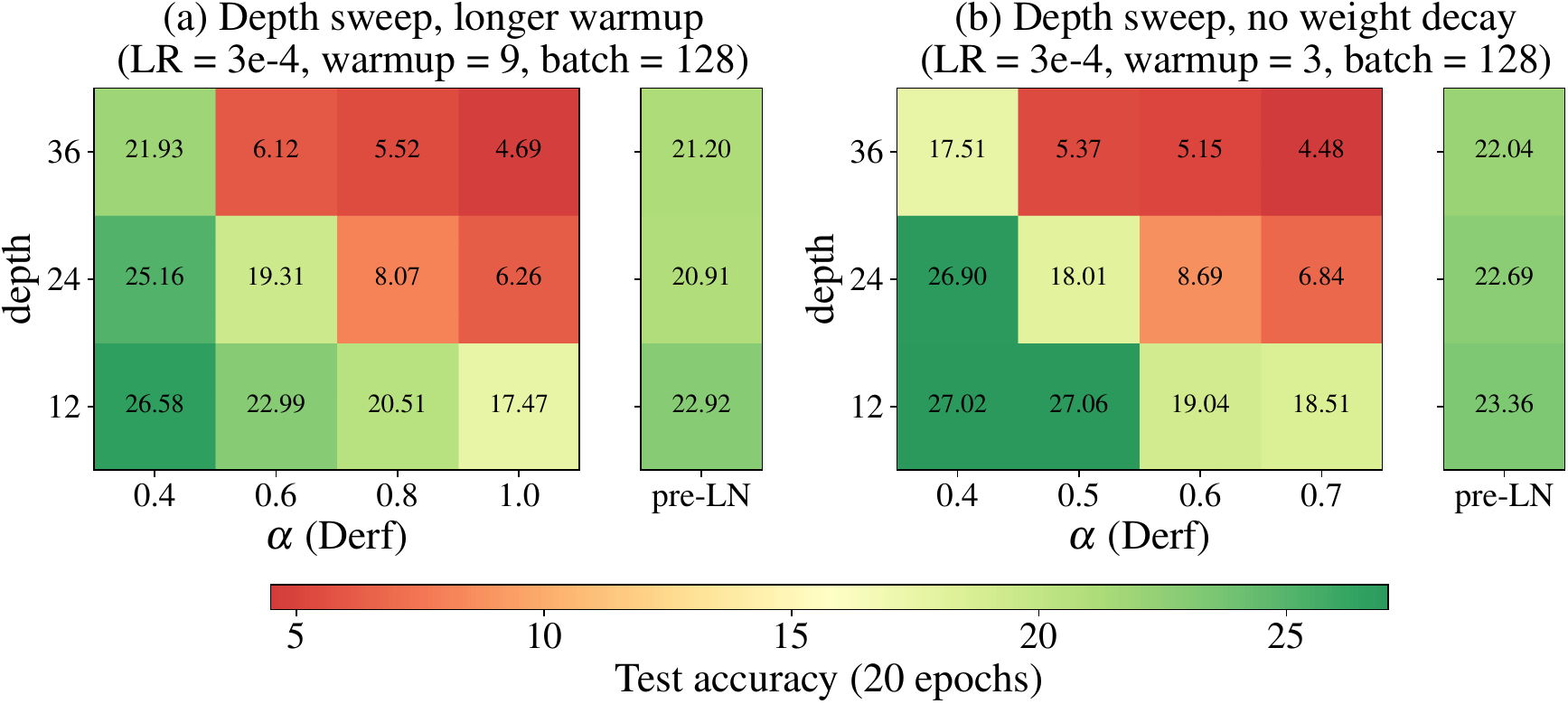}
  \caption{
  Training stability comparison between Derf and pre-LN as depth varies, using \textbf{(a)} a longer warmup and \textbf{(b)} no weight decay, relative to the main experiment shown in Fig.~\ref{fig:depth_std_tiles}(a). All other parameters are kept identical. Both panels show test accuracy on CIFAR-100 after $20$ training epochs as a function of depth. Compared to Fig.~\ref{fig:depth_std_tiles}(a), increasing the number of warmup epochs improves performance near the instability boundary without substantially shifting it, while removing weight decay does not substantially affect the training stability.
  }
  \label{fig:depth-sweep-warmup-no-wd}
\end{figure}

\textbf{Training stability comparison between Derf and pre-LN under varying depth with other hyperparameter settings.} Fig.~\ref{fig:depth-sweep-lr-and-batch} repeats the training stability comparison between Derf and pre-LN shown in Fig.~\ref{fig:depth_std_tiles}(a), using a lower learning rate and a larger batch size than in the main experiment. Fig.~\ref{fig:depth-sweep-warmup-no-wd} repeats this experiment with a longer warmup and without weight decay.

Lowering the learning rate or increasing the batch size shifts the Derf instability boundary toward larger $\alpha$ and greater depth, while increasing the number of warmup epochs improves performance near the boundary without substantially shifting it. By contrast, weight decay has little effect on training stability in this setting.

\begin{figure}[tbp]
  \centering
\includegraphics[width=12.0cm]{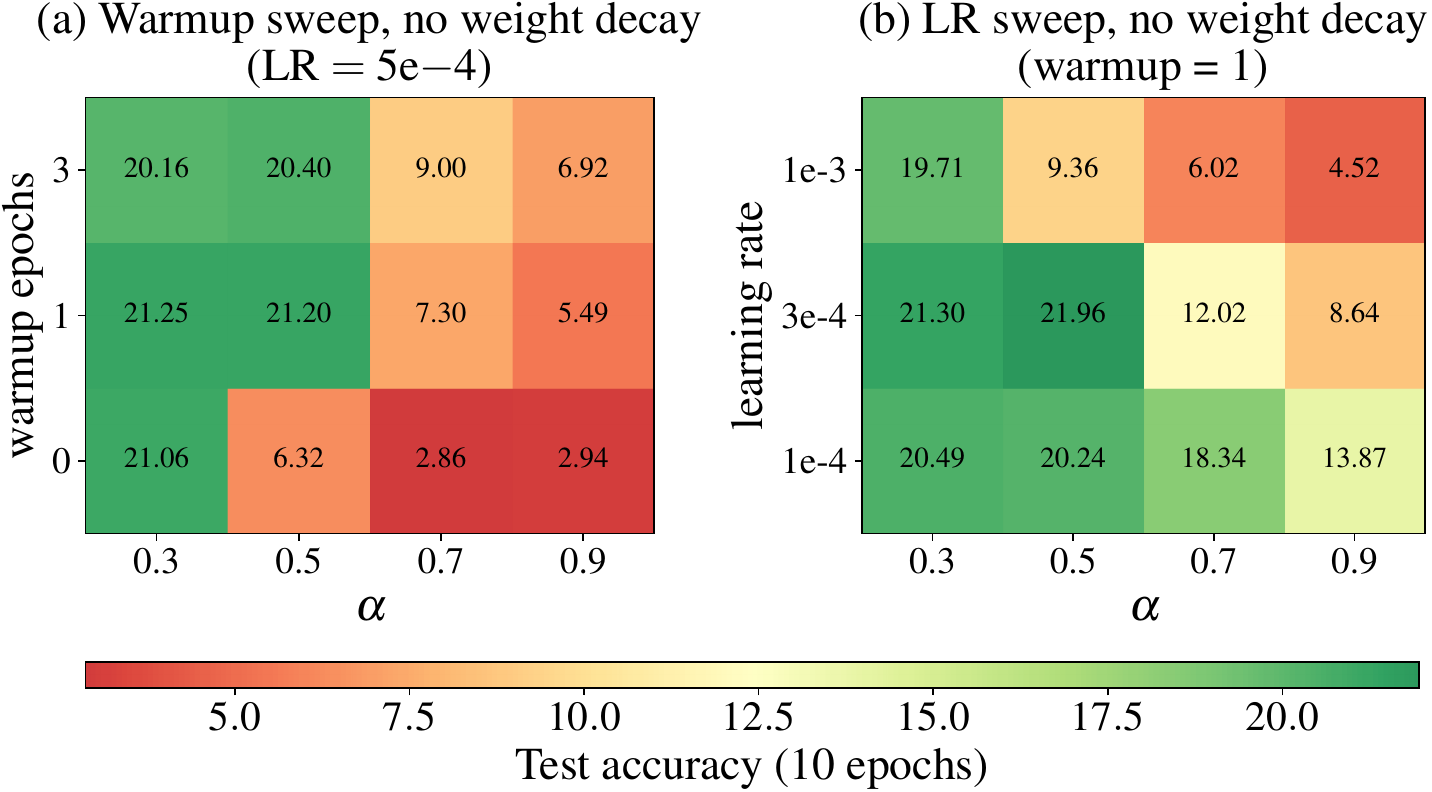}
  \caption{
\textbf{(a)} Effect of the number of warmup epochs and \textbf{(b)} effect of the learning rate on the training stability of Derf for different $\alpha$, \textit{without weight decay}. Comparison to Fig. \ref{fig:warmup_lr_tiles} suggests that moderate weight decay, or its absence, does not significantly affect the training-stability boundary.
  }
  \label{fig:no-wd-warmup-lr-tiles}
\end{figure}

\begin{figure}[tbp]
  \centering
\includegraphics[width=12.0cm]{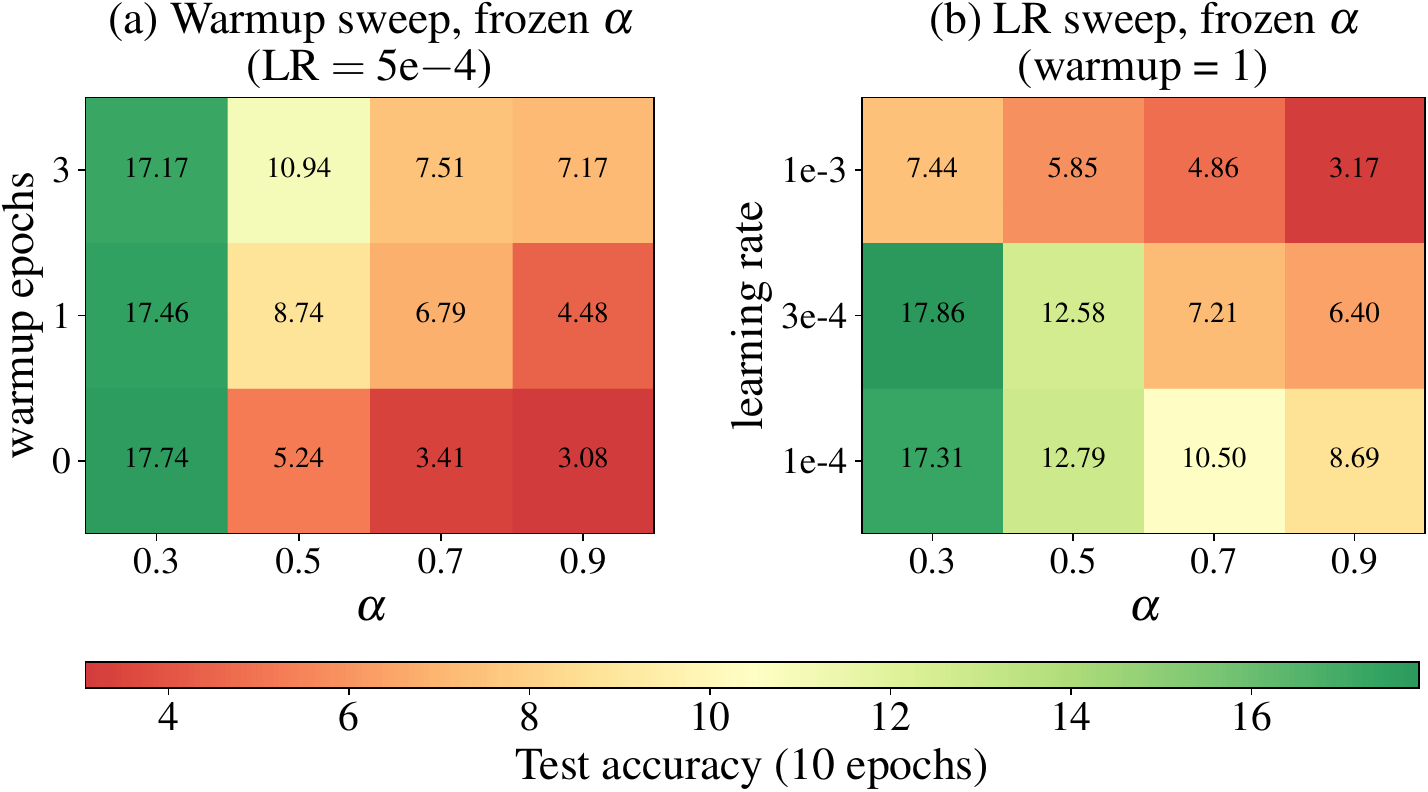}
  \caption{
\textbf{(a)} Effect of the number of warmup epochs and \textbf{(b)} effect of the learning rate on the training stability of Derf for different $\alpha$, \textit{with $\alpha$ frozen during training}. Comparison to Fig.~\ref{fig:warmup_lr_tiles} suggests that making $\alpha$ trainable accelerates training convergence and improves stability at larger $\alpha$ and higher learning rates.
  }
  \label{fig:frozen-alpha-warmup-lr-tiles}
\end{figure}

\textbf{Warmup and learning-rate sweep experiments for Derf in other settings.}
Fig.~\ref{fig:no-wd-warmup-lr-tiles} repeats the experiments in Figs.~\ref{fig:warmup_lr_tiles}(a) and (b) without weight decay in Adam, showing that removing weight decay does not significantly affect training stability. Fig.~\ref{fig:frozen-alpha-warmup-lr-tiles} repeats the experiments in Figs.~\ref{fig:warmup_lr_tiles}(a) and (b) with a non-trainable parameter $\alpha$ fixed at its initialization value. Comparing with Fig.~\ref{fig:warmup_lr_tiles} suggests that making $\alpha$ untrainable worsens training convergence.

\begin{figure}[tbp]
  \centering
\includegraphics[width=14.0cm]{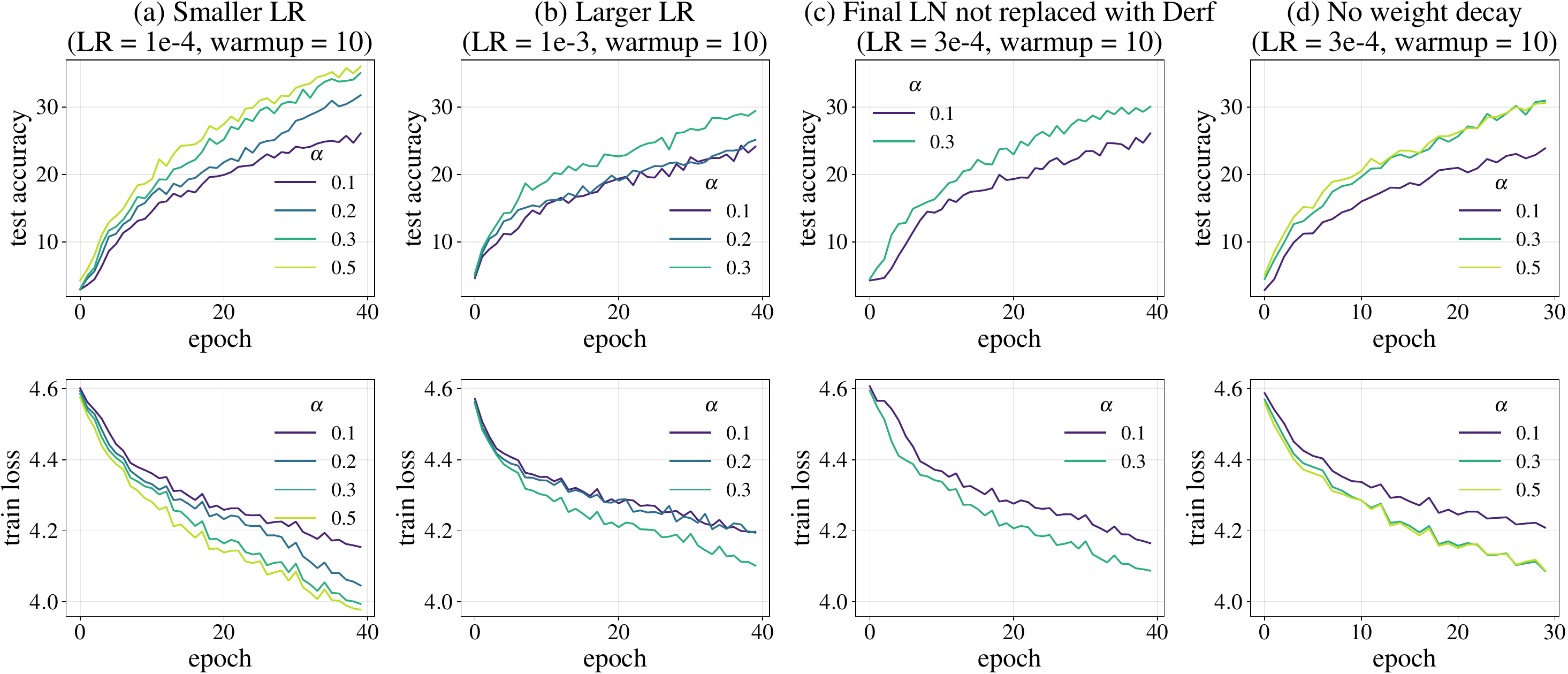}
  \caption{
Effect of initializing Derf with an overly small 
$\alpha$. \textbf{(a)} Smaller learning rate and \textbf{(b)} larger learning rate than in the main experiment in Fig.~\ref{fig:warmup_lr_tiles}(c). \textbf{(c)} Same setup as Fig.~\ref{fig:warmup_lr_tiles}(c), but with the final LayerNorm not replaced by Derf, to rule out an effect of small $\alpha$ acting directly at the output. \textbf{(d)} No weight decay. All four panels suggest that choosing $\alpha$ too small in Derf can slow convergence.
  }
  \label{fig:appendix-small-alpha}
\end{figure}

\textbf{Small-$\alpha$ sweep experiment in other settings.}
Fig.~\ref{fig:appendix-small-alpha} repeats the experiment from Fig.~\ref{fig:warmup_lr_tiles}(c) under several alternative settings: smaller and larger learning rates, no weight decay, and no replacement of the final LayerNorm with Derf. Across these settings, the results suggest that choosing $\alpha$ too small in Derf can slow convergence.

\begin{figure}[tbp]
  \centering
\includegraphics[width=14.0cm]{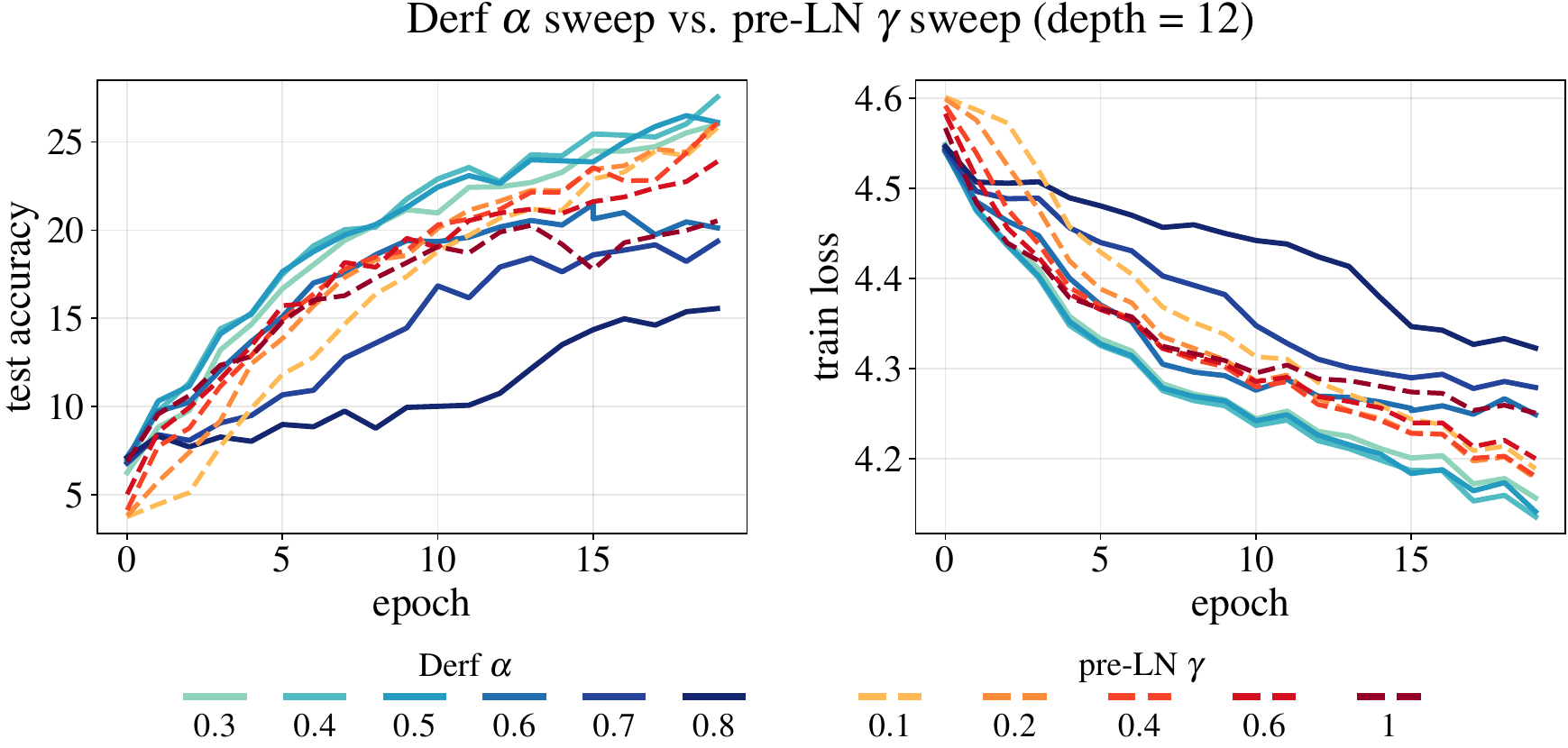}
  \caption{
  Comparison of a Derf $\alpha$ sweep and a pre-LN $\gamma$ sweep at depth $12$, showing test accuracy and training loss over 20 training epochs.
  }
  \label{fig:gamma-sweep}
\end{figure}

\textbf{Pre-LN $\gamma$ sweep vs.\ Derf $\alpha$ sweep.} Fig.~\ref{fig:gamma-sweep} compares pre-LN models with different values of the initialization parameter $\gamma$ against Derf models with different values of $\alpha$. Here, $\bm{\gamma}$ is defined as $\bm{\gamma} = \gamma \bm{1}$.
Note that, in this experiment, the inputs in the pre-LN variant are additionally normalized with LayerNorm before the first transformer block, unlike in the other experiments. As a result, the training curves and test accuracy for the $\gamma=1$ variant may differ from those in the other experiments, such as those shown in Figs.~\ref{fig:train-test-curves-depth}(a) and \ref{fig:train-test-curves-sigma}(a).

\clearpage

\end{document}